%% file: paper.tex
\documentclass[]{bytedance}

\usepackage[toc,page,header]{appendix}

\input{resources/packages}

\input{resources/math_macro}
\usepackage{mathrsfs}
\usepackage{adjustbox}
\usepackage{multirow}
\usepackage{multirow}
\usepackage{multicol}
\usepackage{tcolorbox}
\usepackage{changepage}
\usepackage{graphicx}
\usepackage{amssymb}
\usepackage{array}
\usepackage{bm}
\usepackage{colortbl}
\usepackage{xcolor}

\usepackage{minitoc}

\usepackage{pifont}
\definecolor{mygreen}{RGB}{112, 180, 143}
\definecolor{myred}{RGB}{242, 128, 128}
\newcommand{\cmark}{\textcolor{mygreen}{\ding{51}}}%
\newcommand{\xmark}{\textcolor{myred}{\ding{55}}}%

\title{\methodName{}: Unified Sequential Modeling Activates Multimodal Understanding and Generation}

\author[1]{Huichao Zhang*}
\author[1]{Liao Qu*}
\author[1]{Yiheng Liu*}
\author[1]{Hang Chen*}
\author[1]{Yangyang Song*}
\author[1]{Yongsheng Dong*}
\author[1,2]{Shikun Sun*}
\author[1]{Xian Li*}
\author[1,\dagger]{Xu Wang}
\author[1,\dagger]{Yi Jiang}
\author[1]{Hu Ye}
\author[1]{Bo Chen}
\author[1]{Yiming Gao}
\author[1]{Peng Liu}
\author[1,3]{Akide Liu}

\author[1]{Zhipeng Yang}
\author[1]{Qili Deng}
\author[1]{Linjie Xing}
\author[1]{Jiyang Liu}
\author[1]{Zhao Wang}

\author[1]{Yang Zhou}
\author[1]{Mingcong Liu}
\author[1]{Yi Zhang}
\author[1]{Qian He}
\author[1]{Xiwei Hu}
\author[1]{Zhongqi Qi}
\author[1]{Jie Shao}

\author[1]{Zhiye Fu}
\author[1]{Shuai Wang}
\author[1]{Fangmin Chen}
\author[1]{Xuezhi Chai}
\author[1]{Zhihua Wu}
\author[1]{Yitong Wang}
\author[1]{Zehuan Yuan}
\author[1]{Daniel K. Du}
\author[1]{Xinglong Wu}

\affiliation[1]{ByteDance}
\affiliation[2]{TsingHua University}
\affiliation[3]{Monash University}

\contribution[*]{Core Contributors}
\contribution[\dagger]{Project Lead\vspace{1em}}

\abstract{
We present \methodName{}, a unified decoder-only autoregressive transformer trained on 6 trillion interleaved text-image discrete tokens. By leveraging a unified vision representation within a unified autoregressive architecture, \methodName{} natively activates multimodal understanding and generation capabilities, unlocking abilities of image editing, interleaved content and video generation. 
Motivated by the distinct nature of modalities—where text is strictly sequential and images are inherently hierarchical—we retain next-token prediction for text but adopt next-scale prediction for visual generation. This departs from traditional raster-scan methods, enabling the generation of $1024 \times 1024$ images in just 5 seconds—orders of magnitude faster than comparable AR models. We address the instabilities of multi-scale generation through a robust training recipe. Furthermore, we introduce a prefix-tuning strategy for reinforcement learning. Experiments demonstrate that \methodName{} achieves state-of-the-art performance among unified models and rivals specialized diffusion baselines in visual quality. 
}

\checkdata[Official Page]{\url{https://github.com/ByteVisionLab/NextFlow} }
 
\newcommand{\methodName}{NextFlow}

\begin{document}
\begin{CJK*}{UTF8}{gbsn}

\maketitle

\definecolor{chinese_red}{HTML}{8B4513}
\definecolor{english_blue}{HTML}{4169E1}

\definecolor{mycolor_blue}{HTML}{E7EFFA}
\definecolor{mycolor_green}{HTML}{E6F8E0}
\definecolor{mycolor_gray}{HTML}{ECECEC}
\definecolor{pearDark}{HTML}{2980B9}
\definecolor{lightergray}{HTML}{D3D3D3}

\vspace{-2em}
\begin{figure}[htp]
    \begin{center}
    \includegraphics[width=0.58\linewidth]{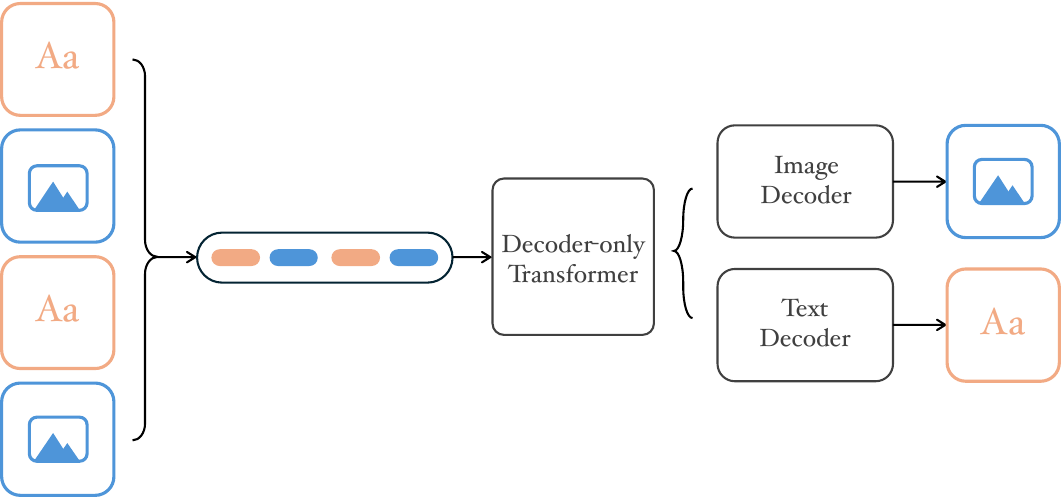}
    \end{center}
    \caption{\textbf{Architecture of \methodName{}.} \methodName{} processes interleaved text-image discrete token sequences as input and generates interleaved multimodal outputs. Text tokens are predicted via next-token modeling, while visual tokens are generated through next-scale prediction.}
    \label{fig:overall_framework}
\end{figure}

\begin{figure}[pt]
\begin{center}
\vspace*{-1.0cm} 
\includegraphics[width=1.0\linewidth]{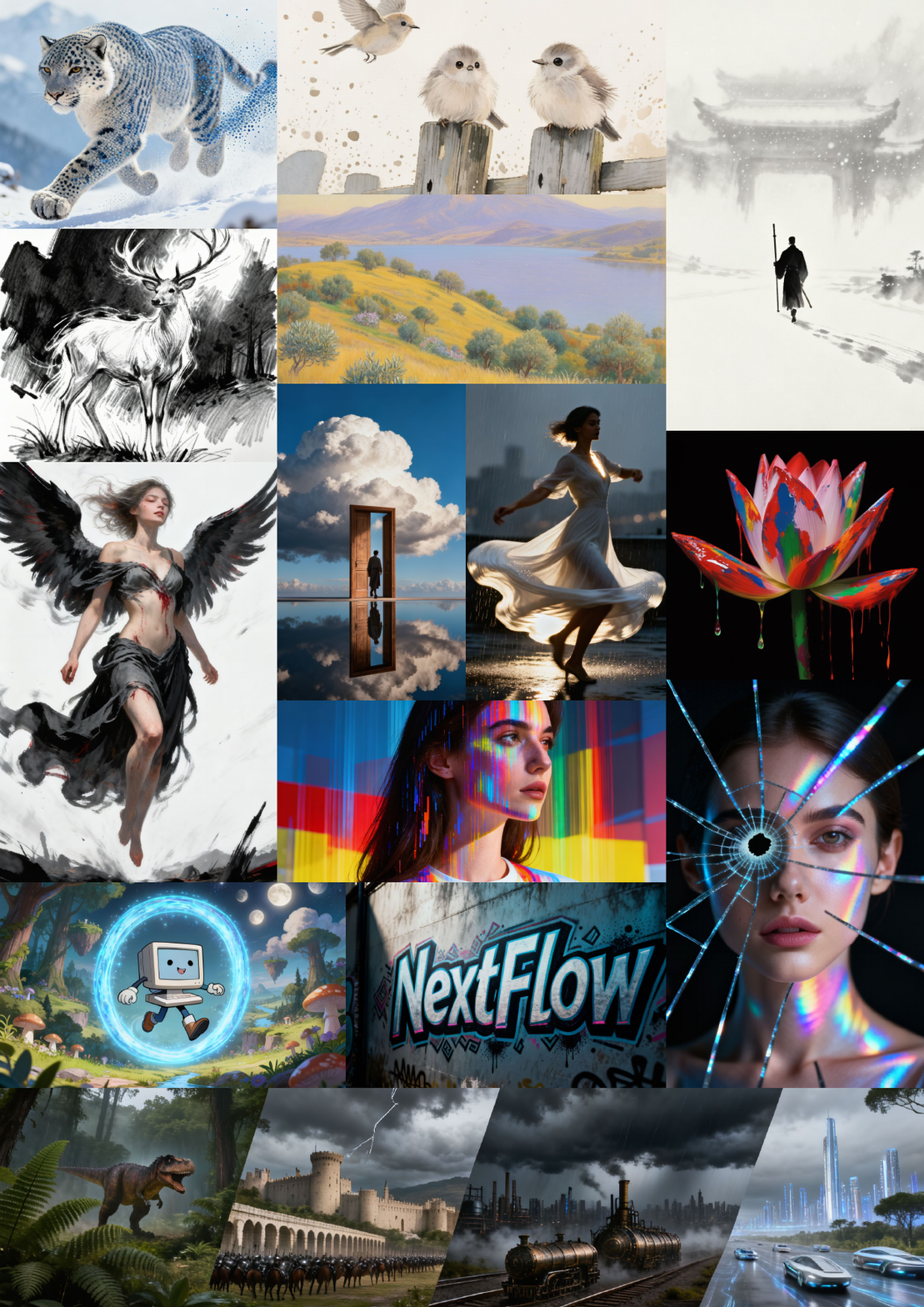}
\end{center}
\label{fig:teaser}
\vspace{-0.6em}
\caption{\textbf{\methodName{} visualization.} Our approach generates high-fidelity images via a pure discrete autoregressive framework, achieving production-grade visual quality.}
\end{figure}

\begin{figure}[pt]
\begin{center}
\vspace*{-1.0cm} 
\includegraphics[width=1.0\linewidth]{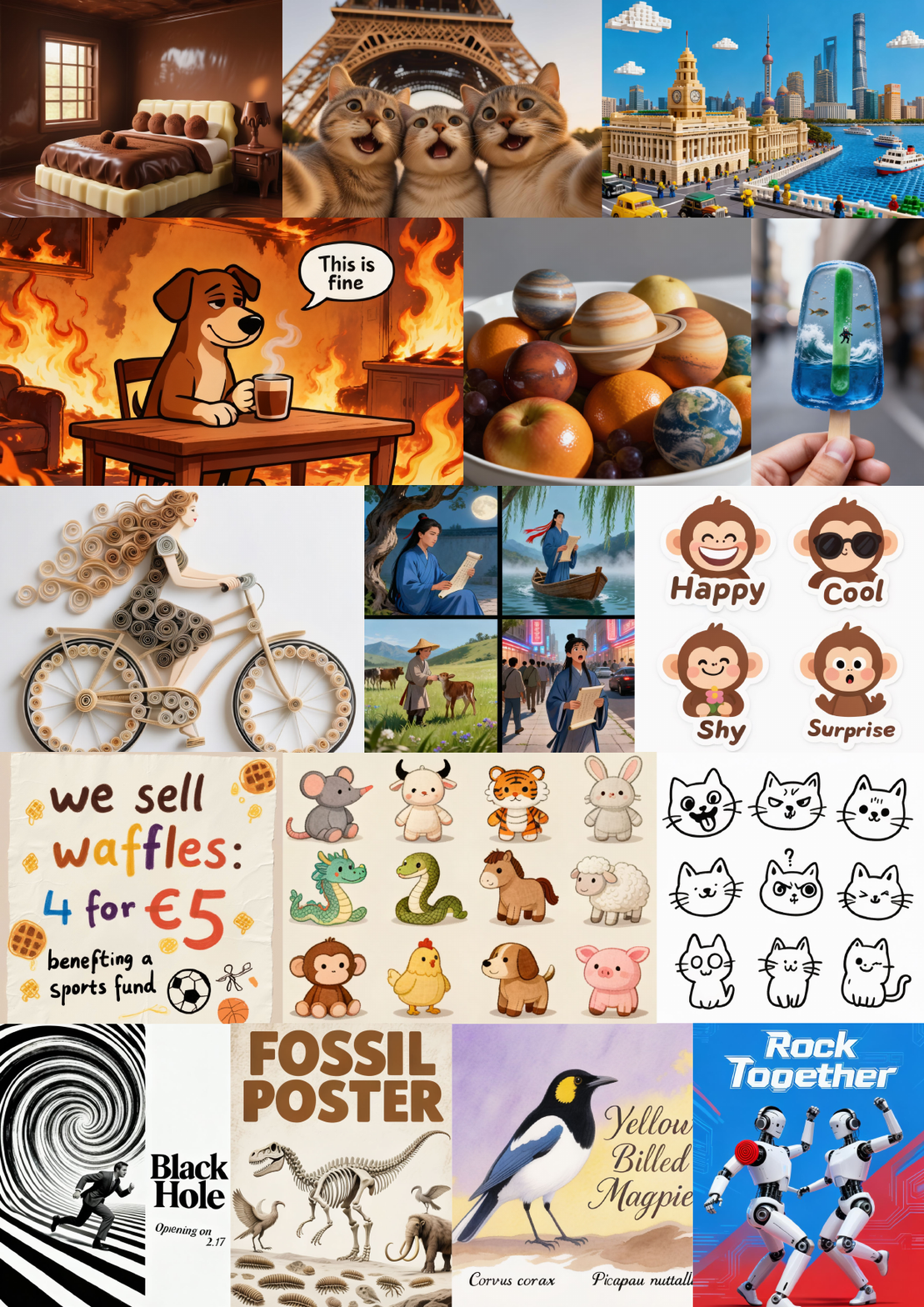}
\end{center}
\label{fig:teaser2}
\vspace{-0.6em}
\caption{\textbf{\methodName{} visualization.} Utilizing next-scale prediction for visual generation, our model can efficiently synthesizes high-quality $1024 \times 1024$ images under 5 seconds.}
\end{figure}

\begin{figure}[pt]
    \centering
    \makebox[\linewidth][c]{\includegraphics[width=1.2\linewidth]{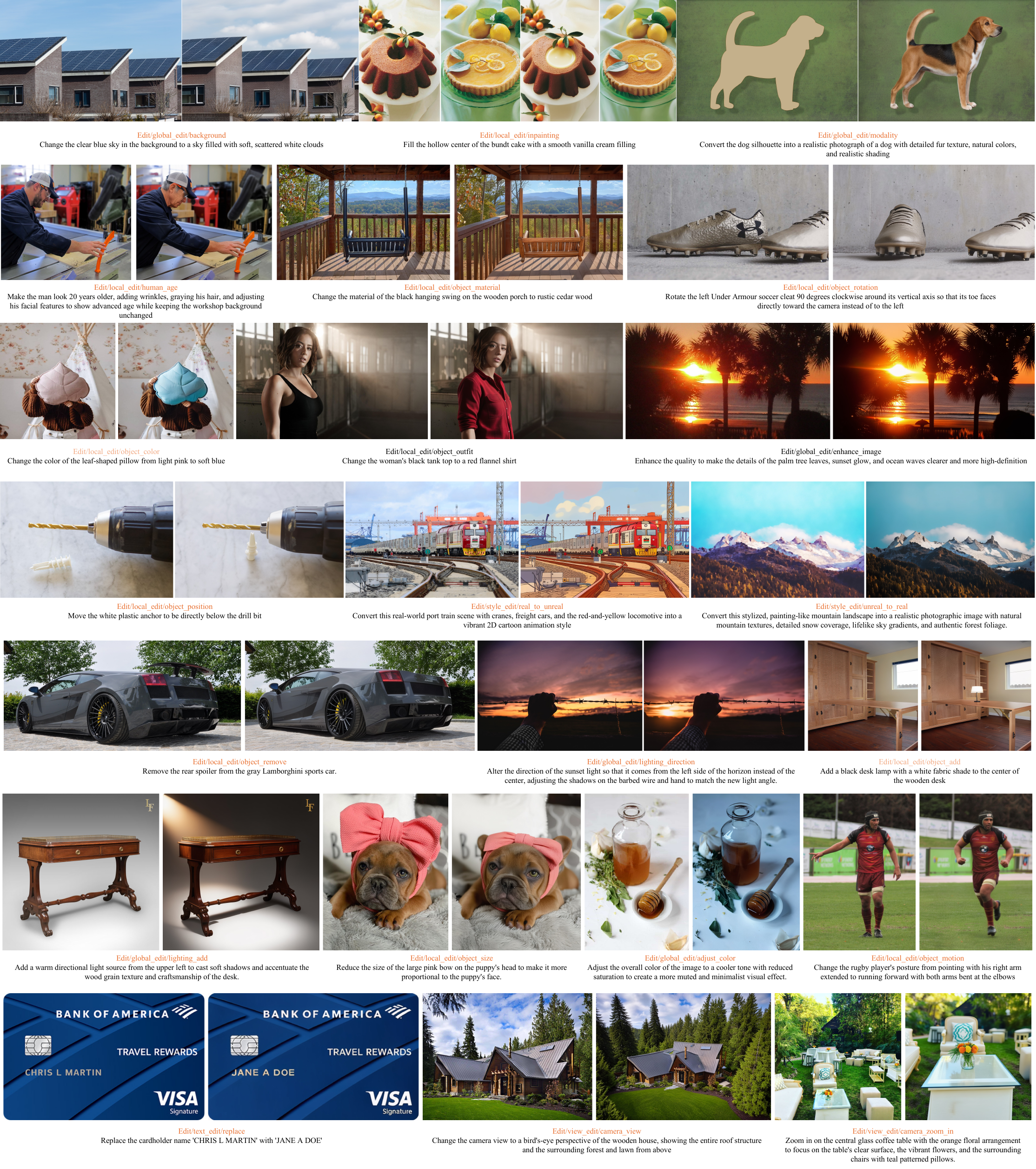}}
    \caption{\textbf{Edit results of \methodName{} on EditCanvas benchmark.}}
    \label{fig:edit_showcase1}
\end{figure}

\clearpage

\tableofcontents

\newpage

\input{sections/A_Introduction}

\input{sections/B_model_architecture}

\input{sections/C_training}

\input{sections/D_infrastructure}

\input{sections/E_data}

\input{sections/F_Model_Performance}

\clearpage

\bibliographystyle{plainnat}
\bibliography{main}

\clearpage

\beginappendix
\input{sections/G_appendix}

\end{CJK*}
\end{document}

%% file: resources/packages.tex
\usepackage{natbib}

\usepackage{CJKutf8}

\usepackage{xargs}  

\usepackage{todonotes}  

\usepackage{multirow}

\usepackage{cleveref}

\usepackage{amsmath}
\usepackage{dsfont}

\usepackage{subcaption}

\usepackage{svg}

%% file: sections/A_Introduction.tex
\section{Introduction}

The pursuit of artificial general intelligence (AGI) has long envisioned a unified system capable of perceiving, reasoning, and creating across diverse modalities. While Large Language Models (LLMs) have achieved mastery in text understanding and reasoning, and Diffusion Models \cite{rombach2022high, SD3, flux} have revolutionized visual generation, these two distinct paradigms remain largely separated. This separation creates a friction: diffusion models excel at pixel-level fidelity but lack the inherent logical reasoning and in-context learning capabilities of LLMs, whereas traditional multimodal LLMs are often restricted to perception only.

Recently,  AR-Diffusion hybrid architecture including Transfusion \cite{transfusion} and Bagel \cite{bagel} demonstrated promising results for unified understanding and generation. 
However, the reliance on two different representations creates a gap between generation and understanding. This separation imposes re-encoding overheads for interleaved tasks and might fundamentally constrain the potential for deep multimodal integration.
On the other side, pure autoregressive efforts like Chameleon \cite{chameleon}, EMU3 \cite{emu3}, and EMU3.5 \cite{cui2025emu35} remain constrained by two fundamental bottlenecks that hinder their practical deployment and multimodal capability.
First, the reliance on the standard raster-scan next-token prediction paradigm for visual generation incurs a prohibitive computational cost at high resolutions. Unlike text, the sequence length of flattened visual tokens grows quadratically with resolution. Consequently, generating a single $1024 \times 1024$ image via raster-scan autoregression can take over 10 minutes \cite{emu3, cui2025emu35}, making these models significantly slower than their diffusion counterparts and impractical for interactive applications.
Second, the visual representations in these models are typically derived from reconstruction-oriented VQ tokenizers. While these tokenizers optimize for pixel-level fidelity, the resulting discrete codes often lack high-level semantic density. This semantic gap fundamentally limits the model's performance on multimodal understanding tasks, as the visual tokens fail to capture the abstract concepts necessary for complex reasoning and alignment with the textual latent space.

In this work, we introduce \textbf{\methodName{}}, a unified sequential modeling framework that activates both multimodal understanding and generation within a single decoder-only transformer. To address the efficiency bottleneck, \methodName{} departs from raster-scan generation, adopting a \textbf{next-scale prediction} paradigm \cite{var}. This hierarchical approach generates visual content from coarse structural layouts to fine-grained details, significantly reducing inference latency to just \textbf{5 seconds for a $\textbf{1024} \times \textbf{1024}$ image}—orders of magnitude faster than raster-scan counterparts. To bridge the semantic gap, we employ a dual-codebook tokenizer \cite{qu2025tokenflow} that decouples semantic and pixel-level features, ensuring both high-level conceptual alignment and fine-grained visual fidelity within a dynamic resolution framework.

Introduction of next-scale prediction in a unified decoder-only structure is non-trivial and presents unique challenges. We scale \methodName{} on a massive corpus of \textbf{6 trillion tokens}, comprising text, image-text pairs, and interleaved multimodal data. Throughout this journey, we identify and resolve key instabilities inherent to next-scale AR generation.
Furthemore, we implement a rigorous post-training pipeline to boost the model performance. Uniquely, we propose a prefix-tuning strategy for Group Reward Policy Optimization (GRPO), which focuses optimization on the coarse-scale "prefixes" that determine global structure. This approach stabilizes RL training and effectively aligns the model with downstream objectives. 
For scenarios demanding hyper-realistic detail, we further integrate an optional diffusion-based decoder that refines the discrete output, pushing the boundaries of visual fidelity without compromising the unified architecture.

\methodName{} demonstrates that a unified AR model can rival state-of-the-art diffusion models in visual quality while retaining the reasoning power of LLMs. Our 7B parameter model achieves competitive performance on text-to-image benchmarks and outperforms specialized models in image editing. Crucially, the unified architecture naturally supports interleaved text-image tasks. \methodName{} can perform Chain-of-Thought (CoT) reasoning to refine prompts before generation or enable in-context learning for zero-shot image editing. Analysis reveals that \methodName{} is highly efficient, requiring $6\times$ fewer FLOPs during inference compared to MMDiT-based diffusion models \cite{SD3} at $1024^2$ resolution, achieving a generation speed of 5s per image.

Our contributions are summarized as follows:
\begin{itemize}
    \item We propose \textbf{\methodName{}}, a unified decoder-only Transformer that activates multimodal understanding, generation, and editing. It employs next-scale prediction for efficient generation and a dual-codebook tokenizer to ensure high semantic density.
    \item We present a robust training recipe validated on \textbf{6 trillion tokens}, introduce a novel RL strategy for multi-scale generation that focuses optimization on coarse scales, and an optional diffusion decoder for enhanced visual detail, establishing a comprehensive training pipeline.
    \item Extensive experiments demonstrate that \methodName{} achieves state-of-the-art performance, outperforming specialized image editing models and rivaling top-tier diffusion models in visual quality. Crucially, we show that a unified AR architecture can be both computationally efficient and structurally simple, offering a superior alternative to complex hybrid architectures.
\end{itemize}

%% file: sections/B_model_architecture.tex
\section{Model Architecture}

\subsection{Tokenizer}

The \methodName{} tokenizer adopts a dual-codebook architecture building upon TokenFlow \cite{qu2025tokenflow}, which simultaneously achieves high-fidelity image reconstruction and semantically rich discrete representations for multimodal understanding. 

Specifically, this design decouples the learning of semantic and pixel-level features while maintaining their alignment via a shared-mapping mechanism. Unlike standard VQ-VAE that relies solely on pixel reconstruction, our quantization process is jointly constrained by both reconstruction fidelity and semantic consistency. By minimizing the weighted summation of semantic and pixel-level distances during codebook lookup, we ensure that the discrete tokens encapsulate both high-level concepts (distilled from the semantic teacher) and fine-grained visual details. 

To accommodate dynamic resolution processing, we upgrade the semantic encoder initialization from \texttt{siglip-so400m}\footnote{https://huggingface.co/google/siglip-so400m-patch14-384} to \texttt{siglip2-so400m-naflex}\footnote{https://huggingface.co/google/siglip2-so400m-patch16-naflex}, enabling variable resolution and aspect ratio processing. When combined with our CNN-based pixel branch, this architecture enables fully dynamic spatial processing, allowing our AR model to train directly at native resolutions without the constraints of fixed input ratios. We employ multi-scale VQ \cite{var} to further enhance the quantization quality. The scale settings are detailed in the appendix \ref{appendix:scale_schedule}.

\subsection{Decoder-Only Transformer}

Our autoregressive framework builds upon a standard decoder-only Transformer architecture, initialized from Qwen2.5-VL-7B \cite{Qwen2.5-VL} to leverage its strong multimodal priors. We extend this architecture to support visual token prediction using a \textbf{next-scale prediction} paradigm. For the newly added visual codebook, we initialize the embeddings directly from the tokenizer's codebook embeddings. Empirically, we find that a unified prediction head achieves comparable performance to separate text and image heads while maintaining architectural simplicity. Therefore, we adopt a single output head for both modalities. The model is trained with cross-entropy loss to predict codebook indices across both modalities.

\begin{figure}[htp]
    \begin{center}
    \includegraphics[width=\linewidth]{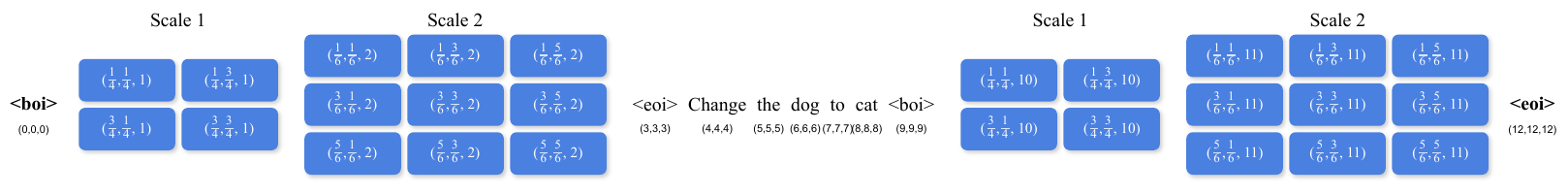}
    \end{center}
    \caption{\textbf{Multi-Scale 3D RoPE for interleaved text-image sequences.} Text tokens employ diagonal positions (e.g., position 3 → [3,3,3]), while vision tokens utilize normalized spatial coordinates with augmented scale indices. For clarity, we show 2 scales using square images as examples. Note that the next-scale prediction paradigm excludes 1×1 features in the input, beginning with 2×2 feature maps.}
    \label{fig:pe}
\end{figure}

\textbf{Positional Encoding.} We introduce Multiscale 3D RoPE to handle interleaved text and multiscale vision tokens, as shown in Figure \ref{fig:pe}. For text tokens at position $t$, we simply replicate the position across all three dimensions: $(t, t, t)$. For vision tokens, we encode spatial and scale information explicitly: Each patch at scale $s$ with grid coordinates $(i, j)$ receives position 
\begin{equation}
(p_x, p_y, p_s) = (\frac{C}{\sqrt{HW}}(i + 0.5), \frac{C}{\sqrt{HW}}(j + 0.5), s)
\end{equation}
where $H \times W$ is the grid size and $C$ is a constant range factor (we omit this constant in \ref{fig:pe} for simplicity). The 0.5 offset means to center the positional encoding within each patch.

This normalized formulation enables resolution-invariant training: by mapping all spatial positions to a fixed range $[0, C]$, grids of different resolutions (e.g. $16 \times 16$ and $32 \times 32$) share the same coordinate space. This eliminates positional extrapolation during high-resolution fine-tuning, as the model encounters only previously learned position ranges.

Following \cite{var}, we incorporate additional learnable scale embeddings for vision tokens. To enhance the model’s ability to adapt to varying resolutions during both training and inference, we further introduce \textit{scale length positional embeddings}. Specifically, we employ sinusoidal encoding over the number of scales, which serves to disambiguate the target output resolution. This design is particularly critical in the VAR framework, where the generative process for images of different sizes follows distinct scale-wise patterns. By explicitly encoding scale length, the model can accordingly adjust its generation paradigm, leading to more coherent and resolution-aware synthesis.

\textbf{Scale Reweight.}
In the next-scale prediction paradigm, earlier scales play a crucial role in determining the overall layout and structure of the image. However, these scales contain significantly fewer tokens compared to later scales, creating an imbalanced loss landscape during training. With uniform token weighting, the model tends to prioritize abundant fine-scale tokens at the expense of coarse-scale structural information, leading to layout degradation, particularly severe at high resolutions where the token count disparity becomes extreme.

To address this imbalance, we introduce scale-aware loss reweighting while maintaining the total vision loss constant. Specifically, we assign scale-dependent weights as: 
\begin{equation}\label{eq:reweight}
k_s = \frac{1}{(h_s \times w_s)^\alpha}
\end{equation}
where $h_s \times w_s$ represents the spatial resolution at scale $s$ and $\alpha$ is a hyperparameter controlling the reweighting strength. This formulation substantially increases the importance of early-scale predictions, ensuring stable structural generation.

\textbf{Self Correction with Residual Features.}
During inference in the next-scale prediction paradigm, tokens within each scale are sampled independently using top-k/top-p sampling. This independence can introduce local conflicts: adjacent positions may sample semantically redundant content due to the lack of joint modeling. More fundamentally, autoregressive models suffer from exposure bias \cite{bengio2015scheduled, ren2025beyond, chen2024diffusion, huang2025self}: during training with teacher forcing, the model always conditions on ground-truth tokens from previous steps, but at inference time, it must condition on its own predictions. This train-test mismatch means that errors from earlier scales, which the model was never exposed to during training, tend to propagate and amplify through the generation process \cite{qu2025tokenflow, liu2025detailflow}.

To address this, we introduce a self-correction mechanism during training, inspired by \cite{liu2025detailflow,han2025infinity}. Rather than deterministically selecting the closest codebook index during encoding, we sample from a multinomial distribution over the top-k nearest indices, while the model continues to predict the top-1 index as the target. This trains the model to correct suboptimal choices from previous scales.
In original VAR framework, the input features for each scale are accumulated from all previous scales. However, we find that applying self-correction to accumulated features in our decoder-only architecture leads to performance degradation. We hypothesize that self-correction significantly complicates the input feature space, creating a mismatch with text features that are directly retrieved from the codebook.

We modify the visual input to use residual features directly from the codebook without accumulation. The features of each scale are independently retrieved and upsampled as needed. This approach significantly constrains the complexity of visual input feature space, yielding substantial performance improvements and reducing local artifacts in generated images.

\subsection{Optional Diffusion Decoder}

\begin{figure}[htp]
    \begin{center}
    \includegraphics[width=0.6\linewidth]{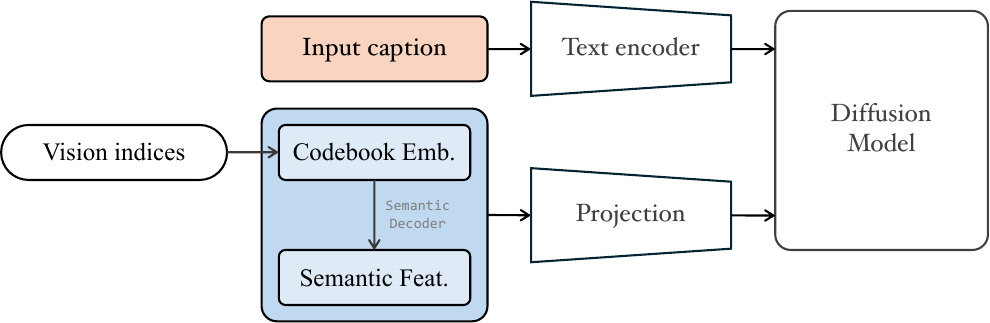}
    \end{center}
    \caption{\textbf{Architecture of the optional diffusion decoder.}}
    \label{fig:diffusion_decoder}
\end{figure}

While our default VQ decoder achieves satisfactory reconstruction fidelity with high inference efficiency, the discrete quantization process inevitably results in the loss of high-frequency details. To push the boundaries of visual quality, particularly for generation tasks requiring photo-realistic details, we introduce an \textbf{optional diffusion-based decoder} acting as a refinement module. 

\textbf{Conditioning Mechanism.} After the next-scale visual index prediction, we obtain the corresponding embeddings from both the semantic and pixel codebooks. The semantic embeddings can be processed through the tokenizer's semantic decoder to yield high-dimensional semantic features—representations that are explicitly aligned with ground-truth semantic features during tokenizer training \cite{qu2025tokenflow}. We concatenate these three elements (semantic embeddings, pixel embeddings, and decoded semantic features), pass them through a linear projection layer, and feed the result into the diffusion model as a visual condition. Simultaneously, the diffusion model integrates the image caption via the text branch.

\textbf{Usage and Trade-offs.}
Within the \methodName{} framework, we observe that the diffusion decoder significantly mitigates detail degradation, particularly in challenging regions such as small-scale faces and text. However, this generative refinement introduces a trade-off: the stochastic nature of the diffusion process may alter fine-grained structures, potentially impacting performance in tasks requiring strict spatial consistency, such as local editing or identity preservation. 
\textit{Unless explicitly stated otherwise, the diffusion decoder is disabled in our reported experiments, including both quantitative evaluations and qualitative visualizations.}

%% file: sections/C_training.tex
\section{Training Odyssey}
\begin{figure}[ht]
    \begin{center}
    \includegraphics[width=\linewidth]{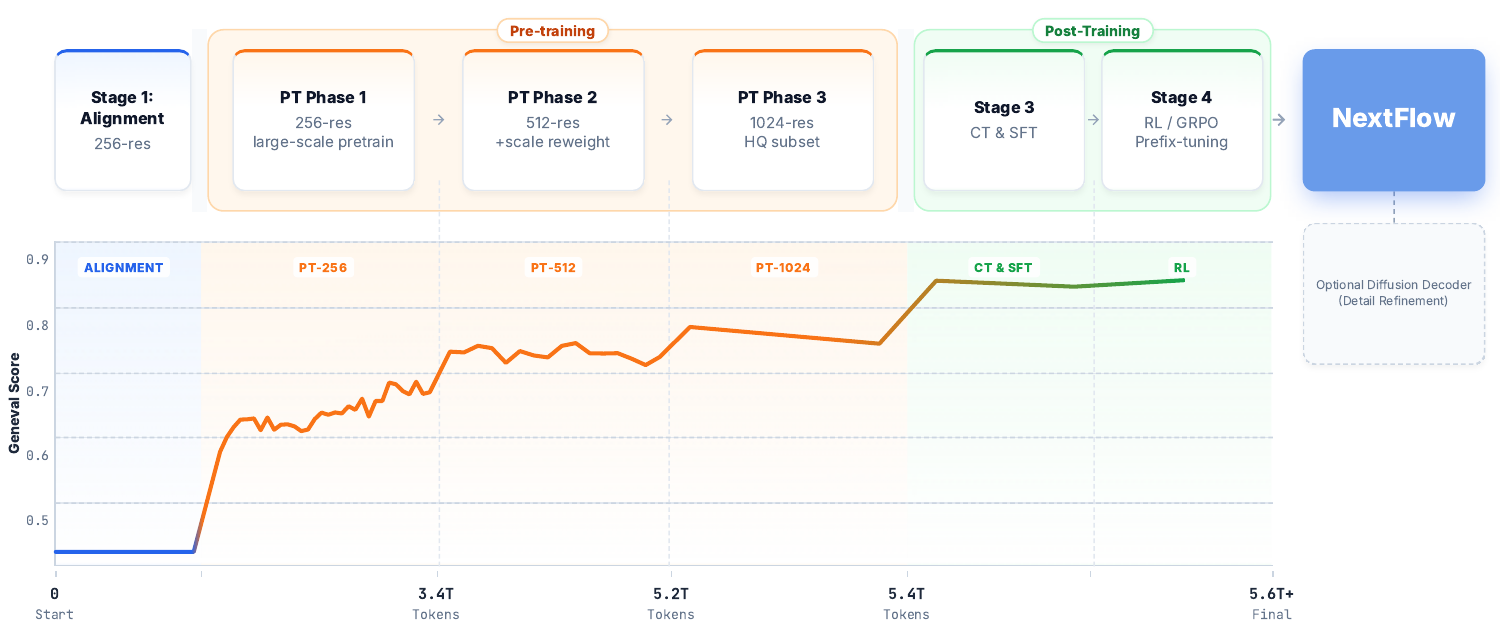}
    \end{center}
    \caption{\textbf{Overall training pipeline and the evolution of the Geneval score.} The plot below tracks the Geneval score \cite{ghosh2023geneval} against the cumulative number of training tokens (in trillions). Shaded regions distinguish the distinct stages of our pipeline.}
    \label{fig:training_pipeline}
\end{figure}

Experimental progress is rarely linear. Throughout training, we encountered a variety of challenges. In this section, we present our training odyssey—a chronological account of our training recipes, the issues that arose at each stage, and the solutions we developed to address them. The overall training pipeline is shown in \ref{fig:training_pipeline}.

\subsection{Tokenizer Training}

The original TokenFlow framework \cite{qu2025tokenflow} initializes the semantic encoder from pretrained model, while leaving the pixel encoder train from scratch.
This leads to semantic features dominating early optimization and ultimately limits reconstruction quality.
To mitigate this, we adopt a multi-stage training strategy: We first independently train the pixel branch to establish strong reconstruction capabilities. We then initialize all components—semantic encoder, pixel encoder/decoder, and pixel codebook—from their pre-trained checkpoints and train them jointly. This ensures that both branches begin from a favorable initialization, significantly accelerating convergence and improving final performance. Finally, we double the capacity of the pixel decoder, following architectural insights from \cite{chang2023muse, qu2025tokenflow}, and fine-tune it separately. This stage substantially reduces local artifacts and enhances fine details such as small faces and text.

The tokenizer is trained on high-fidelity images with oversampling of face-containing samples to improve the preservation of facial detail. We find that randomly dropping 50\% of VAR scales enhances the robustness of numerical distributions in the codebook and results in better reconstruction. Additionally, we enforce \texttt{fp32} precision during quantization to maintain numerical stability.

\begin{figure*}[htp]
    \begin{center}
    \includegraphics[width=\linewidth]{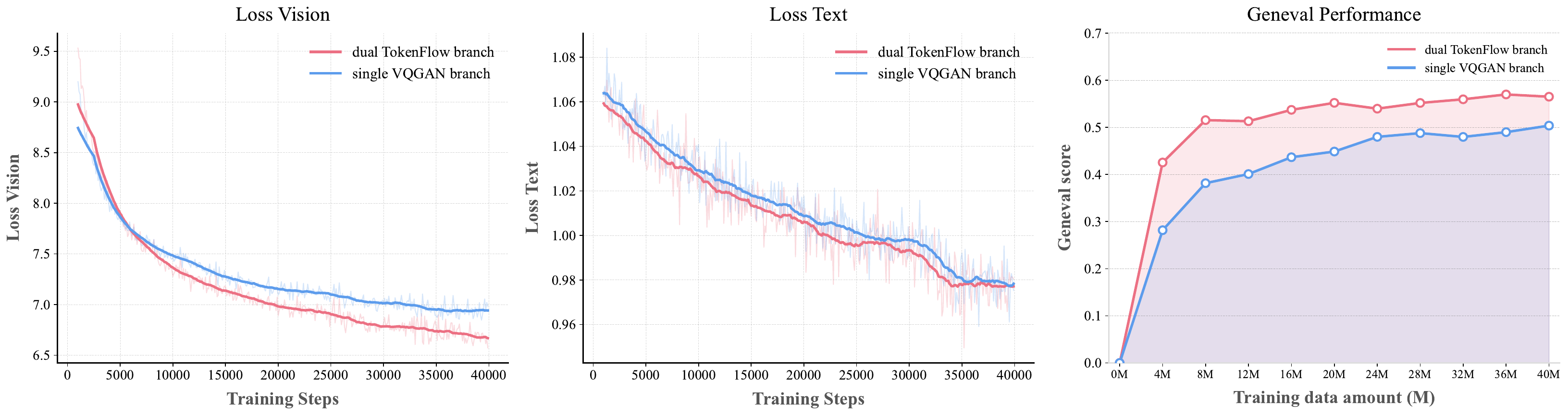}
    \caption{\textbf{Tokenizer Comparison.} We compare the training loss and generative performance (GenEval \cite{ghosh2023geneval}) of our dual branch tokenizer against a single branch VQGAN baseline over 40k steps. Despite slightly lower reconstruction PSNR, the dual-branch architecture demonstrates faster convergence and superior generative capability due to semantically aligned latent structures.}
    \label{fig:single_dual_comparison}
    \end{center}
\end{figure*}

\textbf{Advantage of TokenFlow-style tokenizer.} 
We validate the efficacy of the TokenFlow-style dual-codebook design by comparing it against a standard single-branch VQGAN. While the single-branch baseline yields marginally higher raw reconstruction fidelity (+0.5 PSNR at $256^2$), it proves less effective for generative modeling. As shown in Figure~\ref{fig:single_dual_comparison}, under identical training protocols (40k steps, $\sim$40M samples), the dual-branch tokenizer achieves significantly lower vision loss and consistently superior GenEval scores. We attribute this to the semantic constraints inherent in the dual-branch architecture; these constraints shape a latent space that is structurally easier for the autoregressive model to learn, aligning with findings in REPA \cite{yu2024representation} and VA-VAE \cite{yao2025reconstruction}.

\subsection{Decoder-only Transformer Training} 

\subsubsection{Before the Journey Begins}

\textbf{Output Head Design.} Prior to large-scale training, we evaluated two architectural variants: a \textit{shared output head} that predicts both text and visual tokens within a unified vocabulary, and \textit{separate modality-specific heads} that employ distinct prediction mechanisms for text and vision. This design choice reflects a common trade-off in multimodal modeling: shared heads maximize parameter sharing and simplify training \cite{chameleon,li2025onecat}, while separate heads allow modality-specific optimization \cite{show-o,xie2025show,kou2024orthus}. To systematically compare these approaches in our decoder-only framework, we conducted a controlled ablation under a lightweight setting using 5M alignment and 5M supervised fine-tuning (SFT) samples.
During alignment, we fine-tuned the adapter and the expanded shared output head for the single-head model, keeping the remaining parameters frozen. For the dual-head model, we fine-tuned the additional vision head along with the adapter. All parameters were fine-tuned during the SFT phase.

\begin{figure*}[htp]
    \begin{center}
    \includegraphics[width=\linewidth]{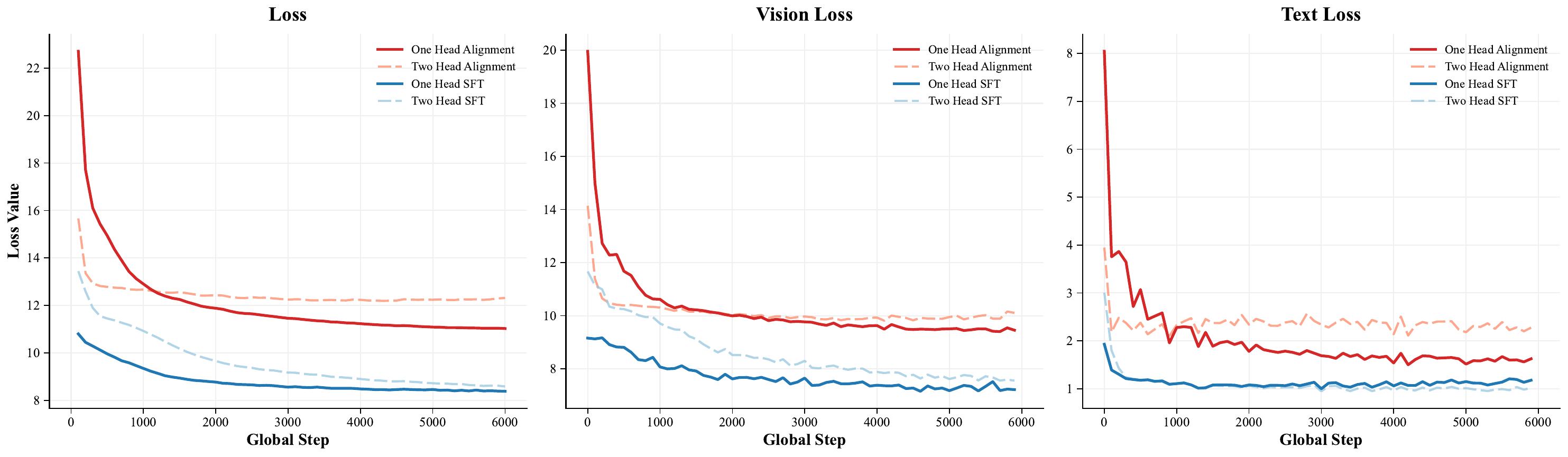}
    \caption{\textbf{Loss curves for total loss, vision loss and text loss during the lightweight alignment/SFT experiments.} Losses are computed using cross-entropy loss.}
    \label{fig:loss}
    \end{center}
\end{figure*}

As illustrated by the training losses in Figure \ref{fig:loss}, the single-head architecture consistently demonstrates superior performance. Specifically, it achieves lower total loss and vision loss throughout both the alignment and SFT phases. Furthermore, its text loss during the SFT phase is comparable to that of the dual-head counterpart. Given its architectural simplicity and empirically better performance, we adopted the shared, single-head design for all subsequent large-scale experiments.

\textbf{Impact of Self Correction.} 
We validate self-correction strategies on a 2B parameter model. All experiments are conducted using a high-quality subset of LAION \cite{schuhmann2022laion} and utilize a tokenizer with a 16,384-size codebook. Each 10,000 training steps corresponds to 5M image-text pairs.

\begin{figure}[htp]
    \begin{center}
    \includegraphics[width=0.5\linewidth]{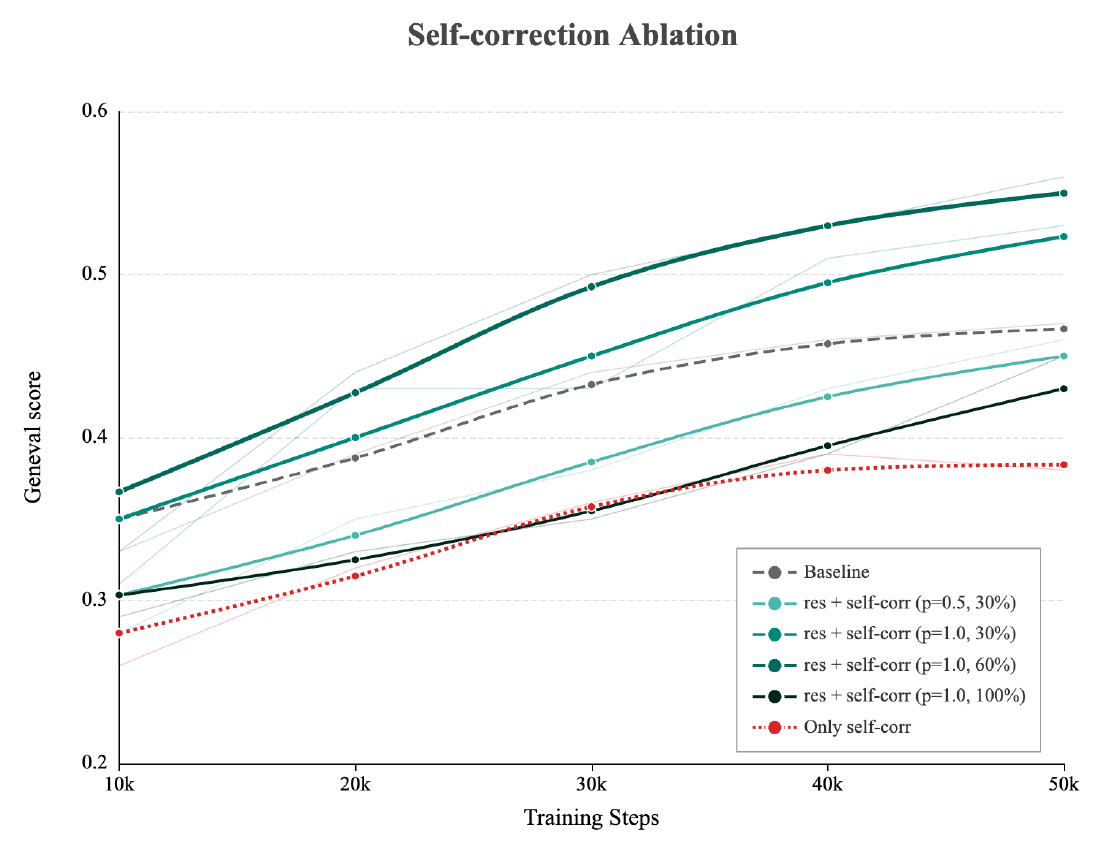}
    \end{center}
    \caption{\textbf{Ablations of self-correction strategy.} Direct application of self-correction degrades performance, while combining it with residual features yields substantial improvements. Using self-correction on 100\% training samples and 60\% tokens per scale achieves optimal performance.}
    \label{fig:ablation1}
\end{figure}

Figure~\ref{fig:ablation1} reveals an intriguing phenomenon: directly applying self-correction with accumulated features (green line, $p=1.0$, 30\%) degrades performance below the non-corrected baseline (origin line). However, replacing accumulated VAR features with residual features leads to significant improvements. We attribute this to the fundamental difference in feature space complexity: in our decoder-only architecture, self-correction on accumulated features excessively complicates the input space, creating a mismatch with text tokens that are directly retrieved from the codebook. Residual features effectively constrain this complexity while maintaining consistency with the text modality, enabling self-correction to function as intended.
Further ablations on self-correction intensity reveal that applying correction with probability $p=1.0$ to 60\% of visual tokens per scale achieves optimal performance (0.56 at 50k steps).

\textbf{Impact of Text-only Data.} To preserve the original model’s text capabilities and evaluate whether mixing text-only data harms text-to-image performance, we conducted a small-scale ablation. As shown in Table \ref{tab:text_data_comparison}, incorporating 25\% text data during training does not adversely affect text-to-image generation quality.

\begin{table}[h]
  \centering
  \caption{\textbf{Geneval scores comparing the pure-t2i baseline and the setting with 25\% text mixed.}}
  \label{tab:text_data_comparison}
  \resizebox{0.8\textwidth}{!}{%
  \begin{tabular}{lcccc}
    \toprule
    Setting & 12M (iter-5k) & 24M (iter-10k) & 36M (iter-15k) & 48M (iter-20k) \\
    \midrule
    t2i data only & 0.266 & 0.384 & 0.441 & 0.505 \\
    +25\% text    & 0.265 & 0.404 & 0.454 & 0.499 \\
    \bottomrule
  \end{tabular}
  }
\end{table}

\subsubsection{Alignment}

During the alignment stage, we replace the original ViT in Qwen2.5-VL-7B \cite{Qwen2.5-VL} with our vision tokenizer and expand the vocabulary to include vision codes and image boundary tokens $\langle$boi$\rangle$, $\langle$eoi$\rangle$. We explore two training strategies to align the tokenizer with the language backbone: (1) a two-stage approach that first aligns the connector module and subsequently fine-tunes the output projection layer, and (2) a joint training strategy that optimizes both components simultaneously. To compare these strategies, we monitor performance metrics after an identical supervised fine-tuning (SFT) phase. Under controlled experimental conditions, both methods yield comparable performance. For simplicity, we adopt the joint alignment strategy in our main experiments.
We used 10 million image-text pairs for bidirectional alignment tasks (image captioning and text-to-image generation) during this phase. Training is conducted at 256-level resolution. 

\subsubsection{Pre‑Training} During pre-training, all model parameters except those of the tokenizer are trainable. The training corpus consisted of approximately \textbf{6 trillion} tokens drawn from various sources, including pure text, image-text pairs, editing data, and interleaved multimodal data. We adopt a progressive resolution curriculum across three sub-stages: 256-level, 512-level, and 1024-level pre-training. 

\textbf{256-level Pre-training:} In this stage, we conducted large-scale pre-training using around 2 billion text-to-image samples to enable the model to learn a wide range of visual semantics and establish fundamental image generation capabilities. We mixed pure text and multimodal understanding data to maintain the model's original language and visual comprehension abilities. Additionally, we incorporate 147M interleaved samples, which teach the model to understand relationships across multiple images, laying the foundation for advanced editing capabilities.

\begin{figure}[htp]
    \begin{center}
    \includegraphics[width=\linewidth]{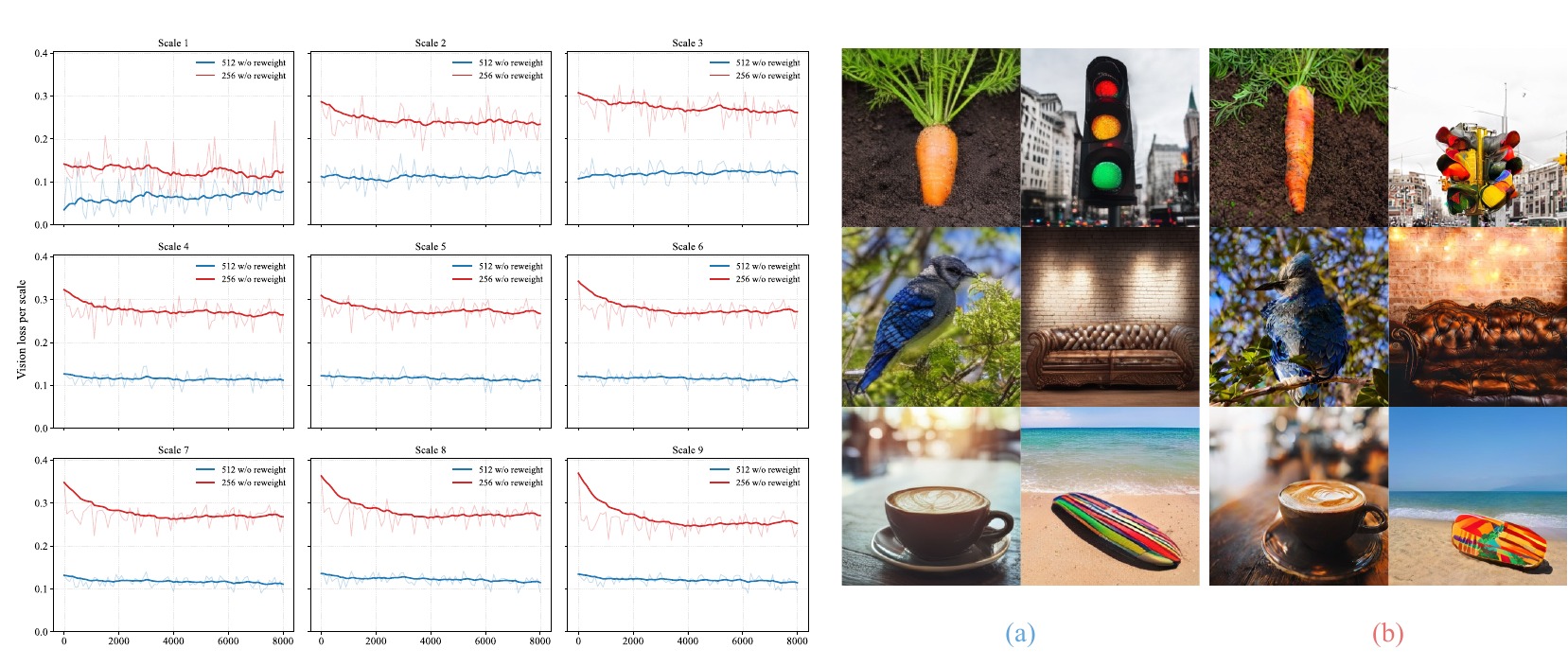}
    \end{center}
    \caption{\textbf{Motivation for scale-reweight strategy.} Left: Comparison of vision loss curves at 256-level and 512-level pre-training stages, including the first 9 VAR scales. A loss increase is observed in lower scales during 512-level training. Right: \textcolor{blue}{(a)} Initial state before 512-level training; \textcolor{red}{(b)} Generation results after training on 200M tokens. Image quality degrades with increased artifacts and structural anomalies, indicating a need for reweighted loss scaling.}
    \label{fig:motivation_reweight}
\end{figure}

\textbf{512-level Pre-training:} \label{512pretrain}
Upon transitioning to $512\times512$ resolution, we observed a significant increase in artifacts and structural degradation in generated images. Although overall vision loss decreased, lower-scale losses exhibited an upward trend (Fig.~\ref{fig:motivation_reweight}). The Geneval score also dropped from 0.67 to 0.57, which we attribute to the substantial increase in the number of tokens per image (approximately 3000 additional tokens) at higher resolution. In VAR architecture, earlier scales play a critical role in the determination of global layout~\cite{var,jiao2025flexvar,qu2025tokenflow}. Equal weighting of tokens disproportionately suppressed learning on these foundational scales. Inspired by flow matching techniques that emphasize difficult intermediate timesteps~\cite{SD3,flux2024}, we introduced a scale-reweighting strategy (Eq.~\ref{eq:reweight}) with $\alpha = 0.9$. This adjustment ensured stable loss reduction on all scales and eliminated localized artifacts.

\textbf{1024-level Pre-training:} At $1024\times1024$ resolution, the exponential growth in token count necessitated the use of a carefully curated subset of 40 million high-quality samples. Despite the reduced size of the data set, the tokenizer achieved significantly finer reconstruction of local details. This stage required minimal data to enable high-resolution generation while substantially improving visual fidelity compared to previous stages.

\subsubsection{Continue-Training and Supervised Fine-Tuning}
After pre-training, we employ a two-phase post-training strategy to refine the models capabilities. First, we conduct continued training (CT) \cite{gong2025seedream, gao2025seedream} on a curated subset of high-quality data to improve the aesthetic quality of the generated images. While pretrained models demonstrate strong general capabilities, they often produce outputs with inconsistent aesthetic standards due to the heterogeneous nature of pre-training datasets. The CT phase addresses this limitation by fine-tuning aesthetically superior samples while preserving prompt adherence and structural accuracy. 
Subsequently, we performed supervised fine-tuning (SFT) using a small set of high-quality conversational data. During SFT, we format the data in a dialogue structure and apply supervision exclusively to the model's responses, enabling more natural and contextually appropriate interactions while further improving generation quality.

\begin{figure}[htp]
    \begin{center}
    \includegraphics[width=0.5\linewidth]{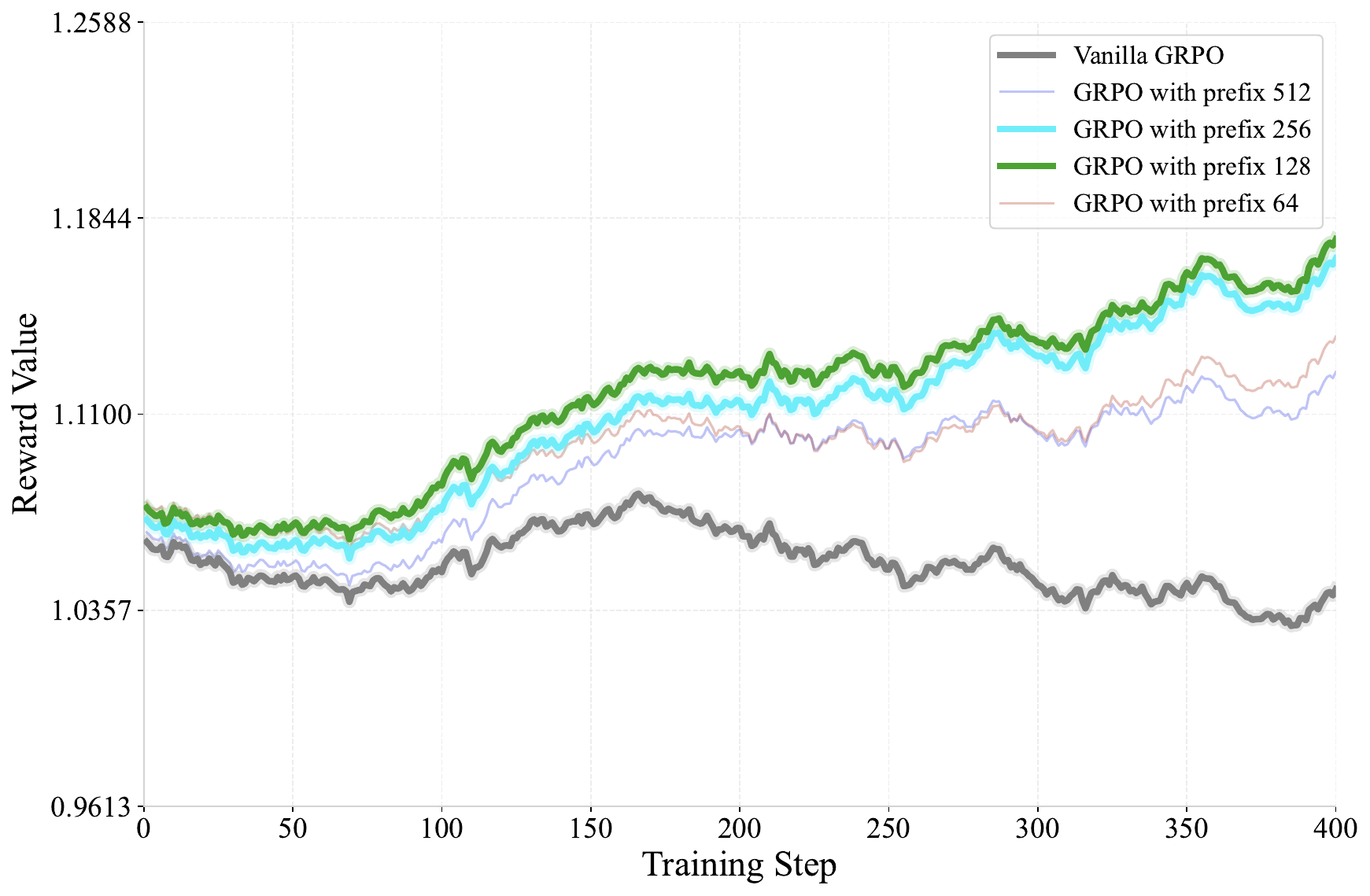}
    \end{center}
    \caption{\textbf{Prefix-tuning strategy for RL fine-tuning in \methodName{}.} Only the policies for the coarse scales are optimized, while policies for the finer scales remain frozen. This approach stabilizes training by focusing high-variance RL updates on the most semantically critical generation steps.}
    \label{fig:rl_prefix}
\end{figure}
    
\subsubsection{Reinforcement Learning}
As our architecture follows a pure autoregressive approach, unlike Bagel-like unified models \cite{bagel, transfusion, liao2025mogao}, we can directly employ Group Reward Policy Optimization (GRPO) \cite{shao2024deepseekmath} for reinforcement learning \cite{argrpo,wang2025simplear,jiang2025t2i,ma2025stage,tong2025delving,zhang2025group}. 
The generation of a VAR sequence can be formulated as a multi-step Markov Decision Process (MDP), where each action $\mathbf{a}_t$ corresponds to generating the token grid for the next resolution level. The policy is defined as:

\begin{equation}
\pi^\theta_t(\mathbf{a}_t \mid \mathbf{s}_{t}) = \prod_{(i,j)} \pi^\theta_{t,(i,j)}(\mathbf{a}_{t,(i,j)} \mid \mathbf{s}_{t}),
\end{equation}

where $\mathbf{a}_{t,(i,j)}$ denotes the token at position $(i,j)$, and $\mathbf{s}_t$ represents the prefix $\{\mathbf{a}_i\}_{i=1}^{t-1}$.
However, applying RL to a multi-scale VAR generation process introduces unique challenges.

As discussed in \ref{512pretrain}, early steps in VAR architecture produce coarse grids with few tokens, while later steps generate fine-grained grids with orders of magnitude more tokens. This imbalance causes learning signals from later steps to dominate optimization, destabilizing training and hindering effective policy updates for the initial steps that define global structure. To address this, we introduce two targeted strategies to stabilize the RL fine-tuning process. The first is \textbf{scale reweight}, same as the strategy adopted at the pretraining stage. The second is \textbf{prefix-tuning strategy}. Since low-resolution steps are most critical for the global layout and semantics of the generated image, we concentrate our RL updates on these formative stages. Given the condition $\mathbf{c}$, which may also contains condition images, we roll out a group of image token sequences $\{\mathbf{s}_T^i\}_{i=1}^G$, and the corresponding decoded images $\{\mathbf{x}^i\}_{i=1}^G$. Within each group, the advantage can be calculated by:
\begin{equation}
    A_i = \frac{R(\mathbf{x}^i,\mathbf{c}) - \textrm{mean}(\{R(\mathbf{x}^i,\mathbf{c})\}_{i=1}^G)}{\textrm{std}(\{R(\mathbf{x}^i,\mathbf{c})\}_{i=1}^G)},
\end{equation}
where $R$ is the reward model. Then the GRPO optimization target is that
\begin{equation}
\begin{aligned}
        L_{GRPO}(\theta) =& \mathbb{E}_{\mathbf{c}\sim \mathcal{C}, \{\mathbf{s}_T^i\}_{i=1}^G\sim \mathbf{\pi}_\theta}\frac{1}{G}\sum_{t=1}^{m}k_t \min\left(\frac{p_\theta(\mathbf{s}_{t+1}^i | \mathbf{s}^i_{t},\mathbf{c})}{p_{\theta_{old}}(\mathbf{s}^i_{t+1} | \mathbf{s}^i_{t},\mathbf{c})}A_i, \mathrm{clip}\left(\frac{p_\theta(\mathbf{s}_{t+1}^i | \mathbf{s}^i_{t},\mathbf{c})}{p_{\theta_{old}}(\mathbf{s}^i_{t+1} | \mathbf{s}^i_{t},\mathbf{c})}, 1-\epsilon, 1+\epsilon\right)A_i \right) \\&-\beta D_{KL}(\mathbf{\pi}_\theta,\mathbf{\pi}_{ref}).
\end{aligned}
\end{equation}

As shown in Fig. \ref{fig:rl_prefix}, we find that fine-tuning only the policies for the first \(m\) (e.g., \(m=8\)) scales, while keeping the policies for the remaining \(T-m\) finer scales frozen, is highly effective. This strategy, termed "prefix-tuning," focuses the limited and high-variance RL signal on the most impactful decisions, avoiding noisy updates to later-stage policies that primarily refine local details. This not only accelerates convergence but also better preserves the generative quality of the pretrained model while adapting its high-level attributes.
Together, these two techniques enable stable and efficient RL-based fine-tuning of \methodName{}, allowing for precise alignment with downstream objectives without sacrificing generative coherence.

\begin{table}[b!]
\centering
\caption{\textbf{Training recipe of \methodName{}.} "M" refers to million. "T" refers to trillion.}
    \label{tab:training_recipe}
\resizebox{1.0\textwidth}{!}{%
\begin{tabular}{c c c c c c c c c c}
\toprule
\multirow{2}{*}{\textbf{Stage}} & \multirow{2}{*}{\textbf{Resolution}} & \multirow{2}{*}{\textbf{Epochs}} & \multirow{2}{*}{\textbf{Learning rate}} & \multicolumn{5}{c}{\textbf{Data volume}} & \multirow{2}{*}{\textbf{Train tokens}} \\
\cmidrule(lr){5-9}
& & & & \textbf{Text} & \textbf{T2I} & \textbf{I2T} & \textbf{Editing} & \textbf{Interleaved} & \\
\midrule
Alignment & 256-level & 1 & 1e-3 & - & 5M & 5M & - & - & 0.01T \\
Pretrain & 256-level & 1 & 1e-4 & 662M & 1891M & 520M & - & 147M & 3.4T \\
Pretrain & 512-level & 1 & 1e-4 & 700M & 345M & - & 20M & 5M & 1.8T \\
Pretrain & 1024-level & 1 & 1e-4 & 47M & 10M & - & 2M & 0.5M & 0.2T \\
\midrule
CT & 1024-level & 1 & 5e-5 & 47M & 9M & - & 2M & - & 0.2T \\
SFT & 1024-level & 1 & 1e-5 & 5M & 1M & - & 0.1M & - & 0.02T \\
\bottomrule
\end{tabular}}
\end{table}

\subsection{Diffusion Decoder Training} 

We explore three model variants: a 1B parameter UNet-based model, and two Transformer-based models with 12B and 18B parameters, respectively.
To preserve the text-conditional capabilities of the pre-trained backbones, we inject the discrete tokens strictly via the visual conditioning branch. We avoid concatenating discrete tokens with text embeddings, as our preliminary experiments indicated this would corrupt the textual semantic information. Structurally, the UNet variant adapts its input convolution to match the input channel dimensions, while the Transformer variants utilize an MLP layer to project the discrete tokens into the model's native dimension.

\textbf{Training Strategy.}
We adopt different training strategies based on model scale and computational constraints. The 1B model undergoes full parameter fine-tuning, whereas the 12B and 18B models are trained using Low-Rank Adaptation (LoRA) \cite{hu2022lora}.
Data selection is also tailored to model capacity: the 1B model is trained on high-quality synthetic data, as its limited capacity struggles to fit the complex distribution of real-world data, often resulting in generation artifacts. Conversely, the larger 12B and 18B models are trained on real-world data to maximize reconstruction fidelity.
We employ a two-stage curriculum: models are first trained at the base resolution, followed by fine-tuning on a $2\times2$ upsampling task to yield sharper high-frequency details. We rely exclusively on the standard diffusion loss, explicitly avoiding pixel-level losses (e.g., MSE), which we found to inflate PSNR metrics while degrading perceptual quality.

%% file: sections/D_infrastructure.tex
\section{Infrastructure}

\methodName{} is pre-trained on a cluster equipped with 1024 GPUs. We employ DeepSpeed ZeRO \cite{rajbhandari2020zero} with gradient checkpointing to enable efficient distributed training at scale.

\begin{figure}[htp]
    \begin{center}
    \includegraphics[width=\linewidth]{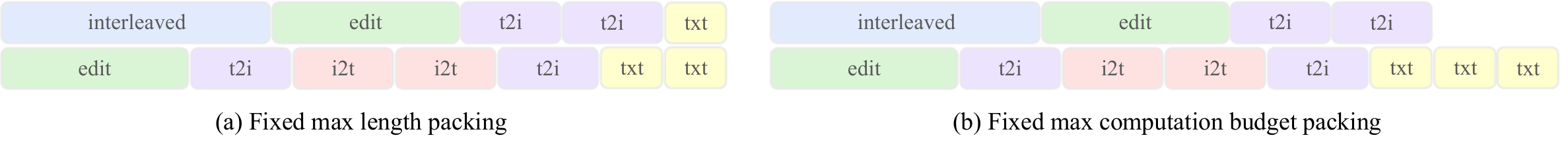}
    \end{center}
    \caption{\textbf{Illustration of different packing strategy.} Fixed max computation budget packing reduces inter-GPU idle time and improving overall training throughput.}
    \label{fig:packing}
\end{figure}

\textbf{Workload Balance.}\label{infra_workload_balance}
The pre-training of \methodName{} involves heterogeneous data types (pure text vs. text-to-image vs. interleaved, varying resolution), which introduces significant computational imbalance across GPUs. We address this through a workload balancing strategy during data packing, as shown in Figure \ref{fig:packing}. Specifically, we precompute the TFLOPS for all sequence lengths and balance workloads accordingly. As demonstrated in Table \ref{tab:packing_efficiency}, this strategy significantly reducing inter-GPU idle time, achieving a $4.1\times$ speedup compared to naive approach.

\begin{table}[h]
    \centering
    \small
    \caption{\textbf{Throughput comparison of different packing strategies on 512-level pretraining stage.} \textit{Tpt} denotes throughput per GPU.}
    \label{tab:packing_efficiency}
    \resizebox{0.6\linewidth}{!}{%
    \begin{tabular}{lcc}
    \toprule
    \textbf{Strategy} & \textbf{Tpt (tokens/s)} & \textbf{Speedup} \\
    \midrule
    Batch Padding & 620.3 & 1.0$\times$ \\
    Fixed Length Packing & 2109.5 & 3.4$\times$ \\
    \textbf{Fixed computation budget packing} & \textbf{2517.4} & \textbf{4.1$\times$} \\
    \bottomrule
    \end{tabular}%
    }
\end{table}

\textbf{High-Performance Kernel.} 
The large codebook size results in significant memory consumption when computing output logits during training. To address this challenge, we adapt FusedLinearCrossEntropy \cite{hsu2025ligerkernel}, which fuses the final linear projection and cross-entropy loss computation into a single kernel, reducing peak memory usage by $\sim$ 20GB per GPU by avoiding the storage of the full logit tensor in memory.
We further identified memory-bound operations and fused them into single kernels to minimize redundant memory access, such as RoPE, RMS norm and Flash-Attention \cite{dao2023flashattention2}. These fused kernels store intermediate results in registers or shared memory, significantly improving arithmetic intensity.

\textbf{Pre-extract image indices.}
To minimize computational overhead during large-scale training, we pre-extract image indices offline across all training stages, thereby eliminating online encoding latency. For each image-text sequence, we precompute and store the index sequence along with the original ordering of the data. This approach allows the encoders to remain offloaded from the GPU during training, significantly reducing memory requirements.
To support self-correction during training, we incorporate a predefined self-correction probability (100\% samples, 60\% tokens per scale). The corresponding input indices and ground-truth indices are stored accordingly.
Once pre-extraction is complete, we perform offline data packing using these precomputed samples, further improving training efficiency as detailed in Sec. \ref{infra_workload_balance}.

%% file: sections/E_data.tex
\section{Data}
\methodName{} is trained on a large scale multimodal dataset which covers a wide range of tasks involving images and texts, including visual understanding, image generation, image editing, image and text interleaved documents generation, text generation. We elaborate how each component of our dataset is built in the following sections.

\subsection{Visual Understanding}
In the pre-training stage, the model learns to perceive and understand images from text supervision. And thus we collect a large number of image captioning samples from open-source image-text datasets. In addition, we further supplement it with 1) text-rich images, comprising scene text and documents of various types such as table, chart, plot, and so on, to enhance the OCR capability of our model; 2) images with associated world knowledge. The image captions are rewritten using Vision-Language Models (VLM) to improve the text quality and comprehensiveness of the description.

\begin{figure}[ht]
    \begin{center}
    \includegraphics[width=1.0\linewidth]{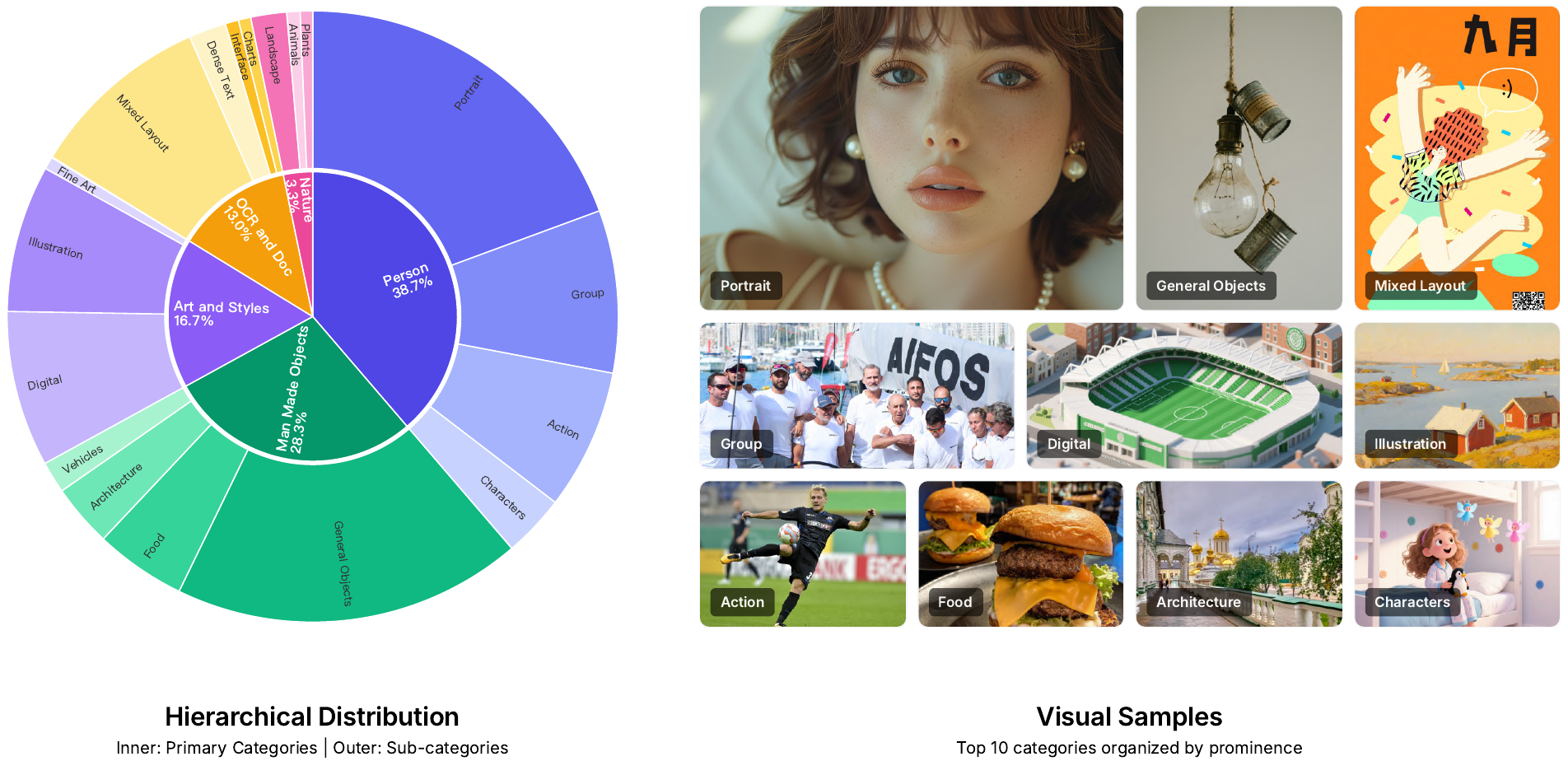}
    \end{center}
    \caption{\textbf{Hierarchical distribution and visual samples of the pre-training dataset.} The left panel illustrates the data composition, where the inner ring represents primary categories and the outer ring details fine-grained sub-categories. The right panel presents representative visual samples across top categories.}
    \label{fig:t2i_distribution}
\end{figure}

\subsection{Image Generation}
We curate a billion scale dataset for text to image generation, with great diversity in terms of image content and image types. The images are gathered from open-source datasets, which exhibit a broad coverage over the content on the Internet; and Megalith \cite{BoerBohan2024Megalith10m}, CommonCatalog \cite{gokaslan2023commoncanvas}, which provide valuable high-quality photographs; as well as our in-house image gallery which further extends the distribution of our data. We first apply a set of heuristic filters to the images and run aesthetic models to rule out images of poor visual quality. For sources whose images we observe have noticeably imbalanced content distribution, we employ a image topic classifier using pretrained SigLip2 in zero-shot manner, to categorize images into topics like landscape, human, animal, plant, food\&drinks and so on. Then data resample is performed to obtain a balanced distribution. As shown in recent works \cite{SD3}, caption plays a crucial role in training image generation models with strong instruction following ability. Therefore we caption all the images using VLM to obtain accurate and detailed descriptions. In the CT stage, we introduce a small amount of synthetic data to further boost the aesthetics of our model's generation.

\begin{figure}[ht]
    \begin{center}
    \includegraphics[width=1.0\linewidth]{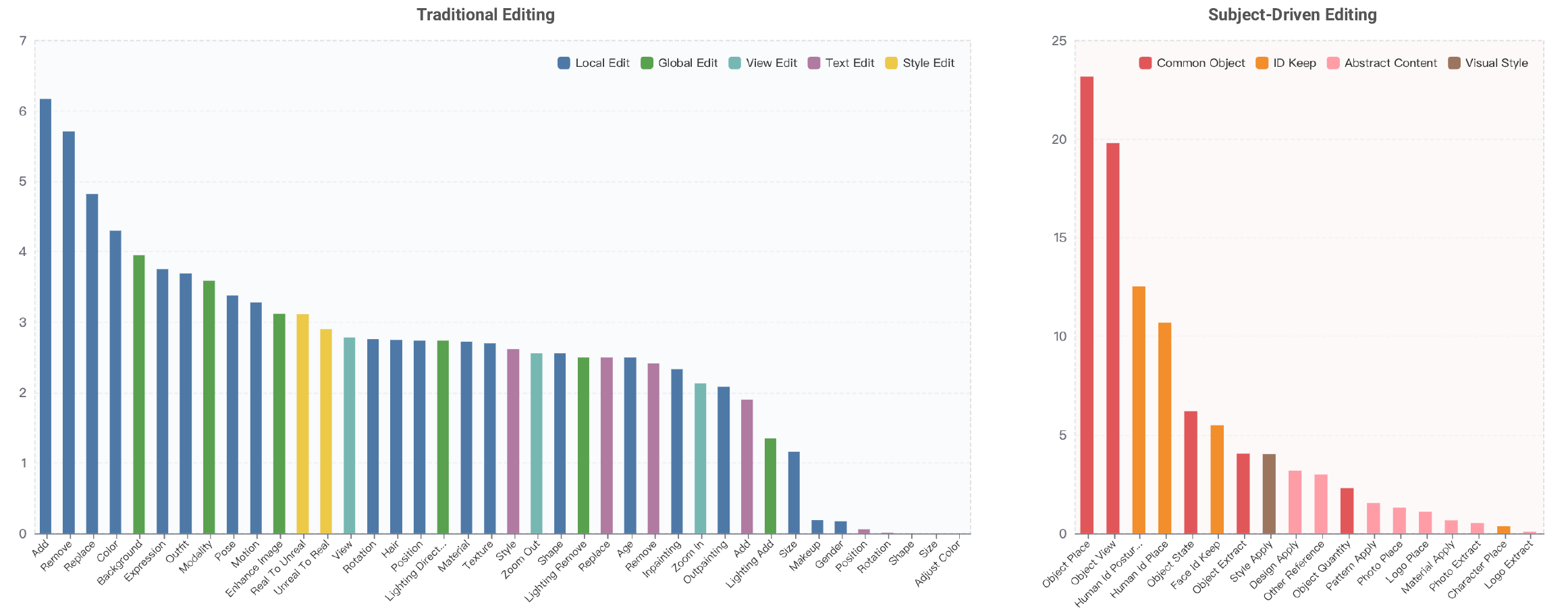}
    \end{center}
    \caption{\textbf{Task distribution of the image editing dataset.} The data is mainly categorized into two streams: Traditional Editing (left) and Subject-Driven Editing (right). The left panel illustrates the frequency of specific editing operations (e.g., Add, Remove, Replace) colored by their high-level semantic types (Local, Global, View, Text, Style). The right panel displays the distribution of subject-driven tasks, highlighting the diversity in preserving object identity (ID Keep), common objects, abstract content, and visual styles.}
    \label{fig:edit_distribution}
\end{figure}

\subsection{Image Editing}
For clarity, we categorize image editing into two distinct tasks: traditional editing and subject driven generation. Traditional editing tackles the problem of transforming an image either locally or globally, while the resulted image always follows the input image in spirit, for example, local editing like object removal, object insertion and global editing like lighting modification, stylization, as well as view editing like zooming in/out, changing the camera's view angle. In contrast, subject driven image generation aims to create an image of new content which only shares the target subject with the input image, for example, generating an image of the dog running on a grass field, given its close-up photo as input.

\subsubsection{Traditional Editing}
We start with open-source traditional editing datasets, including UniWorld-V1 \cite{lin2025uniworld}, RealEdit \cite{sushko2025realedit} and ShareGPT-4o-Image \cite{chen2025sharegpt}. First, we filter edit pairs with mismatched image resolution. Then we identify and remove low quality subsets which contain over 5\% bad edits from these datasets through manual examination, followed by a VLM assessment at a granularity of samples. An edit is considered as "bad" if the target image does not respond to or closely follow the edit instruction. The resulted dataset exhibits a task distribution biased towards common and simple editing tasks such as adding, removing an object or replacing an object with another, which substantially limits the model's capability to perform complex, varied real-world editing. As a result, we build an additional synthetic dataset which span over a diverse spectrum of traditional editing tasks. The task distribution is significantly more balanced than the open-source counterpart as illustrated in Figure \ref{fig:edit_distribution}.

\subsubsection{Subject-driven Generation}
We observe that subjects like human, animal, objects and even abstract content, often reoccur across real world data. And this fact indicates there are numerous natural demonstrations for diverse tasks of subject-driven generation. We collect images that co-occur on the Internet and utilize VLM to identify image pairs sharing common subjects, which turns out to be a scalable way to build the training data. Compared to workflow methodology which heavily relies on generative models, our method inherently produces data with a significant advantage in terms of task diversity as well as subject consistency.

\subsection{Interleaved Generation}

To endow the model with the ability to generate coherent interleaved image-text sequences, we construct a large-scale video-text dataset. We treat video clips as sequences of interleaved frames and leverage the temporal continuity of video to model narrative progression. Our data pipeline aggregates sources including OmniCorpus-CC, OmniCorpus-YT \cite{li2024omnicorpus}, and Koala36M \cite{wang2025koala}, processed through a rigorous multi-stage filtering strategy.

\noindent\textbf{Quality and Heuristic Filtering.} 
We apply strict quality controls to the raw video corpus, specifically targeting the Koala36M subset. To minimize fragmentation errors during training, we discard long clips exceeding 20 seconds. Visual quality is ensured by filtering for high aesthetic scores ($>4.3$, retaining the top 30\%) and clarity ($>0.7$, retaining the top 50\%). Furthermore, to avoid degenerate solutions where the model generates static images, we require a motion score greater than 4. This pipeline removes approximately 75\% of the raw data, resulting in a refined subset of 8M clips free from blur, static scenes, and low-aesthetic content.

\noindent\textbf{Semantic Balancing.}
To prevent the model from overfitting to dominant categories, we employ SigLIP \cite{siglip} to classify video content. We downsample overrepresented classes, removing 50\% of clips containing "person" (reducing the total volume by 20\%) and 50\% of "television news broadcast" clips (reducing the total by 10\%). This results in a balanced corpus of approximately 5.3M clips.

\noindent\textbf{Motion-Adaptive Frame Selection.}
We extract frames at a baseline rate of 0.5 FPS (one frame every 2 seconds). However, uniform sampling often captures redundant frames in static scenes or blurry frames during rapid camera movement. We introduce an optical flow-based filtering mechanism using RAFT \cite{teed2020raft} to refine frame selection:
\begin{itemize}
    \item \textit{Static Filtering:} Frames with negligible optical flow magnitude are discarded.
    \item \textit{Camera vs. Object Motion:} We analyze the variance of flow directions. Low variance implies global camera movement (pan/zoom), which we discard to focus on semantic content changes. High variance indicates independent object motion, which is preserved.
    \item \textit{Large Displacements:} Frames exhibiting significant structural changes (high flow magnitude) are retained to capture scene transitions.
\end{itemize}
During training, we constrain the input to a maximum of 5 frames at $512$px resolution or 3 frames at $1$k resolution. 

\noindent\textbf{Transition Text Generation.}
Finally, to bridge the visual gaps between selected frames, we utilize VLMs to generate coherent transition texts, effectively converting the video clips into interleaved image-text documents.

\begin{table*}[ht]
    \centering
    \caption{\textbf{Comparison of text-to-image generation ability on DPG \cite{dpg} benchmark.} For AR Models, the best results are in \textbf{bold} and the second-best are \underline{underlined}.}
    \resizebox{0.7\linewidth}{!}{%
    \begin{tabular}{lcccccc}
        \toprule
        \textbf{Model}           & \textbf{Global} & \textbf{Entity} & \textbf{Attribute} & \textbf{Relation} & \textbf{Other} & \textbf{Overall$\uparrow$} \\
        \midrule
        \multicolumn{7}{l}{\textbf{Diffusion Models}} \\
        \addlinespace
        SDXL \citep{sdxl}             & 83.27  & 82.43  & 80.91     & 86.76    & 80.41 & 74.65    \\
        DALL-E 3 \citep{dalle3}         & 90.97  & 89.61  & 88.39     & 90.58    & 89.83 & 83.50     \\
        FLUX.1 [Dev] \citep{flux}       & 74.35  & 90.00     & 88.96     & 90.87    & 88.33 & 83.84    \\
        SD3 Medium \citep{SD3}       & 87.90   & 91.01  & 88.83     & 80.70     & 88.68 & 84.08    \\
        GPT Image 1 [High]~\citep{gptimage} &  88.89     & 88.94  & 89.84      & 92.63    & 90.96     & 85.15     \\
        HiDream-I1-Full \citep{cai2025hidream}          & 76.44  & 90.22  & 89.48     & 93.74    & 91.83 & 85.89    \\
        Seedream 3.0 \citep{gao2025seedream} &  94.31     & 92.65  & 91.36      & 92.78    & 88.24     & 88.27     \\
        Qwen-Image \cite{QwenImage} & 91.32 & 91.56 & 92.02 & 94.31 & 92.73 & 88.32 \\

        \midrule
        
        \multicolumn{7}{l}{\textbf{AR Models}} \\
        \addlinespace
        TokenFlow-XL \cite{qu2025tokenflow} & 73.38 & 78.72 & 79.22 & 81.29 & 85.22 & 71.20 \\
        Emu3-Gen \citep{emu3}         & 85.21  & 86.68  & 86.84     & 90.22    & 83.15 & 80.60     \\
        Janus-Pro-7B \citep{januspro2025}     & 86.90   & 88.90   & 89.40      & 89.32    & 89.48 & 84.19    \\
        NextStep-1 \cite{team2025nextstep} &  - & - & - & - & - & 85.28 \\
        EMU3.5 \cite{cui2025emu35} &  - & - & - & - & - & \underline{88.26} \\

        \rowcolor{mycolor_blue}
        \textbf{\methodName{}} & \textbf{92.40} & \underline{90.05} & \underline{90.51} & \textbf{92.72} & \underline{91.14} & 86.00 \\
        \rowcolor{mycolor_blue}
        \textbf{\methodName{}-RL} & \underline{92.09} & \textbf{93.48} & \textbf{91.92} & \underline{92.08} & \textbf{93.22} & \textbf{88.32} \\
        \bottomrule
    \end{tabular}}
    \label{tab:dpg}
\end{table*}

\subsection{Text Generation}

To ensure \methodName{} maintains strong pure-text instruction following and reasoning capabilities, we incorporate high-quality text-only data into the training mix. We source general-purpose instruction data from Nemotron-CC HQ \cite{su2025nemotron} to maintain conversational fluency and general knowledge. Additionally, we integrate mathematical reasoning data from MegaMath \cite{zhou2025megamath} to enhance the model's logical reasoning and problem-solving abilities.

%% file: sections/F_Model_Performance.tex
\begin{figure}[pt]
\begin{center}
\vspace*{-1.0cm} 
\includegraphics[width=1.0\linewidth]{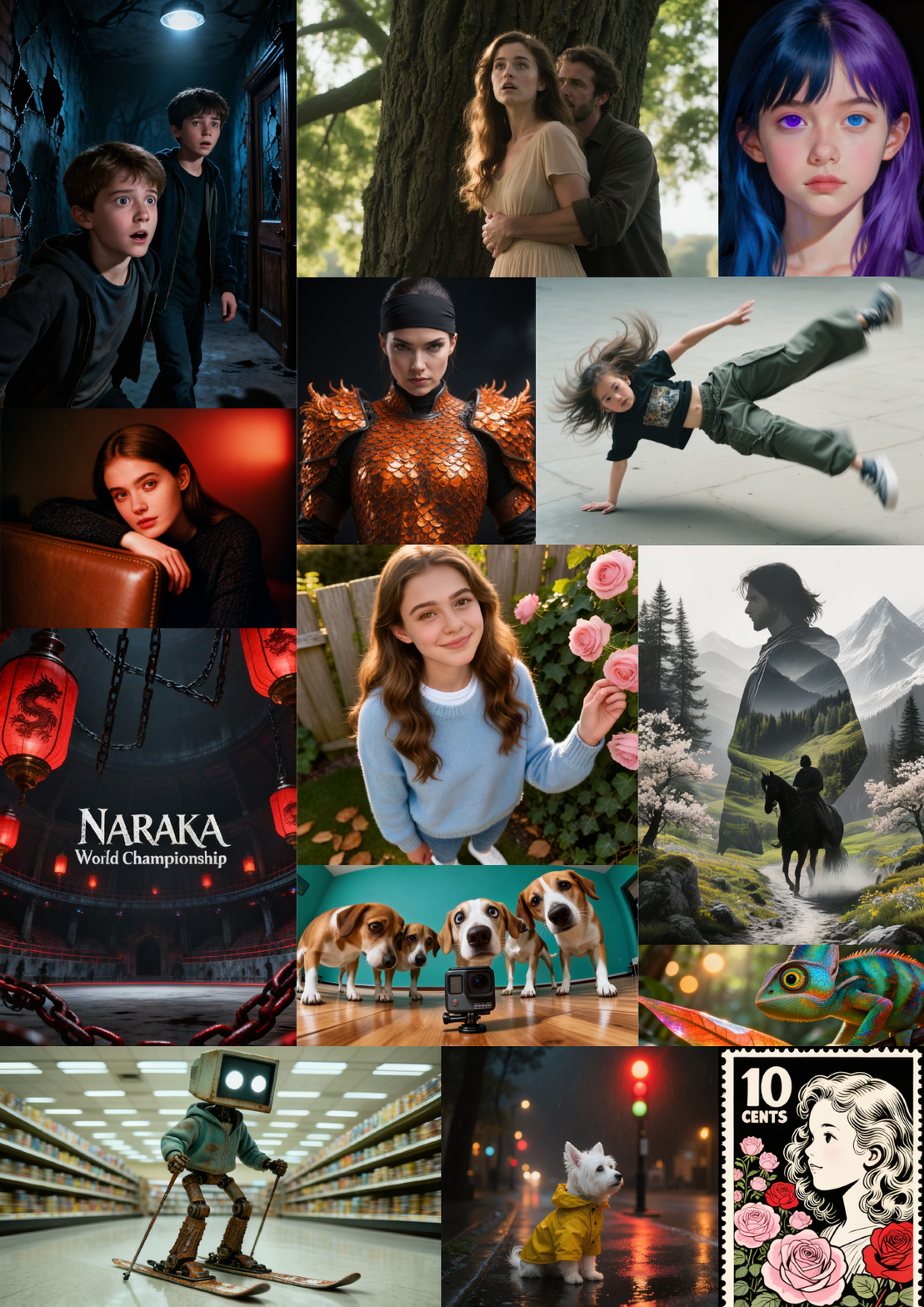}
\end{center}
\label{fig:teaser3}
\vspace{-1pt}
\caption{\textbf{\methodName{} text-to-image visualization.}}
\end{figure}

\begin{figure}[pt]
    \centering
    \makebox[\linewidth][c]{\includegraphics[width=1.1\linewidth]{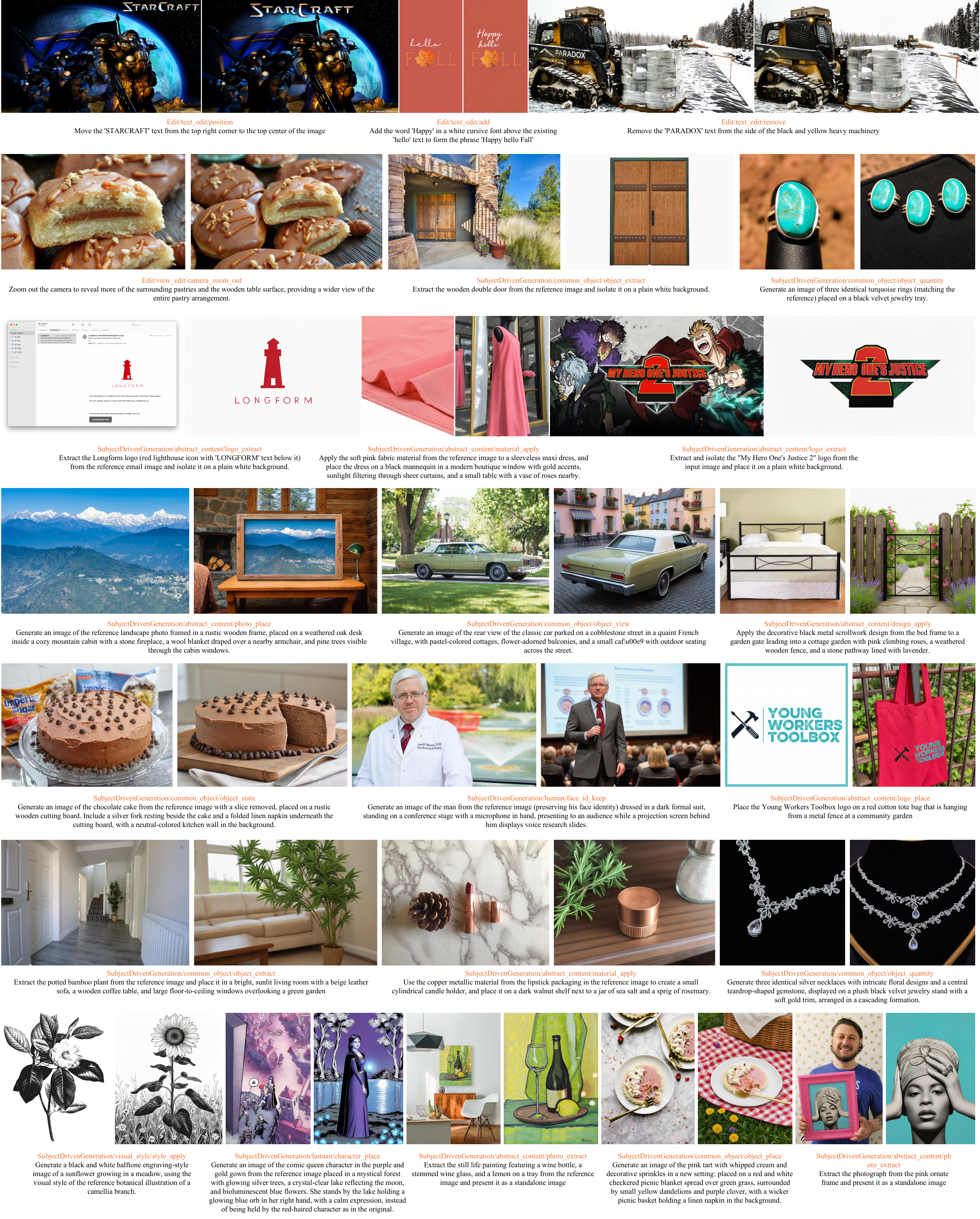}}
    \caption{\textbf{Edit results of \methodName{} on EditCanvas benchmark.}}
    \label{fig:edit_showcase2}
\end{figure}

\section{Model Performance}

\subsection{Image Generation}

We conduct a comprehensive evaluation of \methodName{} across multiple dimensions, including prompt following, world knowledge, and aesthetic quality. The quantitative results demonstrate that our autoregressive approach, particularly when enhanced with reinforcement learning, achieves state-of-the-art performance comparable to or exceeding top-tier models.

As shown in Tables \ref{tab:geneval} and \ref{tab:dpg}, \methodName{} demonstrates exceptional prompt following capabilities, with our RL-finetuned model achieving state-of-the-art scores (0.84 on GenEval \cite{ghosh2023geneval} and 88.32 on DPG \cite{dpg}) that outperform strong diffusion baselines like FLUX.1-dev \cite{flux} and match top-tier performers.

A critical advantage of our large-scale pre-training on diverse multimodal data is reflected in the WISE benchmark \cite{niu2025wise} (Table \ref{tab:wise_comparison}). \methodName{} RL achieves an overall score of \textbf{0.62}, matching the performance of Qwen-Image and significantly outperforming other autoregressive models like Show-o (0.30) and Janus-Pro-7B (0.35). The model demonstrates robust understanding across cultural, spatial, and physical domains, suggesting that our unified architecture effectively internalizes complex world knowledge.

On PRISM-Bench \cite{prismbench} (Table \ref{tab:prism_comparison}), which evaluates broader generative aspects including style and affect, \methodName{} RL achieves an overall score of \textbf{78.8}. This result places our model on par with top-tier systems like Seedream 3.0 \cite{seedream3} and Qwen-Image \cite{QwenImage}, demonstrating that our unified architecture achieves competitive aesthetic quality and text rendering capabilities alongside its strong instruction following.

\begin{table*}[ht]
\centering
\small
\setlength{\tabcolsep}{4pt}
\caption{\textbf{Comparison of text-to-image generation ability on GenEval \cite{ghosh2023geneval} benchmark.}
$\dagger$ refer to the methods using LLM rewriter. For AR Models, the best results are in \textbf{bold} and the second-best are \underline{underlined}.
}
\resizebox{0.7\linewidth}{!}{%
\begin{tabular}{lccccccc}
 \toprule
 \textbf{Model} & \textbf{Single Obj.} & \textbf{Two Obj.} & \textbf{Count.} & \textbf{Colors} & \textbf{Pos.} & \textbf{Color Att.} & \textbf{Overall$\uparrow$} \\
 \midrule
 \multicolumn{8}{l}{\textbf{Diffusion Models}} \\
 \addlinespace
 SDXL~\cite{sdxl} & 0.98 & 0.74 & 0.39 & 0.85 & 0.15 & 0.23 & 0.55 \\
 DALL-E $3$~\cite{dalle3} & 0.96 & 0.87 & 0.47 & 0.83 & 0.43 & 0.45 & 0.67 \\
 SD3-Medium~\cite{SD3} & 0.99 & 0.94 & 0.72 & 0.89 & 0.33 & 0.60 & 0.74 \\
 FLUX.1-dev$^{\dagger}$~\cite{flux} & 0.98 & 0.93 & 0.75 & 0.93 & 0.68 & 0.65 & 0.82 \\
 HiDream-I1-Full~\cite{cai2025hidream} & 1.00 & 0.98 & 0.79 & 0.91 & 0.60 & 0.72 & 0.83 \\
 GPT Image 1 [High]~\cite{gptimage} & 0.99 & 0.92 & 0.85 & 0.92 & 0.75 & 0.61 & 0.84 \\
 Seedream 3.0~\cite{gao2025seedream} & 0.99 & 0.96 & 0.91 & 0.93 & 0.47 & 0.80 & 0.84 \\
 Qwen-Image~\cite{QwenImage} & 0.99 & 0.92 & 0.89 & 0.88 & 0.76 & 0.77 & 0.87 \\
 \midrule
 \multicolumn{8}{l}{\textbf{AR + Diffusion Models}} \\
 \addlinespace
 Transfusion~\cite{transfusion} & - & - & - & - & - & - & 0.63 \\
 BAGEL$^{\dagger}$ \cite{bagel} & 0.98 & 0.95 & 0.84 & 0.95 & 0.78 &0.77 & 0.88 \\
 \midrule
 \multicolumn{8}{l}{\textbf{AR Models}} \\
 \addlinespace
 Chameleon~\cite{chameleon} & - & - & - & - & - & - & 0.39 \\
 Emu$3$-Gen ~\cite{emu3} & 0.98 & 0.71 & 0.34 & 0.81 & 0.17 & 0.21 & 0.54 \\
 TokenFlow-XL$^{\dagger}$~\cite{qu2025tokenflow} & 0.95 & 0.60 & 0.41 & 0.81 & 0.16 & 0.24 & 0.55 \\
 Show-o~\cite{show-o} & 0.98 & 0.80 & 0.66 & \underline{0.84} & 0.31 & 0.50 & 0.68 \\
 NextStep-1$^{\dagger}$ \cite{team2025nextstep} & - & - & - & - & - & - & 0.73 \\
 Janus-Pro-7B~\cite{januspro2025} & \underline{0.99} & \underline{0.89} & 0.59 & \textbf{0.90} & \textbf{0.79} & 0.66 & 0.80 \\
 Infinity-8B$^{\dagger}$ \cite{han2025infinity} & \underline{0.99} & \underline{0.89} & 0.59 & \textbf{0.90} & \textbf{0.79} & 0.66 & 0.80 \\
EMU3.5$^{\dagger}$ \cite{cui2025emu35} & - & - & - & - & - & - & \textbf{0.86} \\

 \rowcolor{mycolor_blue}
 \textbf{\methodName{}$^{\dagger}$} & 0.98 & \textbf{0.92} & \underline{0.73} & \textbf{0.90} & \underline{0.77} & \underline{0.69} & 0.83\\
 \rowcolor{mycolor_blue}
 \textbf{\methodName{}-RL$^{\dagger}$} & \textbf{1.00} & \textbf{0.92} & \textbf{0.75} & \textbf{0.90} & 0.76 & \textbf{0.70} & \underline{0.84}\\
 \bottomrule
\end{tabular}}
\label{tab:geneval}
\end{table*}

\begin{table*}[htbp]
    \centering
    \small
    \setlength{\tabcolsep}{4pt}
    \caption{\textbf{Comparison of text-to-image generation ability on WISE \cite{niu2025wise} benchmark.} WISE examines the complex semantic understanding and world knowledge for T2I generation. For AR Models, the best results are in \textbf{bold} and the second-best are \underline{underlined}.}
    \resizebox{0.8\linewidth}{!}{%
    \begin{tabular}{lccccccc}
        \toprule
        \textbf{Model}  & \textbf{Cultural} & \textbf{Time} & \textbf{Space} & \textbf{Bio.} & \textbf{Phys.} & \textbf{Chem.} & \textbf{Overall$\uparrow$} \\
        \midrule
        \multicolumn{8}{l}{\textbf{Diffusion Models}} \\
        \addlinespace
        SDXL \cite{sdxl} & 0.43 & 0.48 & 0.47 & 0.44 & 0.45 & 0.27 & 0.43 \\
        SD3-Medium \cite{SD3} & 0.42 & 0.44 & 0.48 & 0.39 & 0.47 & 0.29 & 0.42 \\
        FLUX.1-dev \cite{flux} & 0.48 & 0.58 & 0.62 & 0.42 & 0.51 & 0.35 & 0.50 \\
        GPT Image 1 [High] \cite{gptimage} & 0.81 & 0.71 & 0.89 & 0.83 & 0.79 & 0.74 & 0.80 \\
        Qwen-Image \cite{QwenImage} & \textbf{0.62} & \textbf{0.63} & \textbf{0.77} & \textbf{0.57} & \textbf{0.75} & \textbf{0.40} & \textbf{0.62} \\
        \midrule
        \multicolumn{8}{l}{\textbf{AR + Diffusion Models}} \\
        \addlinespace
        BAGEL \cite{bagel} & 0.44 & 0.55 & 0.68 & 0.44 & 0.60 & 0.39 & 0.52 \\
        \midrule
        \multicolumn{8}{l}{\textbf{AR Models}} \\
        \addlinespace
        Show-o \cite{show-o} & 0.28 & 0.36 & 0.40 & 0.23 & 0.33 & 0.22 & 0.30 \\
        Emu$3$-Gen \cite{emu3} & 0.34 & 0.45 & 0.48 & 0.41 & 0.45 & 0.27 & 0.39 \\
        Janus-Pro-7B \cite{januspro2025} & 0.30 & 0.37 & 0.49 & 0.36 & 0.42 & 0.26 & 0.35 \\
        Liquid \cite{wu2024liquid} & 0.38 & 0.42 & 0.53 & 0.36 & 0.47 & 0.30 & 0.41 \\
        \rowcolor{mycolor_blue}
        \textbf{\methodName{}} & \underline{0.62} & \underline{0.60} & \underline{0.70} & \underline{0.54} & \underline{0.58} & \underline{0.38} & \underline{0.59} \\
        \rowcolor{mycolor_blue}
        \textbf{\methodName{}-RL} & \textbf{0.63} & \textbf{0.63} & \textbf{0.77} & \textbf{0.58} & \textbf{0.67} & \textbf{0.39} & \textbf{0.62}\\
        \bottomrule
    \end{tabular}}
    \label{tab:wise_comparison}
\end{table*}

\begin{table*}[ht]
    \centering
    \small
    \setlength{\tabcolsep}{4pt}
    \caption{\textbf{Quantitative results on PRISM-Bench evaluated by GPT-4.1.}    PRISM-Bench \cite{prismbench} measures image generation performance on multiple aspects including imagination, entity accuracy, style, and long-text alignment. 
    Imag., Comp., and Text Rend. denote Imagination, Composition, and Text Rendering, respectively.
    For AR Models, the best result is in \textbf{bold} and the second best is \underline{underlined}.}
    \begin{tabular}{lcccccccc}
        \toprule
        \textbf{Model} & \textbf{Imag.} & \textbf{Entity} & \textbf{Text Rend.} & \textbf{Style} & \textbf{Affect.} & \textbf{Comp.} & \textbf{Long Text} & \textbf{Overall$\uparrow$} \\
        \midrule
        \multicolumn{9}{l}{\textbf{Diffusion Models}} \\
        \addlinespace
        SDXL~\cite{podell2023sdxl} & 58.2 & 70.0 & 25.4 & 73.9 & 78.0 & 75.4 & 41.9 & 60.4 \\
        SD3-Medium~\cite{SD3} & 63.3 & 60.6 & 43.0 & 75.2 & 79.5 & 82.3 & 53.8 & 65.4 \\
        SD3.5-Large~\cite{SD35} & 72.3 & 74.3 & 58.9 & 80.7 & 86.2 & 85.9 & 58.0 & 73.7 \\
        FLUX.1-dev~\cite{flux} & 71.1 & 71.0 & 56.3 & 76.4 & 89.7 & 86.8 & 64.6 & 73.7 \\
        HiDream-I1-Dev~\cite{hidream} & 69.0 & 69.5 & 58.8 & 73.7 & 83.7 & 83.7 & 52.8 & 70.2 \\
        HiDream-I1-Full~\cite{hidream} & 75.0 & 73.4 & 64.3 & 83.1 & 89.5 & 87.8 & 57.9 & 75.9 \\
        SEEDream 3.0~\cite{seedream3} & 76.9 & 77.0 & 63.2 & 85.7 & 89.8 & 89.8 & 75.0 & 79.6 \\
        Qwen-Image~\cite{QwenImage} & 79.6 & 76.3 & 61.6 & 86.6 & 90.4 & 90.3 & 74.5 & 79.9 \\

        \midrule
        \multicolumn{9}{l}{\textbf{AR + Diffusion Models}} \\
        \addlinespace
        Bagel~\cite{bagel} & 68.7 & 54.6 & 37.4 & 69.6 & 81.6 & 81.8 & 61.7 & 65.1 \\
        Bagel-CoT~\cite{bagel} & 71.3 & 61.2 & 31.7 & 67.3 & 83.8 & 83.2 & 58.0 & 65.2 \\
        \midrule
        \multicolumn{9}{l}{\textbf{AR Models}} \\
        \addlinespace
        JanusPro-7B~\cite{januspro} & 68.1 & 59.5 & 26.1 & 72.6 & 75.4 & \underline{72.4} & 51.1 & 60.7 \\
        \rowcolor{mycolor_blue}
        \textbf{\methodName{}} & \underline{84.3} & \underline{71.2} & \textbf{50.3} & \underline{80.2} & \underline{82.8} & 71.2 & \underline{70.8} & \underline{74.7} \\
        \rowcolor{mycolor_blue}
        \textbf{\methodName{}-RL}& \textbf{87.1} & \textbf{79.3} & \underline{49.8} & \textbf{83.8} & \textbf{88.8} & \textbf{90.5} & \textbf{72.4} & \textbf{78.8} \\
        \bottomrule
    \end{tabular}
    \label{tab:prism_comparison}
\end{table*}

\subsection{Image Editing}

We assess the image editing capabilities of \methodName{} across a diverse set of benchmarks, ranging from traditional instruction-based editing to complex subject-driven generation. As evidenced by the quantitative results, our unified autoregressive framework, particularly when enhanced with reinforcement learning, establishes new state-of-the-art performance levels.

On the {ImgEdit} benchmark \cite{ye2025imgedit}, which covers a wide spectrum of editing operations including addition, removal, and style transfer, \methodName{} RL achieves the highest overall score of \textbf{4.49}, surpassing strong baselines such as Qwen-Image (4.27) and Emu3.5 (4.41). Notably, our model demonstrates exceptional precision in the \textit{Adjust} (4.68) and \textit{Remove} (4.67) sub-tasks, indicating a robust ability to manipulate specific image regions without disrupting the global context.
OmniContext benchmark \cite{wu2025omnigen2} evaluate the in-context editing and subject preservation ability. In the single-subject setting (Table \ref{tab:omni_context_single}), \methodName{} RL achieves a Subject Consistency (SC) score of 9.22, significantly outperforming models like OmniGen2 (8.34) and even surpassing the proprietary GPT-4o (9.03) in maintaining subject fidelity. This result highlights the effectiveness of our interleaved pre-training in learning robust identity representations.
On GEdit-Bench \cite{liu2025step1x} (Table \ref{tab:gedit}), which explicitly measures the trade-off between semantic consistency and perceptual quality, \methodName{} RL achieves the best overall score 7.87.

\textbf{Comprehensive Evaluation on EditCanvas.} Finally, we report results on our proposed \textbf{EditCanvas} benchmark (Table \ref{tab:main_results}). To address the limitations of existing datasets that often focus on isolated capabilities or lack granularity, EditCanvas establishes both \textit{Traditional Editing} and \textit{Subject-Driven Generation} across 56 fine-grained tasks with over 5,000 high-quality samples (please refer to Appendix \ref{sec:editcanvas} for detailed dataset statistics and comparisons). 
On this rigorous benchmark, \methodName{} RL demonstrates balanced excellence, achieving an overall score of \textbf{8.04}. It shows particular strength in \textit{Subject-Driven Generation} (8.78). These results confirm that our "next-scale prediction" paradigm allows for precise local modifications while maintaining the high aesthetic quality required for creative workflows, fully leveraging the comprehensive evaluation scope provided by EditCanvas.

\begin{table*}[ht]
  \centering
  \small
  \caption{\textbf{Quantitative comparison results on ImgEdit~\citep{ye2025imgedit}.} All metrics are evaluated by GPT-4.1. ``Overall'' is calculated by averaging all scores across tasks.}
  \resizebox{\linewidth}{!}{
  \begin{tabular}{lcccccccccc}
    \toprule
    \textbf{Model} & \textbf{Add} & \textbf{Adjust} & \textbf{Extract} & \textbf{Replace} & \textbf{Remove} & \textbf{Background} & \textbf{Style} & \textbf{Hybrid} & \textbf{Action} & \textbf{Overall$\uparrow$} \\
    \midrule
    MagicBrush~\citep{zhang2023magicbrush} & 2.84 & 1.58 & 1.51 & 1.97 & 1.58 & 1.75 & 2.38 & 1.62 & 1.22 & 1.90 \\
    Instruct-Pix2Pix~\citep{brooks2023instructpix2pix} & 2.45 & 1.83 & 1.44 & 2.01 & 1.50 & 1.44 & 3.55 & 1.20 & 1.46 & 1.88 \\
    AnyEdit~\citep{yu2025anyedit} & 3.18 & 2.95 & 1.88 & 2.47 & 2.23 & 2.24 & 2.85 & 1.56 & 2.65 & 2.45 \\
    UltraEdit~\citep{zhao2024ultraedit} & 3.44 & 2.81 & 2.13 & 2.96 & 1.45 & 2.83 & 3.76 & 1.91 & 2.98 & 2.70 \\
    OmniGen~\citep{xiao2025omnigen} & 3.47 & 3.04 & 1.71 & 2.94 & 2.43 & 3.21 & 4.19 & 2.24 & 3.38 & 2.96 \\
    ICEdit~\citep{zhang2025context} & 3.58 & 3.39 & 1.73 & 3.15 & 2.93 & 3.08 & 3.84 & 2.04 & 3.68 & 3.05 \\
    Step1X-Edit~\citep{liu2025step1x} & 3.88 & 3.14 & 1.76 & 3.40 & 2.41 & 3.16 & 4.63 & 2.64 & 2.52 & 3.06 \\
    BAGEL~\citep{bagel} & 3.56 & 3.31 & 1.70 & 3.30 & 2.62 & 3.24 & 4.49 & 2.38 & 4.17 & 3.20 \\
    UniWorld-V1~\citep{lin2025uniworldv1} & 3.82 & 3.64 & 2.27 & 3.47 & 3.24 & 2.99 & 4.21 & 2.96 & 2.74 & 3.26 \\
    OmniGen2~\citep{wu2025omnigen2} & 3.57 & 3.06 & 1.77 & 3.74 & 3.20 & 3.57 & \underline{4.81} & 2.52 & 4.68 & 3.44 \\
    FLUX.1 Kontext [Pro]~~\citep{labs2025kontext} & 4.25 & 4.15 & 2.35 & 4.56 & 3.57 & 4.26 & 4.57 & 3.68 & 4.63 & 4.00 \\
    GPT Image 1 [High]~~\citep{gptimage} & \textbf{4.61} & 4.33 & 2.90 & 4.35 & 3.66 & \textbf{4.57} & \textbf{4.93} & 3.96 & \textbf{4.89} & 4.20 \\
    Qwen-Image~\citep{QwenImage} & 4.38 & 4.16 & 3.43 & \underline{4.66} & 4.14 & \underline{4.38} & \underline{4.81} & 3.82 & 4.69 & 4.27 \\
    Emu3.5~\citep{cui2025emu35} & \textbf{4.61} & 4.32 & 3.96 & \textbf{4.84} & \underline{4.58} & 4.35 & 4.79 & 3.69 & 4.57 & 4.41 \\
    
    \rowcolor{mycolor_blue}
    \textbf{\methodName{}} & 4.34 & \textbf{4.68} & \textbf{4.23} & 4.47 & \underline{4.58} & 4.30 & 4.61 & \textbf{4.20} & 4.56 & \underline{4.44} \\
    \rowcolor{mycolor_blue}
    \textbf{\methodName{}-RL} & 4.33 & \textbf{4.68} & \underline{4.22} & 4.50 & \textbf{4.67} & \underline{4.38} & 4.74 & \underline{4.11} & \underline{4.77} & \textbf{4.49} \\
    \bottomrule
  \end{tabular}
  }
  \label{tab:imgedit}
\end{table*}

\begin{table*}[h]
    \centering
    \small
    \caption{\textbf{Comparison on task type SINGLE from OmniContext~\cite{wu2025omnigen2}.} Prompt Following (PF), Subject Consistency (SC), and Overall scores are reported (higher is better, $\uparrow$). }
    \resizebox{\linewidth}{!}{
    \begin{tabular}{lccccccccc}
        \toprule
        \multirow{3}{*}{\textbf{Model}} & \multicolumn{9}{c}{\textbf{SINGLE$\uparrow$}}\\
        \cmidrule(lr){2-10}
        & \multicolumn{3}{c}{\textbf{Character}} & \multicolumn{3}{c}{\textbf{Object}} & \multicolumn{3}{c}{\textbf{Average}} \\
        \cmidrule(lr){2-4}
        \cmidrule(lr){5-7}
        \cmidrule(lr){8-10}
        & \textbf{PF} & \textbf{SC} & \textbf{Overall} & \textbf{PF} & \textbf{SC} & \textbf{Overall} & \textbf{PF} & \textbf{SC} & \textbf{Overall} \\
        \midrule
        Flux.1 Kontext max~\cite{labs2025kontext} & 7.98 & \textbf{9.24} & 8.48 & 8.78 & \underline{8.76} & 8.68 & 8.38 & \textbf{9.00} & 8.58 \\
        Gemini-2.0-flash~\cite{gemini2flash} & 5.54 & 5.98 & 5.06 & 6.17 & 5.89 & 5.17 & 5.86 & 5.93 & 5.11 \\
        GPT-4o~\cite{gpt4o} & \textbf{8.89} & 9.03 & \textbf{8.90} & \textbf{9.40} & 8.74 & \textbf{9.01} & \textbf{9.14} & 8.88 & \textbf{8.95} \\
        \midrule
        InfiniteYou~\cite{jiang2025infiniteyou} & 7.81 & 5.15 & 6.05 & - & - & - & - & - & - \\
        UNO~\cite{uno} & 7.56 & 6.48 & 6.60 & 7.78 & 6.65 & 6.83 & 7.67 & 6.56 & 6.72 \\
        BAGEL~\cite{bagel} & 7.72 & 4.86 & 5.48 & 8.56 & 6.06 & 7.03 & 8.14 & 5.46 & 6.25 \\
        OmniGen~\cite{xiao2025omnigen} & 7.12 & 7.58 & 7.21 & 7.66 & 5.04 & 5.71 & 7.39 & 6.31 & 6.46 \\
        OmniGen2~\cite{wu2025omnigen2} & \underline{8.04} & 8.34 & 8.05 & 8.44 & 7.26 & 7.58 & 8.24 & 7.80 & 7.81 \\
        \rowcolor{mycolor_blue}
        \textbf{\methodName{}} & 7.66 & 9.04 & 8.23 & 8.96 & \textbf{8.94} & \underline{8.89} & 8.31 & \underline{8.99} & 8.56 \\
        \rowcolor{mycolor_blue}
        \textbf{\methodName{}-RL} & 7.96 & \underline{9.22} & \underline{8.50} & \underline{9.00} & \underline{8.76} & 8.84 & \underline{8.48} & \underline{8.99} & \underline{8.67} \\
        \bottomrule
    \end{tabular}
    }
    \label{tab:omni_context_single}
    \vspace{-1em}
\end{table*}

\begin{table*}[h]
    \centering
    \small
    \caption{\textbf{Comparison on GEdit-Bench~\cite{liu2025step1x}.} G\_SC: Semantic Consistency, G\_PQ: Perceptual Quality, G\_O: Overall Score (geometric mean of G\_SC and G\_PQ). All metrics evaluated by GPT-4.1.}
    \resizebox{0.4\linewidth}{!}{
    \begin{tabular}{lccc}
        \toprule
        \textbf{Model} & \textbf{G\_SC} & \textbf{G\_PQ} & \textbf{G\_O$\uparrow$} \\
        \midrule
        Instruct-Pix2Pix~\citep{brooks2023instructpix2pix} & 3.58 & 5.49 & 3.68 \\
        AnyEdit~\citep{yu2025anyedit} & 3.18 & 5.82 & 3.21 \\
        MagicBrush~\citep{zhang2023magicbrush} & 4.68 & 5.66 & 4.52 \\
        UniWorld-v1~\citep{lin2025uniworldv1} & 4.93 & 7.43 & 4.85 \\
        OmniGen~\citep{xiao2025omnigen} & 5.96 & 5.89 & 5.06 \\
        OmniGen2~\citep{wu2025omnigen2} & 7.16 & 6.77 & 6.41 \\
        Gemini 2.0~\citep{googleGemini2} & 6.73 & 6.61 & 6.32 \\
        BAGEL~\citep{bagel} & 7.36 & 6.83 & 6.52 \\
        FLUX.1 Kontext [Pro]~\citep{labs2025kontext} & 7.02 & 7.60 & 6.56 \\
        Step1X-Edit~\citep{liu2025step1x} & 7.66 & 7.35 & 6.97 \\
        GPT Image 1 \citep{gptimage} & {7.85} & {7.62} & {7.53} \\
        Qwen-Image~\citep{QwenImage} & 8.00 & 7.86 & 7.56 \\
        EMU3.5 \cite{cui2025emu35} & 8.11 & 7.70 & 7.59 \\
        \rowcolor{mycolor_blue}
        \textbf{\methodName{}} & \underline{8.16} & \underline{7.90} & \underline{7.60} \\
        \rowcolor{mycolor_blue}
        \textbf{\methodName{}-RL} & \textbf{8.37} & \textbf{8.10} & \textbf{7.87} \\
        \bottomrule
    \end{tabular}
    }
    \label{tab:gedit}
\end{table*}

\begin{table*}[h]
\centering
\caption{\textbf{Comparison on our proposed EditCanvas benchmark.} We report the overall score for each sub-category. Abbreviations for subject-driven sub-categories are: Abs. (abstract content), Obj. (common object), Hum. (human), and Sty. (visual style). For the aggregate metrics, we report Prompt Following (PF), Perceptual Quality (PQ), Subject Consistency (SC), and the Overall score. The best results are in \textbf{bold} and the second-best are \underline{underlined}.}
\label{tab:main_results}
\resizebox{\textwidth}{!}{
\begin{tabular}{l|ccccc|ccc|cccc|ccc|c} 
\toprule
\multirow{3}{*}{\textbf{Method}} & \multicolumn{8}{c|}{\textbf{Edit Task}} & \multicolumn{7}{c|}{\textbf{Subject-Driven Generation}} & \multirow{3}{*}{\textbf{Overall}} \\ 
\cmidrule(lr){2-9} \cmidrule(lr){10-16}
 & \multicolumn{5}{c|}{Sub-Categories} & \multicolumn{3}{c|}{Aggregate} & \multicolumn{4}{c|}{Sub-Categories} & \multicolumn{3}{c|}{Aggregate} & \\ 
\cmidrule(lr){2-6} \cmidrule(lr){7-9} \cmidrule(lr){10-13} \cmidrule(lr){14-16}
 & Global & Local & Style & Text & View & PF & PQ & \textbf{Ov.} & Abs. & Obj. & Hum. & Sty. & PF & SC & \textbf{Ov.} & \\ 
\midrule
Bagel \cite{bagel} & 5.72 & 5.50 & 5.08 & 3.79 & 6.89 & 5.66 & 7.27 & 5.28 & 6.91 & 7.92 & 8.43 & 7.86 & 8.16 & 7.50 & 7.69 & 6.49 \\
HiDream \cite{hidream} & 5.30 & 5.25 & 5.80 & 3.12 & 4.58 & 5.21 & 5.86 & 4.81 & 5.53 & 6.30 & 8.04 & 7.32 & 6.94 & 6.38 & 6.52 & 5.67 \\
OmniGen2 \cite{wu2025omnigen2} & 5.04 & 4.99 & 5.17 & 3.39 & 4.50 & 4.91 & 6.78 & 4.65 & 5.25 & 6.55 & 8.37 & 7.09 & 7.14 & 6.37 & 6.58 & 5.62 \\
Qwen-Image \cite{QwenImage} & 7.32 & 6.78 & 7.56 & 5.86 & 8.19 & 7.34 & 7.83 & 6.83 & 7.90 & 8.45 & 8.92 & 8.49 & 8.83 & 8.12 & 8.38 & 7.60 \\
Flux.1 Kontext [dev] \cite{labs2025flux1kontextflowmatching} & 6.19 & 5.35 & 6.47 & 5.32 & 5.92 & 6.08 & 7.64 & 5.59 & 7.48 & 8.06 & 8.51 & 7.89 & 8.32 & 7.79 & 7.96 & 6.77 \\
Nano-Banana & 6.69 & 6.49 & 5.30 & 5.62 & 7.18 & 6.83 & \underline{8.05} & 6.34 & \underline{9.22} & 9.13 & \textbf{9.21} & \underline{9.05} & \underline{9.51} & \underline{8.91} & \underline{9.17} & 7.76 \\
GPT-Image-1 & \textbf{8.37} & \textbf{7.96} & \textbf{8.09} & \textbf{7.59} & \underline{8.75} & \textbf{8.35} & \textbf{8.35} & \textbf{8.02} & \textbf{9.41} & \textbf{9.31} & \textbf{9.21} & \textbf{9.10} & \textbf{9.72} & \textbf{8.96} & \textbf{9.31} & \textbf{8.67} \\
Emu3.5 \cite{cui2025emu35} & \underline{8.06} & \underline{7.61} & 7.71 & \underline{7.17} & \textbf{8.92} & \underline{8.34} & 7.80 & \underline{7.70} & 8.86 & \underline{9.16} & 9.18 & 8.70 & 9.50 & 8.68 & 9.04 & \underline{8.37} \\
\midrule
\rowcolor{mycolor_blue}
\textbf{\methodName{}} & 7.82 & 7.10 & 7.45 & 6.17 & 8.01 & 7.91 & 7.26 & 7.13 & 8.69 & 8.73 & 8.82 & 8.58 & 9.09 & 8.46 & 8.73 & 7.93 \\
\rowcolor{mycolor_blue}
\textbf{\methodName{}-RL} & 7.92 & 7.28 & \underline{7.79} & 6.41 & 8.16 & 8.11 & 7.26 & 7.31 & 8.72 & 8.77 & 8.91 & 8.54 & 9.17 & 8.47 & 8.78 & 8.04 \\
\bottomrule
\end{tabular}%
}
\end{table*}

\begin{table}[htbp]
  \centering
  \caption{\textbf{Tokenizer reconstruction metrics on ImageNet-1K validation set \cite{russakovsky2015imagenet} and in-house reconstruction benchmark.}}
  \label{tab:image_recon_comparison}
  \resizebox{0.6\linewidth}{!}{ 
    \begin{tabular}{llccc}
      \toprule
      \textbf{Benchmark} & \textbf{Method} & \textbf{Resolution} & \textbf{PSNR} $\uparrow$ & \textbf{SSIM} $\uparrow$ \\
      \midrule
      \multirow{2}{*}{ImageNet} 
        & Original TokenFlow \cite{qu2025tokenflow} & 512  & 23.147 & 0.761 \\
      & \methodName{} & 512  & 25.228 & 0.820 \\
      & \methodName{} & 1024 & 27.410 & 0.884 \\
      \midrule
      \multirow{3}{*}{In-house benchmark} 
      & Original TokenFlow \cite{qu2025tokenflow} & 512  & 23.320 & 0.760 \\
      & \methodName{} & 512  & 26.472 & 0.870 \\
      & \methodName{} & 1024 & \textbf{28.038} & \textbf{0.900} \\
      \bottomrule
    \end{tabular}
  }
\end{table}

\subsection{Interleaved Generation}

\begin{figure}[htp]
    \begin{center}
    \includegraphics[width=1.0\linewidth]{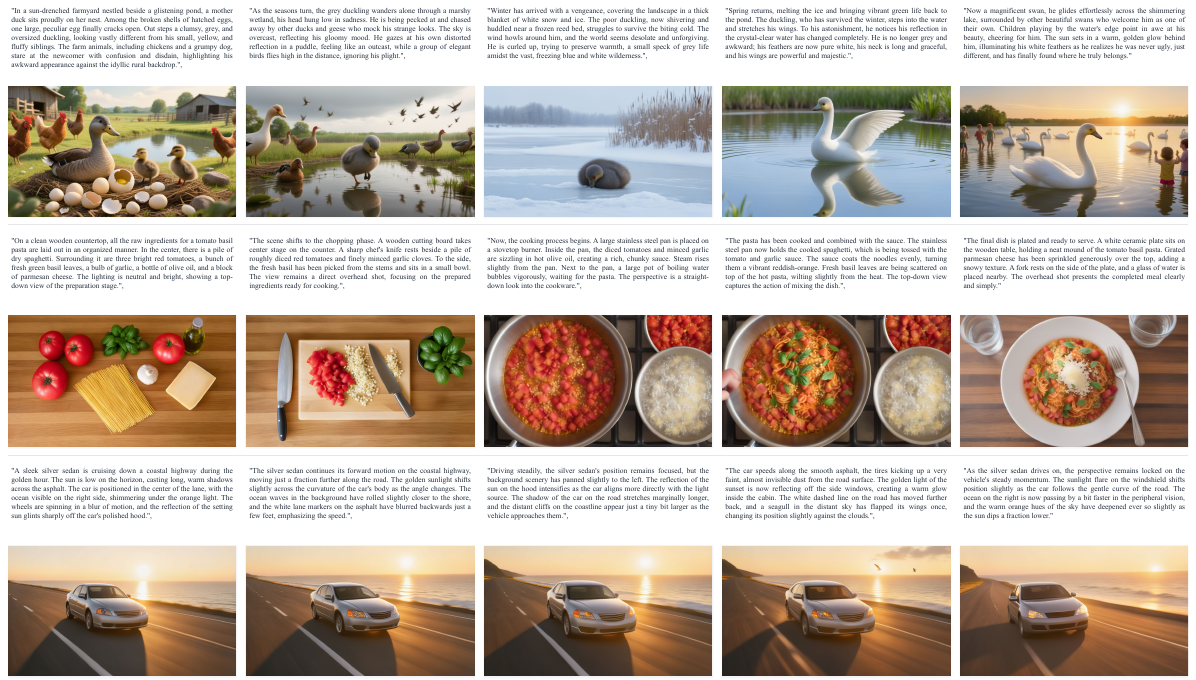}
    \end{center}
    \caption{\textbf{\methodName{} interleaved generation results.} Our unified sequential modeling framework can produce text descriptions and corresponding images in an alternating sequence. We show diverse applications including storytelling (top), recipe instructions (middle), and dynamic scene generation (bottom).}
    \label{fig:interleaved}
\end{figure}

To evaluate the unified modeling capabilities of \methodName{}, we test its performance on interleaved multimodal generation tasks. Unlike pipeline approaches that rely on separate text and image models, our unified autoregressive framework treats text and visual tokens as a single continuous sequence. Figure \ref{fig:interleaved} demonstrates the model's ability to generate coherent, alternating sequences of text and images across distinct domains.

\subsection{CoT Reasoning}

\begin{figure}[htp]
    \begin{center}
    \includegraphics[width=0.8\linewidth]{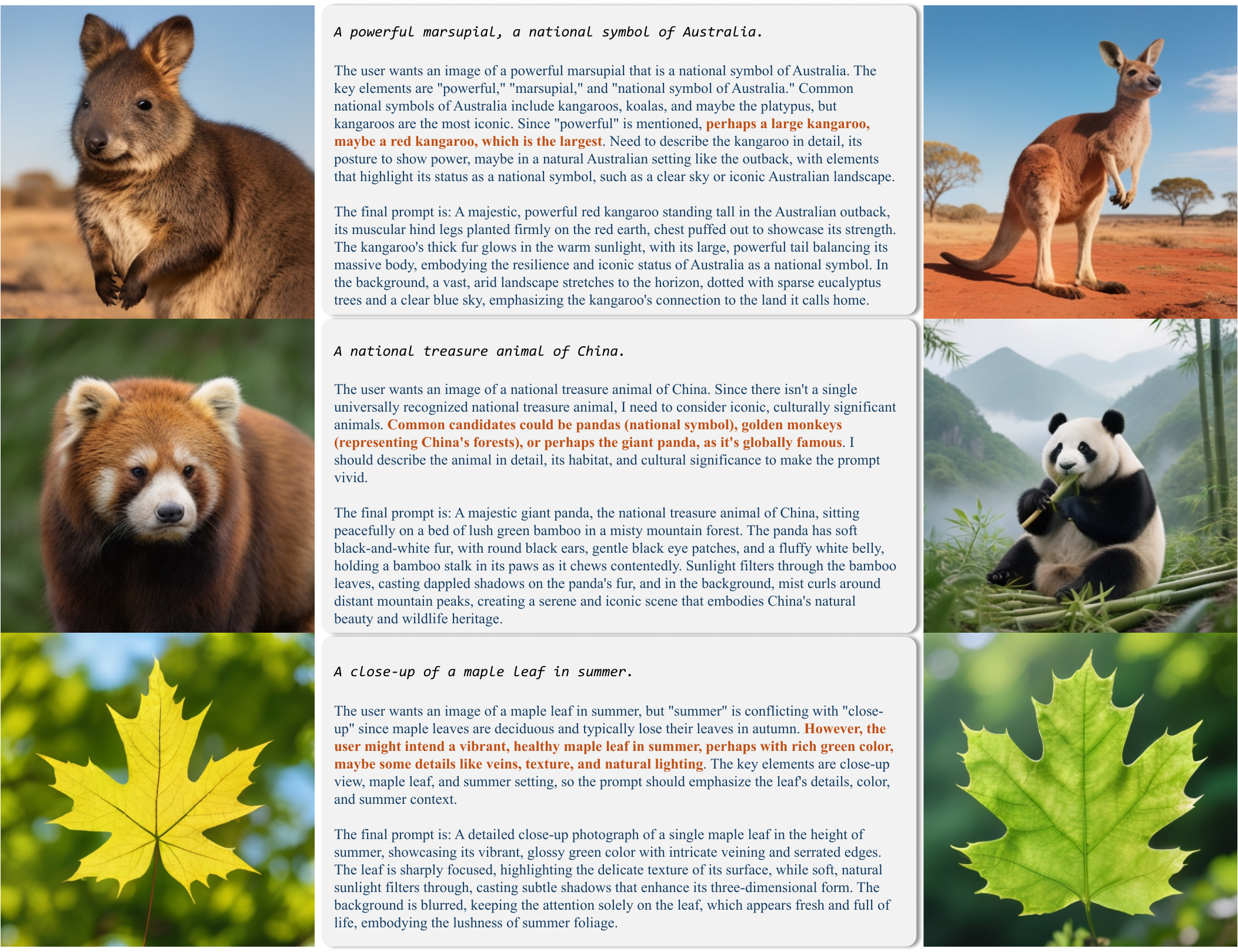}
    \end{center}
    \caption{\textbf{Qualitative comparison of text-to-image generation with and without reasoning.} Left: Baseline outputs generated directly from prompts, often failing to capture implicit constraints or cultural context (e.g., generating a Red Panda instead of a Giant Panda for "China's national treasure"). Middle: The input prompt coupled with the generated reasoning traces. Right: Final outputs leveraging the thinking process.}
    \label{fig:t2i_thinking}
\end{figure}

A distinct advantage of our \methodName{} architecture is its native support for interleaved text-image generation, which allows for the seamless integration of Chain-of-Thought (CoT) reasoning prior to visual synthesis. To investigate the potential of this paradigm in handling implicit constraints and cultural nuances, we conducted an exploratory study using a curated dataset of 1M instruction-reasoning-image triplets.

As illustrated in Figure \ref{fig:t2i_thinking}, standard generation often struggles with prompts containing latent ambiguities. For instance, baseline models may render a Red Panda for "China's national treasure" or generate autumnal red leaves for a "maple leaf in summer," failing to resolve logical conflicts. By fine-tuning \methodName{} to articulate a reasoning trace—analyzing the prompt for cultural context and physical constraints—before predicting visual tokens, the model effectively self-corrects these semantic discrepancies.

We evaluated this on WISE \cite{niu2025wise}. This experiment is conducted as a controlled study on an intermediate model checkpoint. The integration of CoT yielded a substantial performance boost, raising the WISE score from 0.60 to 0.70. This significant relative improvement demonstrates that enabling the model to "think" before "drawing" creates a strong inductive bias for logical consistency and superior prompt adherence.

\subsection{In-context Learning}

\begin{figure}[htp]
    \begin{center}
    \includegraphics[width=0.8\linewidth]{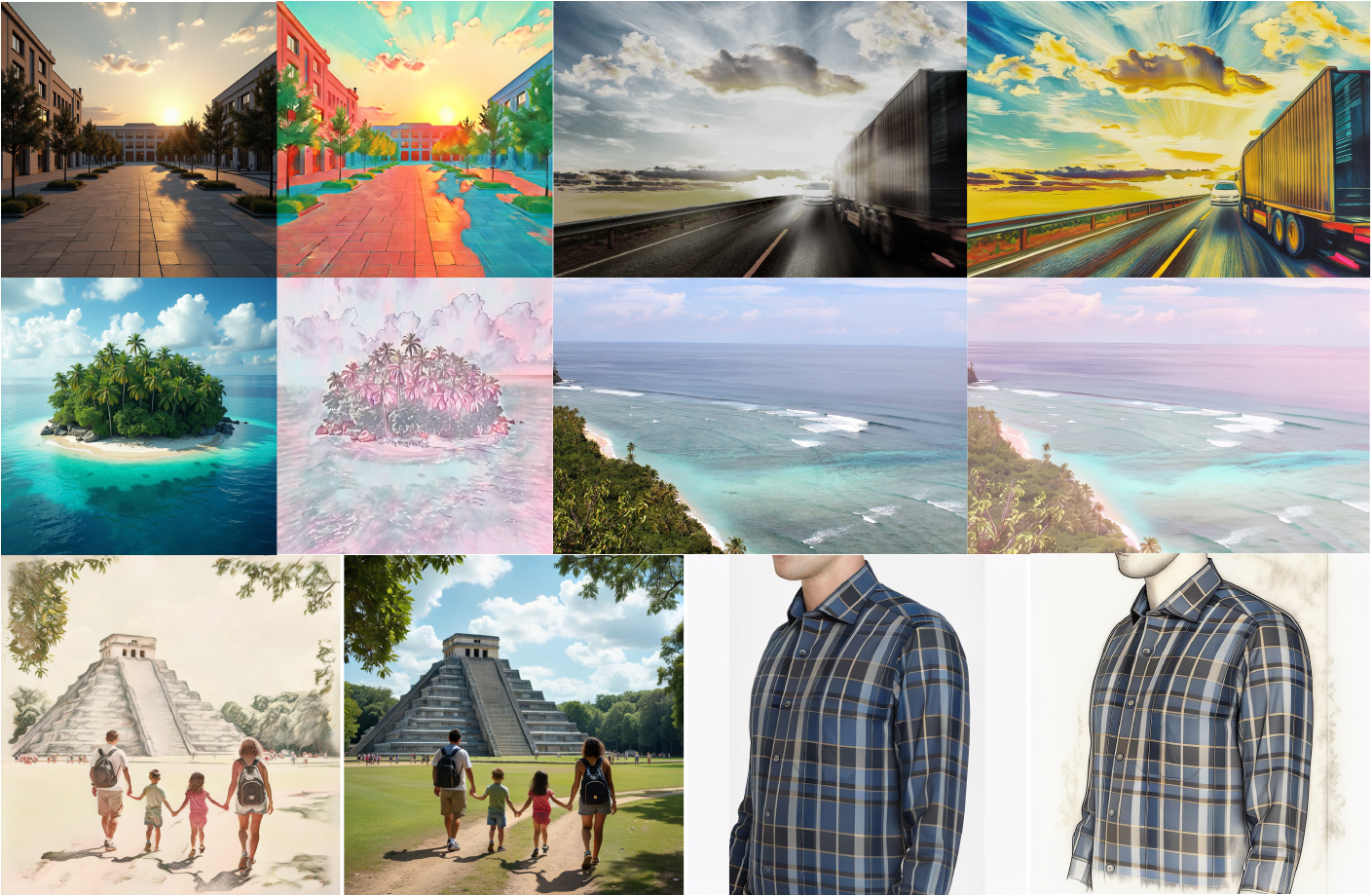}
    \end{center}
    \caption{\textbf{Visualization of in-context learning.} The model can learn the transformation pattern in the first two example pairs and applies a similar modification to the input image.}
    \label{fig:exp_incontext}
\end{figure}

Benefiting from large-scale pre-training on interleaved image-text data, \methodName{} demonstrates the capability to perform in-context learning. As shown in Fig. \ref{fig:exp_incontext}, when prompted with instructions to "infer the transformation pattern from the preceding examples and apply it to the target image," the model is able to capture the stylistic mapping and modifies the input accordingly.

\subsection{Image Reconstruction}

\textbf{Quantitative Evaluation.} As detailed in Table \ref{tab:image_recon_comparison}, our optimized training recipe yields substantial gains over the baseline TokenFlow architecture. On ImageNet-1K ($512^2$), \methodName{} achieves a marked improvement of $+2.08$ dB in PSNR, validating the efficacy of our multi-stage training and scale dropout strategies. Scaling to $1024^2$ further enhances fidelity, reaching a PSNR of 28.04 on our internal benchmark, demonstrating the tokenizer's robustness in high-resolution scenarios.

\begin{figure*}[ht]
    \centering
    \includegraphics[width=1.0\linewidth]{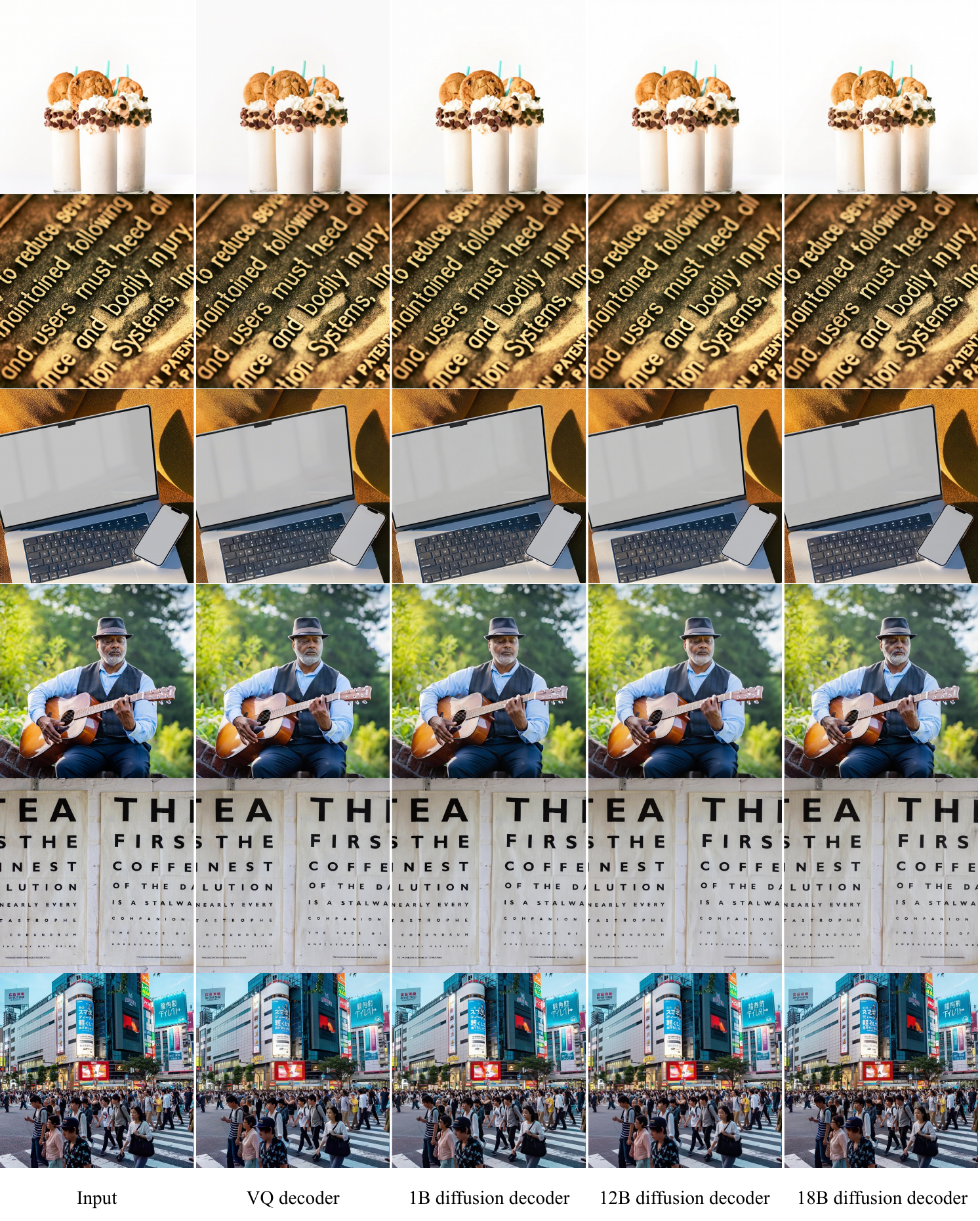}
    \caption{\textbf{Qualitative comparison of VQ decoder versus diffusion decoders on image reconstruction.} The VQ decoder yields satisfactory reconstruction results, although its performance on fine details is moderate. Serving as an optional enhancement, the diffusion decoder improves the reconstruction quality on local details, such as small text and small faces, as its size increases.}
    \label{fig:decoder_comparison_recon}
\end{figure*}

\textbf{Qualitative Comparison.} Figure \ref{fig:decoder_comparison_recon} presents the visual comparison. The default VQ decoder demonstrates robust reconstruction capabilities, faithfully preserving the structural layout and semantic integrity of the input. The optional diffusion decoder acts as a perceptual enhancer, further refining high-frequency textures in complex regions, such as small faces and texts. However, we observe that this generative refinement involves a trade-off: while it boosts perceptual realism, the VQ decoder ensures higher fidelity to the original signal, avoiding the potential detail alteration inherent in diffusion-based re-synthesis.

\subsection{Multimodal Understanding}

We observe that simultaneously supporting multimodal understanding and image generation within a dense 7B decoder-only model imposes a significant capacity bottleneck. Furthermore, due to the scarcity of our high-quality pre-training data for multimodal understanding, we minimize the understanding training in the late pre-training stage. We only employed a task inversion where text-to-image samples were converted into image captioning samples with a probability of 10\%. 

To evaluate the understanding capabilities of our base model, we conducted SFT experiments with increasing data scales: (1) \textbf{0.7M} standard LLaVA-1.5 SFT data~\cite{liu2024improved}; (2) \textbf{21M} LLaVA-OneVision SFT data; and (3) \textbf{40M} composite data, consisting of a 19M high-quality captioning mid-train followed by the 21M SFT. 
As shown in Table~\ref{tab:multimodal_results}, despite the reduced model size, our 7B model fine-tuned on just 0.7M data delivers performance comparable to the 13B LLaVA-1.5 baseline. While scaling to 21M significantly boosts document-oriented tasks (e.g., ChartQA, OcrBench), the integration of the 19M high-quality captioning data (40M total) yields the most robust improvements across all benchmarks, demonstrating the potential of our base model.

\begin{table*}[ht]
\centering
\small
\caption{\textbf{Quantitative comparison results on multimodal understanding benchmarks.} We compare different training data configurations.}
\resizebox{\linewidth}{!}{
\begin{tabular}{lcccccccccc}
\toprule
\textbf{Model} & \textbf{\#Training Data} & \textbf{Model Size} & \textbf{MMStar} & \textbf{ChartQA} & \textbf{OCRBench} & \textbf{MME} & \textbf{MME-P} & \textbf{MMB} & \textbf{MMMU} & \textbf{TextVQA} \\
\midrule
LLaVA 1.5 \cite{liu2024improved} & 0.7M & 13B & 36.7 & 20.4 & 36.2 & 1826.7 & 1531.3 & 67.7 & 36.4 & 61.3 \\
\midrule
\multirow{3}{*}{\textbf{\methodName{}}} & 0.7M & 7B & 44.0 & 18.7 & 35.6 & 1752.1 & 1462.1 & 65.3 & 35.1 & 49.5 \\
 & 21M & 7B & 42.8 & 50.4 & 44.8 & 1744.9 & 1386.3 & 61.7 & 35.8 & 52.7 \\
 & 40M & 7B & 53.0 & 57.7 & 55.1 & 1897.7 & 1453.1 & 66.7 & 37.1 & 58.9 \\
\bottomrule
\end{tabular}
}
\label{tab:multimodal_results}
\end{table*}


\begin{figure*}[ht]
    \centering
    \includegraphics[width=0.8\linewidth]{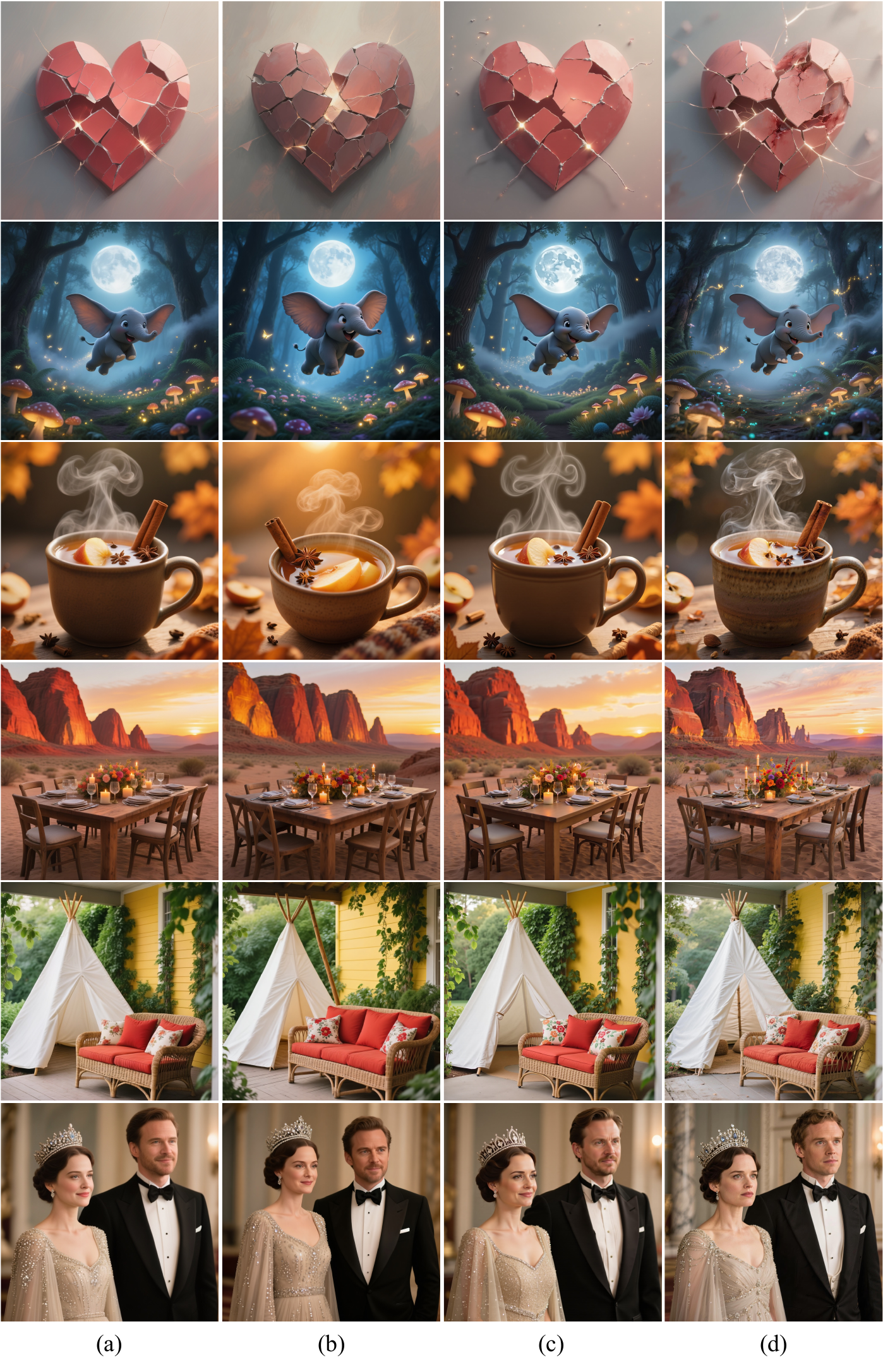}
    \caption{\textbf{Qualitative comparison of VQ decoder versus diffusion decoders on text-to-image scenario.} (a) shows the reconstruction from the baseline VQ decoder. (b)-(d) display results from our diffusion decoders with varying capacities: (b) the 1B parameter UNet-based model, (c) the 12B parameter Transformer-based model, and (d) the 18B parameter Transformer-based model. Note the progression in detail fidelity as model size increases.}
    \label{fig:decoder_comparison}
\end{figure*}

\begin{figure*}[ht]
    \centering
    \includegraphics[width=1.0\linewidth]{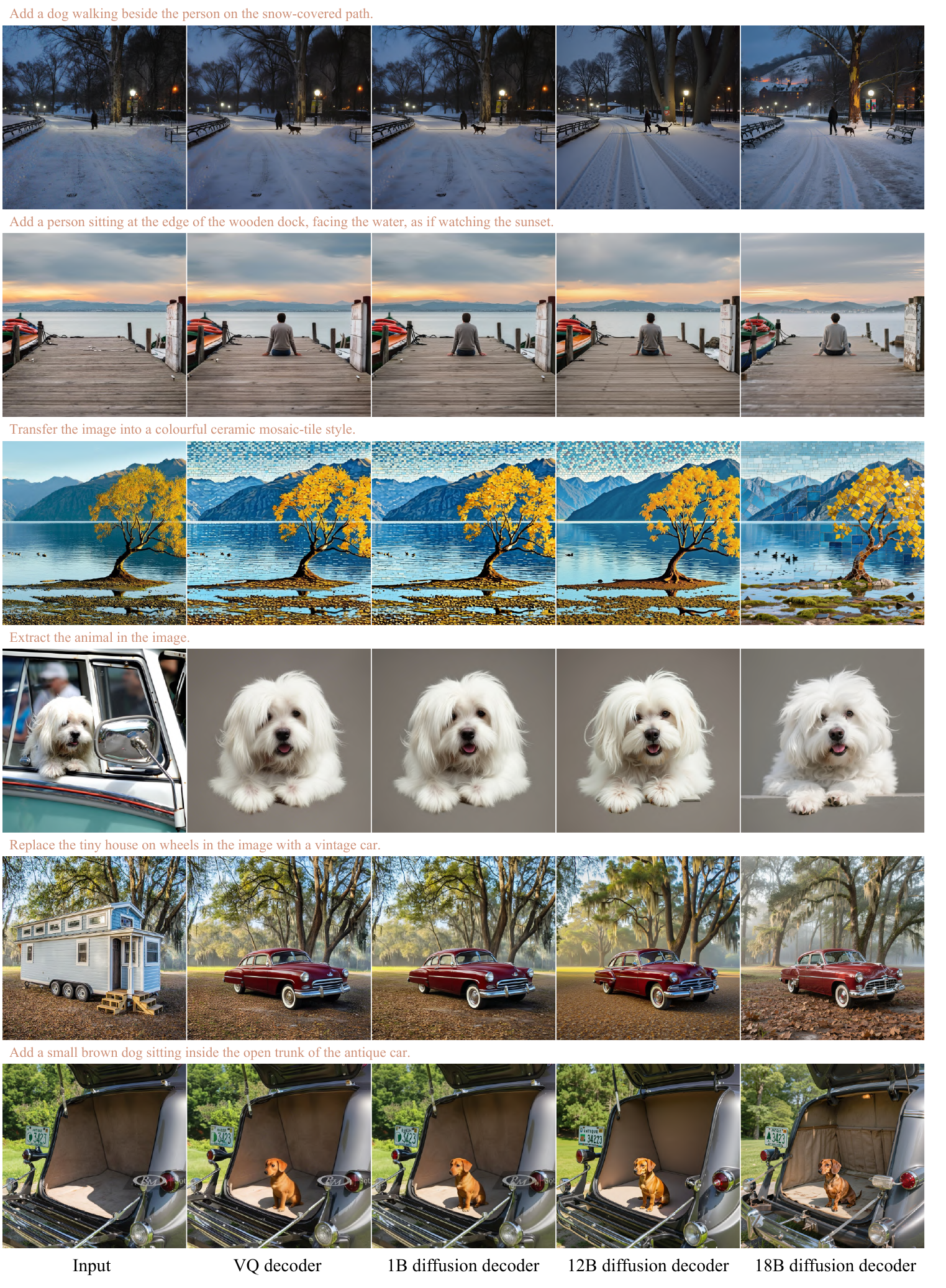}
    \vspace{-2em}
    \caption{\textbf{Qualitative comparison of VQ decoder versus diffusion decoders on image editing scenario.}}
    \label{fig:decoder_comparison_edit}
\end{figure*}


\section{Conclusion, Limitations and Future Directions}

In this work, we introduced \textbf{\methodName{}}, a unified sequential modeling that activates multimodal understanding and generation. By shifting from the traditional raster-scan paradigm to a next-scale prediction framework, we have demonstrated that autoregressive models can achieve inference speeds comparable to, or exceeding, state-of-the-art diffusion models without compromising on visual fidelity or reasoning capabilities. Validated on a massive corpus of 6 trillion tokens, our contributions—ranging from the dual-codebook tokenizer and scale-reweighting training objective to the prefix-tuning GRPO—establish a robust recipe for training our unified multimodal systems. \methodName{} serves as a proof of concept that a single decoder-only transformer can effectively perceive, reason, and create.

Despite these advancements, limitations remain. While our dual-codebook tokenizer significantly improves semantic density, the discrete nature of vector quantization inevitably imposes an information bottleneck compared to continuous latent spaces, occasionally necessitating our optional diffusion decoder for hyper-realistic refinement. Furthermore, balancing the objectives of text generation and visual synthesis within a shared parameter space remains a non-trivial optimization challenge, particularly at lower parameter counts.

Looking forward, we identify three critical avenues to further scale and refine unified autoregressive modeling:
\begin{itemize}
\item \textbf{Data Scaling for Advanced Understanding.} While our current pre-training corpus is extensive, the ratio of high-quality, dense reasoning data remains a limiting factor for complex multimodal comprehension. Future iterations will prioritize the integration of more diverse and high-quality understanding data, specifically dense image captions and complex interleaved reasoning chains. We hypothesize that scaling the density of semantic supervision will unlock emergent reasoning capabilities in the visual domain, mirroring the trajectory of text-only LLMs.

\item \textbf{Model Scaling and Mixture-of-Experts (MoE).} Following neural scaling laws, we anticipate that increasing model capacity will yield predictable gains in both generation fidelity and instruction following. We conducted a toy experiment and proved that transition from dense architectures to Mixture-of-Experts (MoE) frameworks significantly improves the overall generation quality. 

\item \textbf{Next-Generation Tokenization.} The tokenizer remains the fundamental upper bound on autoregressive visual generation performance. We aim to develop improved tokenization strategies that achieve higher compression rates without sacrificing reconstruction quality. This includes exploring variable-rate quantization and semantic-aware compression that can further reduce the sequence length for high-resolution images, thereby compounding the efficiency gains of our next-scale prediction paradigm.

\item \textbf{Native Multimodal Chain-of-Thought and Unified RL.} Our decoder-only architecture naturally extends the "Chain-of-Thought" paradigm to "Thinking with Images," enabling the model to reason via intermediate visual generation. Unlike hybrid architectures such as Transfusion \cite{transfusion}—which grapple with disjoint optimization objectives for discrete text and continuous visual latents—our fully autoregressive framework allows for seamless reinforcement learning across modalities. This unique structural advantage permits the direct application of standard LLM RL techniques to multimodal reasoning paths, ensuring consistent alignment and unlocking the potential for self-improving multimodal intelligence.

\end{itemize}

%% file: sections/G_appendix.tex
\section{Additional Implementation Details and Analysis}

\subsection{Predefined Scale Schedules}
\label{appendix:scale_schedule}

We design a comprehensive set of scale schedules to enable flexible image generation across diverse aspect ratios. As detailed in Table \ref{tab:patch_num_list}, for each target aspect ratio $r$, we define a progressive scale schedule $\mathcal{S}_r = {(h^r_1, w^r_1), (h^r_2, w^r_2), \ldots, (h^r_K, w^r_K)}$, where $K=18$ represents the total number of scale progression steps.

In this design, each tuple $(h^r_k, w^r_k)$ maintains an aspect ratio approximately equal to $r$, with particular emphasis on preserving this ratio at higher scales where visual fidelity is critical. For any given scale index $k$, we ensure that the spatial area $h^r_k \times w^r_k$ remains approximately constant across different aspect ratios $r$. This design choice guarantees uniform training sequence lengths and balanced computational requirements.

For 256-scale images, we utilize the first 12 scales from the corresponding schedule as the scale setting. For 512-scale images, we utilize the first 16 scales as the scale setting. During inference, our method generates photorealistic images spanning an extensive range of aspect ratios from 1:4 to 4:1, following these predefined scale schedules. This approach enables consistent quality across diverse output dimensions while maintaining computational efficiency.

\begin{table*}[h]
\centering
\setlength{\tabcolsep}{2pt}
\captionsetup{skip=5pt} 
\caption{\textbf{Predefined scale schedules for progressive image generation across 40 aspect ratios.} Each row represents a complete 18-step scale progression from initial patch $(h^r_1, w^r_1)$ to final resolution $(h^r_{18}, w^r_{18})$.}
\label{tab:patch_num_list}
\resizebox{.99\linewidth}{!}{
\begin{tabular}{l c c c c c c c c c c c c c c c c c c c}
\toprule
Aspect Ratio & Resolution & \multicolumn{18}{c}{Scale Schedule} \\
\midrule
0.250 (1:4) & 512$\times$2048 & (1,1) & (1,4) & (2,6) & (2,8) & (2,10) & (3,12) & (4,14) & (4,16) & (5,20) & (6,24) & (7,28) & (8,32) & (10,40) & (12,48) & (14,56) & (16,64) & (24,96) & (32,128) \\
0.260 (13:50) & 544$\times$2080 & (1,1) & (1,4) & (2,6) & (2,8) & (2,10) & (3,12) & (4,14) & (4,16) & (5,19) & (6,23) & (7,27) & (8,31) & (10,39) & (12,46) & (14,54) & (16,62) & (24,93) & (32,124) \\
0.267 (4:15) & 576$\times$2160 & (1,1) & (1,4) & (2,6) & (2,8) & (2,9) & (3,11) & (4,13) & (4,15) & (5,19) & (6,22) & (7,26) & (8,30) & (10,38) & (12,45) & (14,52) & (16,60) & (24,90) & (32,120) \\
0.276 (10:36) & 576$\times$2088 & (1,1) & (1,4) & (2,5) & (2,7) & (2,9) & (3,11) & (4,13) & (4,14) & (5,18) & (6,22) & (7,25) & (8,29) & (10,36) & (12,44) & (14,51) & (16,58) & (24,87) & (32,116) \\
0.295 (11:37) & 592$\times$1984 & (1,1) & (1,4) & (2,5) & (2,7) & (3,9) & (3,10) & (4,12) & (4,14) & (6,18) & (7,21) & (8,24) & (9,28) & (11,35) & (14,42) & (16,49) & (18,56) & (27,84) & (36,112) \\
0.311 (9:29) & 624$\times$2001 & (1,1) & (1,3) & (2,5) & (2,6) & (3,8) & (3,10) & (4,11) & (4,13) & (6,16) & (7,20) & (8,23) & (9,26) & (11,32) & (14,39) & (16,46) & (18,52) & (27,78) & (36,104) \\
0.333 (1:3) & 576$\times$1728 & (1,1) & (1,3) & (2,5) & (2,7) & (3,8) & (3,10) & (4,12) & (4,14) & (6,17) & (7,20) & (8,24) & (9,27) & (11,34) & (14,40) & (16,47) & (18,54) & (27,81) & (36,108) \\
0.345 (10:29) & 640$\times$1856 & (1,1) & (1,3) & (2,5) & (2,6) & (3,8) & (4,9) & (4,11) & (5,12) & (6,16) & (8,19) & (9,22) & (10,25) & (12,31) & (15,38) & (18,44) & (20,50) & (30,75) & (40,100) \\
0.417 (5:12) & 800$\times$1920 & (1,1) & (1,3) & (2,4) & (2,6) & (3,8) & (4,9) & (4,10) & (5,12) & (6,15) & (8,18) & (9,21) & (10,24) & (12,30) & (15,36) & (18,42) & (20,48) & (30,72) & (40,96) \\
0.478 (11:23) & 880$\times$1840 & (1,1) & (1,3) & (2,4) & (3,6) & (3,7) & (4,9) & (5,10) & (6,12) & (7,14) & (8,17) & (10,20) & (11,23) & (14,29) & (16,34) & (19,40) & (22,46) & (33,69) & (44,92) \\
0.500 (1:2) & 880$\times$1760 & (1,1) & (1,3) & (2,4) & (3,6) & (3,7) & (4,8) & (5,10) & (6,11) & (7,14) & (8,16) & (10,19) & (11,22) & (14,28) & (16,33) & (19,38) & (22,44) & (33,66) & (44,88) \\
0.524 (11:21) & 896$\times$1709 & (1,1) & (1,3) & (2,4) & (3,5) & (3,7) & (4,8) & (5,9) & (6,10) & (7,13) & (8,16) & (10,18) & (11,21) & (14,26) & (16,32) & (19,37) & (22,42) & (33,63) & (44,84) \\
0.571 (4:7) & 896$\times$1568 & (1,1) & (2,3) & (2,4) & (3,5) & (4,7) & (4,8) & (5,9) & (6,10) & (8,13) & (9,16) & (10,18) & (12,21) & (15,26) & (18,32) & (21,37) & (24,42) & (36,63) & (48,84) \\
0.600 (3:5) & 960$\times$1600 & (1,1) & (2,2) & (2,4) & (3,5) & (4,6) & (4,8) & (5,9) & (6,10) & (8,12) & (9,15) & (10,18) & (12,20) & (15,25) & (18,30) & (21,35) & (24,40) & (36,60) & (48,80) \\
0.685 (13:19) & 816$\times$1192 & (1,1) & (2,2) & (2,4) & (3,5) & (4,6) & (5,7) & (6,8) & (6,10) & (8,12) & (10,14) & (11,17) & (13,19) & (16,24) & (20,28) & (23,33) & (26,38) & (39,57) & (52,76) \\
0.722 (13:18) & 864$\times$1196 & (1,1) & (2,2) & (2,3) & (3,4) & (4,6) & (5,7) & (6,8) & (6,9) & (8,11) & (10,14) & (11,16) & (13,18) & (16,22) & (20,27) & (23,32) & (26,36) & (39,54) & (52,72) \\
0.781 (25:32) & 992$\times$1270 & (1,1) & (2,2) & (3,3) & (4,4) & (4,6) & (5,7) & (6,8) & (7,9) & (9,11) & (10,14) & (12,16) & (14,18) & (18,22) & (21,27) & (24,32) & (28,36) & (42,54) & (56,72) \\
0.824 (14:17) & 1008$\times$1224 & (1,1) & (2,2) & (3,3) & (4,4) & (4,5) & (5,6) & (6,7) & (7,8) & (9,11) & (10,13) & (12,15) & (14,17) & (18,21) & (21,26) & (24,30) & (28,34) & (42,51) & (56,68) \\
0.882 (15:17) & 960$\times$1088 & (1,1) & (2,2) & (3,3) & (4,4) & (5,5) & (6,6) & (7,7) & (8,8) & (9,11) & (11,13) & (13,15) & (15,17) & (19,21) & (22,26) & (26,30) & (30,34) & (45,51) & (60,68) \\
0.937 (15:16) & 960$\times$1024 & (1,1) & (2,2) & (3,3) & (4,4) & (5,5) & (6,6) & (7,7) & (8,8) & (9,10) & (11,12) & (13,14) & (15,16) & (19,20) & (22,24) & (26,28) & (30,32) & (45,48) & (60,64) \\
1.000 (1:1) & 1024$\times$1024 & (1,1) & (2,2) & (3,3) & (4,4) & (5,5) & (6,6) & (7,7) & (8,8) & (10,10) & (12,12) & (14,14) & (16,16) & (20,20) & (24,24) & (28,28) & (32,32) & (48,48) & (64,64) \\
1.067 (16:15) & 1024$\times$960 & (1,1) & (2,2) & (3,3) & (4,4) & (5,5) & (6,6) & (7,7) & (8,8) & (10,9) & (12,11) & (14,13) & (16,15) & (20,19) & (24,22) & (28,26) & (32,30) & (48,45) & (64,60) \\
1.133 (17:15) & 1088$\times$960 & (1,1) & (2,2) & (3,3) & (4,4) & (5,5) & (6,6) & (7,7) & (8,8) & (11,9) & (13,11) & (15,13) & (17,15) & (21,19) & (26,22) & (30,26) & (34,30) & (51,45) & (68,60) \\
1.214 (17:14) & 1224$\times$1008 & (1,1) & (2,2) & (3,3) & (4,4) & (5,4) & (6,5) & (7,6) & (8,7) & (11,9) & (13,10) & (15,12) & (17,14) & (21,18) & (26,21) & (30,24) & (34,28) & (51,42) & (68,56) \\
1.286 (9:7) & 1270$\times$992 & (1,1) & (2,2) & (3,3) & (4,4) & (6,4) & (7,5) & (8,6) & (9,7) & (11,9) & (14,10) & (16,12) & (18,14) & (22,18) & (27,21) & (32,24) & (36,28) & (54,42) & (72,56) \\
1.385 (18:13) & 1196$\times$864 & (1,1) & (2,2) & (3,2) & (4,3) & (6,4) & (7,5) & (8,6) & (9,6) & (11,8) & (14,10) & (16,11) & (18,13) & (22,16) & (27,20) & (32,23) & (36,26) & (54,39) & (72,52) \\
1.462 (19:13) & 1192$\times$816 & (1,1) & (2,2) & (4,2) & (5,3) & (6,4) & (7,5) & (8,6) & (10,6) & (12,8) & (14,10) & (17,11) & (19,13) & (24,16) & (28,20) & (33,23) & (38,26) & (57,39) & (76,52) \\
1.667 (5:3) & 1600$\times$960 & (1,1) & (2,2) & (4,2) & (5,3) & (6,4) & (8,4) & (9,5) & (10,6) & (12,8) & (15,9) & (18,10) & (20,12) & (25,15) & (30,18) & (35,21) & (40,24) & (60,36) & (80,48) \\
1.750 (7:4) & 1568$\times$896 & (1,1) & (3,2) & (4,2) & (5,3) & (7,4) & (8,4) & (9,5) & (10,6) & (13,8) & (16,9) & (18,10) & (21,12) & (26,15) & (32,18) & (37,21) & (42,24) & (63,36) & (84,48) \\
1.909 (21:11) & 1709$\times$896 & (1,1) & (3,1) & (4,2) & (5,3) & (7,3) & (8,4) & (9,5) & (10,6) & (13,7) & (16,8) & (18,10) & (21,11) & (26,14) & (32,16) & (37,19) & (42,22) & (63,33) & (84,44) \\
2.000 (2:1) & 1760$\times$880 & (1,1) & (3,1) & (4,2) & (6,3) & (7,3) & (8,4) & (10,5) & (11,6) & (14,7) & (16,8) & (19,10) & (22,11) & (28,14) & (33,16) & (38,19) & (44,22) & (66,33) & (88,44) \\
2.091 (23:11) & 1840$\times$880 & (1,1) & (3,1) & (4,2) & (6,3) & (7,3) & (9,4) & (10,5) & (12,6) & (14,7) & (17,8) & (20,10) & (23,11) & (29,14) & (34,16) & (40,19) & (46,22) & (69,33) & (92,44) \\
2.400 (12:5) & 1920$\times$800 & (1,1) & (3,1) & (4,2) & (6,2) & (8,3) & (9,4) & (10,4) & (12,5) & (15,6) & (18,8) & (21,9) & (24,10) & (30,12) & (36,15) & (42,18) & (48,20) & (72,30) & (96,40) \\
2.500 (5:2) & 1856$\times$640 & (1,1) & (3,1) & (5,2) & (6,2) & (8,3) & (9,4) & (11,4) & (12,5) & (16,6) & (19,8) & (22,9) & (25,10) & (31,12) & (38,15) & (44,18) & (50,20) & (75,30) & (100,40) \\
2.889 (26:9) & 2001$\times$624 & (1,1) & (3,1) & (5,2) & (6,2) & (8,3) & (10,3) & (11,4) & (13,4) & (16,6) & (20,7) & (23,8) & (26,9) & (32,11) & (39,14) & (46,16) & (52,18) & (78,27) & (104,36) \\
3.000 (3:1) & 1728$\times$576 & (1,1) & (3,1) & (5,2) & (7,2) & (8,3) & (10,3) & (12,4) & (14,4) & (17,6) & (20,7) & (24,8) & (27,9) & (34,11) & (40,14) & (47,16) & (54,18) & (81,27) & (108,36) \\
3.111 (28:9) & 1984$\times$592 & (1,1) & (4,1) & (5,2) & (7,2) & (9,3) & (10,3) & (12,4) & (14,4) & (18,6) & (21,7) & (24,8) & (28,9) & (35,11) & (42,14) & (49,16) & (56,18) & (84,27) & (112,36) \\
3.625 (29:8) & 2088$\times$576 & (1,1) & (4,1) & (5,2) & (7,2) & (9,2) & (11,3) & (13,4) & (14,4) & (18,5) & (22,6) & (25,7) & (29,8) & (36,10) & (44,12) & (51,14) & (58,16) & (87,24) & (116,32) \\
3.750 (15:4) & 2160$\times$576 & (1,1) & (4,1) & (6,2) & (8,2) & (9,2) & (11,3) & (13,4) & (15,4) & (19,5) & (22,6) & (26,7) & (30,8) & (38,10) & (45,12) & (52,14) & (60,16) & (90,24) & (120,32) \\
3.875 (31:8) & 2080$\times$544 & (1,1) & (4,1) & (6,2) & (8,2) & (10,2) & (12,3) & (14,4) & (16,4) & (19,5) & (23,6) & (27,7) & (31,8) & (39,10) & (46,12) & (54,14) & (62,16) & (93,24) & (124,32) \\
4.000 (4:1) & 2048$\times$512 & (1,1) & (4,1) & (6,2) & (8,2) & (10,2) & (12,3) & (14,4) & (16,4) & (20,5) & (24,6) & (28,7) & (32,8) & (40,10) & (48,12) & (56,14) & (64,16) & (96,24) & (128,32) \\
\bottomrule
\end{tabular}
}
\end{table*}

\newpage

\subsection{Inference Efficiency Analysis}

\begin{figure*}[ht]
    \centering
    \includegraphics[width=\linewidth]{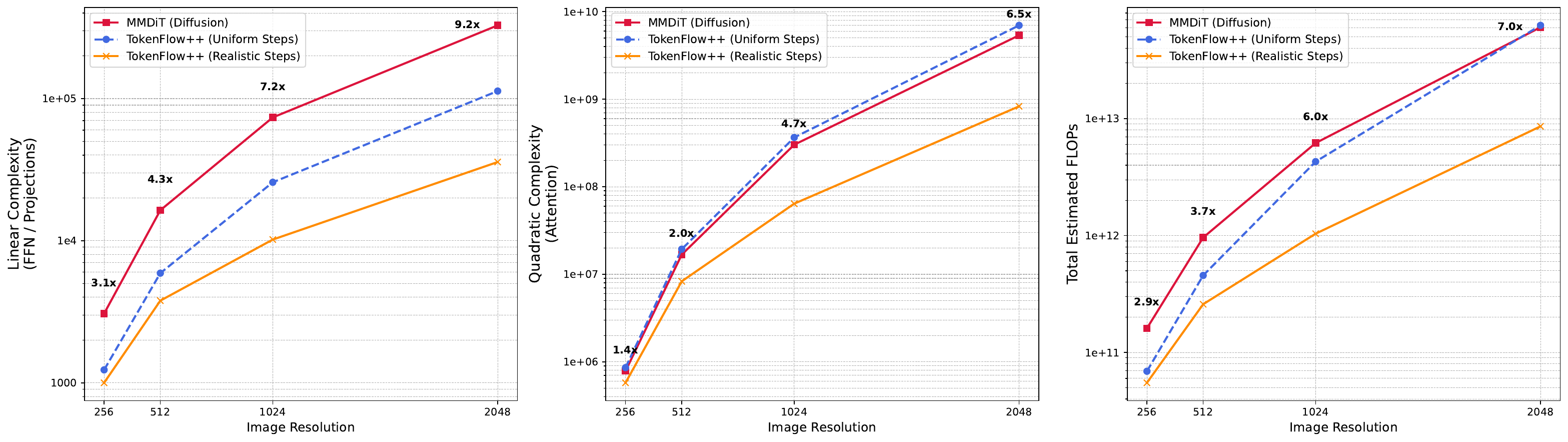}
    \caption{\textbf{Inference efficiency comparison between \methodName{} and MMDiT across different image resolutions.} Both models use the same number of sampling steps for fair comparison. We evaluate two scale scheduling strategies for \methodName{}: (1) \textit{Uniform Steps}, where the scale resolution increases linearly with each step; (2) \textit{Realistic Steps}, which reflects our actual training configuration that allocates more steps to lower scales and fewer steps to higher resolutions. The annotations show the FLOPs ratio of MMDiT to \methodName{}. As resolution increases, the efficiency advantage of \methodName{} becomes more pronounced, achieving up to $6\times$ reduction in total FLOPs.}
    \label{fig:infer_efficiency}
\end{figure*}

In this section, we provide a comprehensive theoretical analysis comparing the computational costs of \methodName{}, which adopts the next-scale prediction paradigm from VAR~\cite{var}, against the Multi-Modal Diffusion Transformer (MMDiT)~\cite{SD3} baseline. We focus primarily on the inference phase, where efficiency is critical for real-world deployment.

\textbf{Inference Efficiency.}
To evaluate the inference cost fairly, we compare the Floating Point Operations (FLOPs) required to generate a $1024 \times 1024$ resolution image (approximately $64 \times 64$ latent tokens). Both models utilize a similar hyperparameter setup: a hidden size $h$ (tested at 2048 and 4096) and a total of $T=18$ sampling steps.

For a standard Transformer layer, the FLOPs for the Attention mechanism and the Feed-Forward Network (FFN) are defined as~\cite{var,dao2023flashattention2}:
\begin{equation}
    \text{FLOPs}_{\text{attn}} \approx 4s_q h \cdot (2h) + 4 s_q s_{kv} h = 8 s_q h^2 + 4 s_q s_{kv} h
\end{equation}
\begin{equation}
    \text{FLOPs}_{\text{MLP}} \approx 2 \cdot 2 s_q h^2 = 4 s_q h^2
\end{equation}
where $s_q$ is the query sequence length and $s_{kv}$ is the key/value sequence length. The total FLOPs per layer is:
\begin{equation}
    \text{FLOPs}_{\text{layer}} = 12 s_q h^2 + 4 s_q s_{kv} h
\end{equation}

\textbf{MMDiT Cost Analysis.}
MMDiT~\cite{SD3} operates on a fixed sequence length $s = 64 \times 64 = 4096$ across all $T=18$ denoising steps. Since $s_q = s_{kv} = s$, the total inference cost is:
\begin{equation}
    \text{FLOPs}_{\text{MMDiT}} = T \cdot L \cdot (12sh^2 + 4s^2h) \approx 884{,}736 \cdot h^2 + 1{,}207{,}959{,}552 \cdot h
\end{equation}

\textbf{\methodName{} Cost Analysis.}
In contrast, \methodName{} employs the next-scale prediction paradigm~\cite{var}. The sequence length $s_q$ varies dynamically at each step (e.g., $1, 4, 9, \dots, 4096$), and thanks to the autoregressive nature, we utilize KV-caching. This means $s_q$ is the number of new tokens generated in the current step, while $s_{kv}$ is the cumulative sum of all previous tokens. The total inference cost is:
\begin{equation}
    \text{FLOPs}_{\methodName{}} = \sum_{t=1}^{T} L \cdot (12 s_{q}^{(t)} h^2 + 4 s_{q}^{(t)} s_{kv}^{(t)} h) \approx 122{,}208 \cdot h^2 + 256{,}215{,}912 \cdot h
\end{equation}

\textbf{Comparative Results.}
By substituting typical hidden sizes (e.g., $h=4096$), the ratio of computational cost is:
\begin{equation}
    \frac{\text{FLOPs}_{\text{MMDiT}}}{\text{FLOPs}_{\methodName{}}} \approx \frac{6.18 \times 10^{12}}{1.04 \times 10^{12}} \approx 5.96
\end{equation}

This analysis demonstrates that MMDiT requires approximately \textbf{6$\times$} more FLOPs than \methodName{} to generate an image of the same resolution. The efficiency gain stems from two key factors: (1) \textit{FFN and Linear Projections}: These costs depend linearly on the current step's token count ($s_q$). Since the next-scale prediction paradigm generates fewer tokens in early steps, the average computational load is significantly lower. In our implementation, MMDiT consumes $\approx 7.2\times$ more FLOPs in these components. (2) \textit{Attention Mechanism}: While memory access for Keys/Values grows cumulatively, the computational cost for Queries is minimal in early steps. Overall, MMDiT's attention computation is $\approx 4.7\times$ that of \methodName{}.

In summary, for deployment, \methodName{} offers a definitive advantage, \textbf{reducing inference FLOPs by a factor of 6} compared to state-of-the-art diffusion transformers~\cite{SD3,flux}.

Figure~\ref{fig:infer_efficiency} visualizes this efficiency comparison across different image resolutions. For fair comparison, both MMDiT and \methodName{} use the same number of sampling steps at each resolution. We evaluate two scale scheduling strategies: (1) \textit{Uniform Steps}, where scale resolution increases linearly from $1 \times 1$ to the target resolution; (2) \textit{Realistic Steps}, which reflects our actual training configuration—allocating more steps to lower scales for better structural modeling and fewer steps to higher resolutions where fine details are refined. The realistic schedule which we use achieves slightly better efficiency while maintaining generation quality.

\section{EditCanvas benchmark}
\label{sec:editcanvas}

To comprehensively evaluate the capabilities of modern image editing models, we introduce \textbf{EditCanvas}, a novel and meticulously structured benchmark. EditCanvas is designed to address the limitations of existing benchmarks by offering a more systematic and fine-grained assessment framework.
A key innovation of EditCanvas is its hierarchical, three-level labeling system, which categorizes editing tasks with high granularity. At the highest level, we distinguish between two fundamental paradigms: \textbf{Traditional Editing} and \textbf{Subject-Driven Generation}.
\begin{itemize}
    \item \textbf{Traditional Editing} focuses on direct modifications to a reference image, akin to benchmarks like GEdit-Bench. This category is further divided into five sub-categories: Text Editing, Global Editing, Local Editing, Viewpoint Editing, and Style Editing.
    \item \textbf{Subject-Driven Generation}, similar in spirit to OmniContext, involves extracting a specific subject from a reference image to generate a new scene. This is broken down into four sub-categories: Style, Character, Object, and Abstract Concept.
\end{itemize}
This structure extends to a third level with 56 fine-grained tasks (40 for Traditional Editing and 16 for Subject-Driven Generation), such as object addition, removal, replacement, and repositioning. This detailed taxonomy allows for a nuanced analysis of a model's performance across a wide spectrum of editing scenarios, providing deeper insights than previously possible.
The EditCanvas dataset comprises 5,221 high-quality samples. We began by sourcing high-resolution images (over 1K) from open datasets like LAION and COYO, filtering out those with plain white backgrounds. We then employed a state-of-the-art Vision-Language Model (VLM) to automatically generate corresponding editing instructions for each image. To ensure the benchmark's reliability and quality, these samples underwent a rigorous two-round manual filtering process, where we assessed factors such as image clarity, content appropriateness, and the suitability of the generated instructions.
For evaluation, EditCanvas adopts a hybrid metric approach inspired by its predecessors. For Traditional Editing tasks, we follow the methodology of GEdit-Bench, using GPT-4.1 to assess \textit{Instruction Following} and the presence of \textit{Artifacts}. For Subject-Driven Generation tasks, we align with OmniContext's protocol, evaluating \textit{Instruction Following} and \textit{Content Consistency} using GPT-4.1.

\begin{table}[t]
\centering
\resizebox{0.5\linewidth}{!}{
\begin{tabular}{lcccc}
\toprule
\textbf{Benchmarks} & \textbf{Size} & \textbf{Real Image} & \textbf{Human Filtering} & \textbf{Sub-tasks} \\
\midrule
EditBench~\cite{wang2023imagen} & 240 & \cmark & \xmark & 1 \\
EmuEdit~\cite{sheynin2024emu} & 3,055 & \cmark & \xmark & 7 \\
HIVE~\cite{zhang2024hive} & 1,000 & \cmark & \cmark & 1 \\
HQ-Eidt~\cite{hui2024hq} & 1,640 & \xmark & \xmark & 7 \\
MagicBrush~\cite{zhang2023magicbrush} & 1,053 & \cmark & \cmark & 7 \\
AnyEdit~\cite{shen2024many} & 1,250 & \cmark & \xmark & 25 \\
ICE-Bench~\cite{pan2025ice} & 6,538 & \cmark & \cmark & 31 \\
Gedit-Bench~\cite{liu2025step1x} & 606 & \cmark & \cmark & 11 \\
\midrule
\textbf{EditCanvas (Ours)} & 5,221 & \cmark & \cmark & 56 \\
\bottomrule
\end{tabular}
}
\caption{\textbf{Key Attributes of Open-source Edit Benchmarks}. The reliance of existing open-source benchmarks on synthetic user inputs and minimal human involvement highlights the necessity of our proposed EditCanvas.}
\label{tab:dataset_comparison}
\end{table}

\begin{figure*}[ht]
    \centering
    \includegraphics[width=0.4\linewidth]{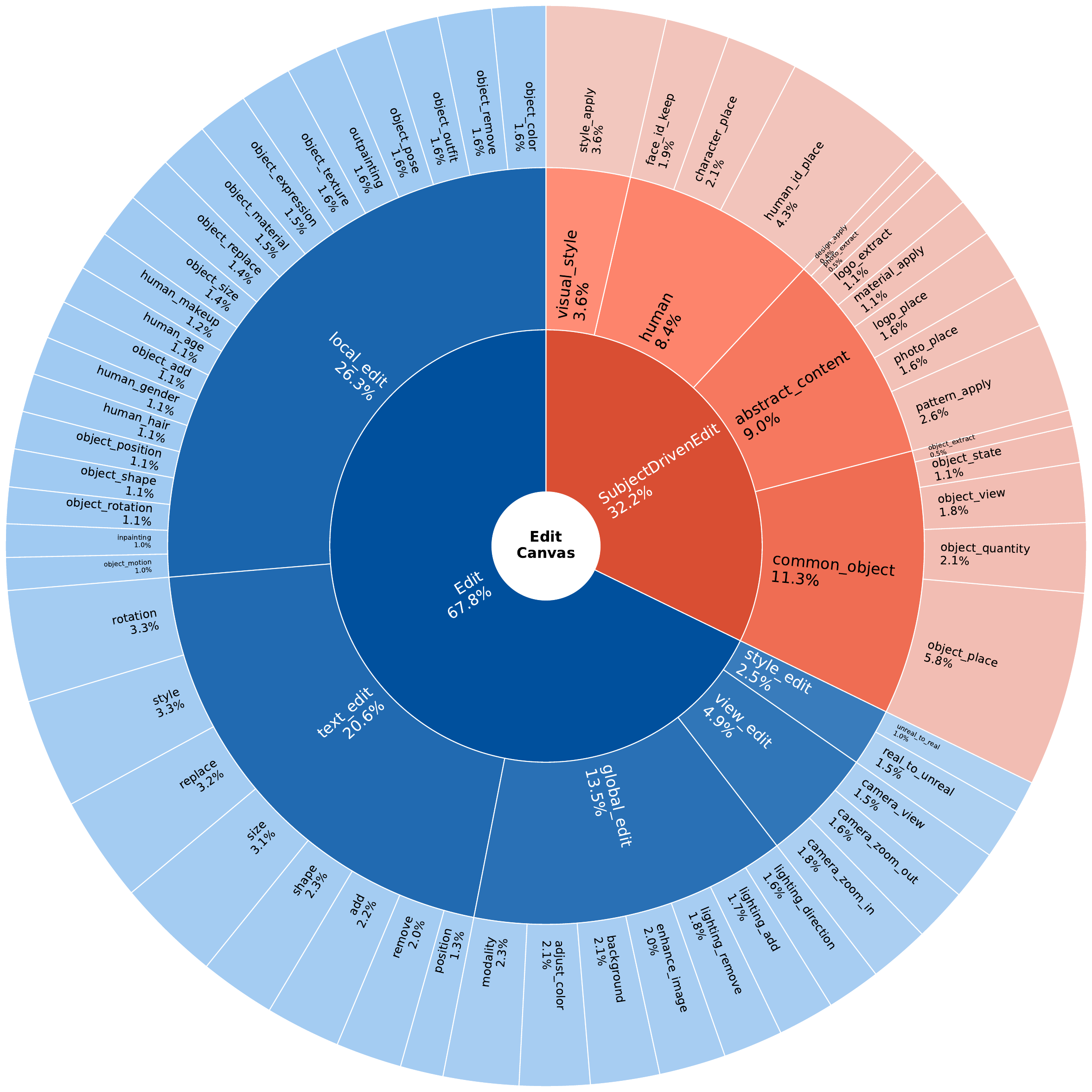}
    \caption{\textbf{Distribution of EditCanvas Tasks.} The chart illustrates the hierarchical structure of the dataset, dividing tasks into Traditional Editing (blue, 67.8\%) and Subject-Driven Editing (orange, 32.2\%). The outer layers represent the sub-categories and specific fine-grained editing tasks.}    \label{fig:editcanvas_class}
\end{figure*}

Compared to existing benchmarks such as GEdit-Bench and OmniContext, EditCanvas presents several key advantages:
\begin{enumerate}
    \item \textbf{Unified and Comprehensive Framework:} While GEdit-Bench focuses on genuine user instructions for direct editing and OmniContext specializes in in-context subject generation, EditCanvas is the first benchmark to systematically integrate both paradigms into a single, unified framework. This allows for a holistic evaluation of a model's diverse editing capabilities, from minor tweaks to complex generative tasks.
    \item \textbf{Granular, Hierarchical Taxonomy:} EditCanvas's three-level classification system is significantly more detailed than the taxonomies of previous benchmarks. While GEdit-Bench offers 11 categories and OmniContext defines 8 subtasks, EditCanvas provides 57 distinct third-level tasks. This granularity enables precise identification of a model's specific strengths and weaknesses, for instance, distinguishing its performance in "object addition" versus "object repositioning."
    \item \textbf{Large-Scale, High-Quality Dataset:} With over 5,000 carefully curated samples, EditCanvas offers a larger and more diverse dataset compared to GEdit-Bench's 606 examples. The semi-automated generation process, combining a powerful VLM with multi-round human verification, ensures both scale and quality, creating a robust testbed for modern models.
\end{enumerate}
By providing a structured, comprehensive, and large-scale benchmark, EditCanvas sets a new standard for evaluating image editing models, enabling a more thorough and insightful understanding of their real-world performance.

%% file: main.bib
@inproceedings{qu2025tokenflow,
  title={Tokenflow: Unified image tokenizer for multimodal understanding and generation},
  author={Qu, Liao and Zhang, Huichao and Liu, Yiheng and Wang, Xu and Jiang, Yi and Gao, Yiming and Ye, Hu and Du, Daniel K and Yuan, Zehuan and Wu, Xinglong},
  booktitle={Proceedings of the Computer Vision and Pattern Recognition Conference},
  pages={2545--2555},
  year={2025}
}

@article{var,
  title={Visual autoregressive modeling: Scalable image generation via next-scale prediction},
  author={Tian, Keyu and Jiang, Yi and Yuan, Zehuan and Peng, Bingyue and Wang, Liwei},
  journal={arXiv preprint arXiv:2404.02905},
  year={2024}
}

@article{Qwen2.5-VL,
  title={Qwen2.5-VL Technical Report},
  author={Bai, Shuai and Chen, Keqin and Liu, Xuejing and Wang, Jialin and Ge, Wenbin and Song, Sibo and Dang, Kai and Wang, Peng and Wang, Shijie and Tang, Jun and Zhong, Humen and Zhu, Yuanzhi and Yang, Mingkun and Li, Zhaohai and Wan, Jianqiang and Wang, Pengfei and Ding, Wei and Fu, Zheren and Xu, Yiheng and Ye, Jiabo and Zhang, Xi and Xie, Tianbao and Cheng, Zesen and Zhang, Hang and Yang, Zhibo and Xu, Haiyang and Lin, Junyang},
  journal={arXiv preprint arXiv:2502.13923},
  year={2025}
}

@article{liu2025detailflow,
  title={DetailFlow: 1D Coarse-to-Fine Autoregressive Image Generation via Next-Detail Prediction},
  author={Liu, Yiheng and Qu, Liao and Zhang, Huichao and Wang, Xu and Jiang, Yi and Gao, Yiming and Ye, Hu and Li, Xian and Wang, Shuai and Du, Daniel K and others},
  journal={arXiv preprint arXiv:2505.21473},
  year={2025}
}

@inproceedings{han2025infinity,
  title={Infinity: Scaling bitwise autoregressive modeling for high-resolution image synthesis},
  author={Han, Jian and Liu, Jinlai and Jiang, Yi and Yan, Bin and Zhang, Yuqi and Yuan, Zehuan and Peng, Bingyue and Liu, Xiaobing},
  booktitle={Proceedings of the Computer Vision and Pattern Recognition Conference},
  pages={15733--15744},
  year={2025}
}

@inproceedings{
hsu2025ligerkernel,
title={Liger-Kernel: Efficient Triton Kernels for {LLM} Training},
author={Pin-Lun Hsu and Yun Dai and Vignesh Kothapalli and Qingquan Song and Shao Tang and Siyu Zhu and Steven Shimizu and Shivam Sahni and Haowen Ning and Yanning Chen and Zhipeng Wang},
booktitle={Championing Open-source DEvelopment in ML Workshop @ ICML25},
year={2025},
url={https://openreview.net/forum?id=36SjAIT42G}
}

@inproceedings{dao2023flashattention2,
  title={Flash{A}ttention-2: Faster Attention with Better Parallelism and Work Partitioning},
  author={Dao, Tri},
  booktitle={International Conference on Learning Representations (ICLR)},
  year={2024}
}

@inproceedings{rajbhandari2020zero,
  title={Zero: Memory optimizations toward training trillion parameter models},
  author={Rajbhandari, Samyam and Rasley, Jeff and Ruwase, Olatunji and He, Yuxiong},
  booktitle={SC20: International Conference for High Performance Computing, Networking, Storage and Analysis},
  pages={1--16},
  year={2020},
  organization={IEEE}
}

@inproceedings{sdxl,
  title={Sdxl: Improving latent diffusion models for high-resolution image synthesis},
  author={Podell, Dustin and English, Zion and Lacey, Kyle and Blattmann, Andreas and Dockhorn, Tim and M{\"u}ller, Jonas and Penna, Joe and Rombach, Robin},
  booktitle = {ICLR},
  year = {2024}
}

@article{dalle3,
  title={Improving image generation with better captions},
  author={Betker, James and Goh, Gabriel and Jing, Li and Brooks, Tim and Wang, Jianfeng and Li, Linjie and Ouyang, Long and Zhuang, Juntang and Lee, Joyce and Guo, Yufei and others},
  journal={OpenAI blog},
  year={2023}
}

@misc{flux,
  title = {FLUX},
  author = {Black Forest Labs},
  year = {2024},
  url = {https://github.com/black-forest-labs/flux}
}

@inproceedings{SD3,
  title={Scaling rectified flow transformers for high-resolution image synthesis},
  author={Esser, Patrick and Kulal, Sumith and Blattmann, Andreas and Entezari, Rahim and M{\"u}ller, Jonas and Saini, Harry and Levi, Yam and Lorenz, Dominik and Sauer, Axel and Boesel, Frederic and others},
  booktitle={ICML},
  year={2024}
}

@article{transfusion,
  title={Transfusion: Predict the next token and diffuse images with one multi-modal model},
  author={Zhou, Chunting and Yu, Lili and Babu, Arun and Tirumala, Kushal and Yasunaga, Michihiro and Shamis, Leonid and Kahn, Jacob and Ma, Xuezhe and Zettlemoyer, Luke and Levy, Omer},
  journal={arXiv preprint arxiv:2408.11039},
  year={2024}
}

@article{emu3,
  title={Emu3: Next-token prediction is all you need},
  author={Wang, Xinlong and Zhang, Xiaosong and Luo, Zhengxiong and Sun, Quan and Cui, Yufeng and Wang, Jinsheng and Zhang, Fan and Wang, Yueze and Li, Zhen and Yu, Qiying and others},
  journal={arXiv preprint arxiv:2409.18869},
  year={2024}
}

@article{chameleon,
  title={Chameleon: Mixed-modal early-fusion foundation models},
  author={Team, Chameleon},
  journal={arXiv preprint arXiv:2405.09818},
  year={2024}
}

@article{show-o,
  title={Show-o: One single transformer to unify multimodal understanding and generation},
  author={Xie, Jinheng and Mao, Weijia and Bai, Zechen and Zhang, David Junhao and Wang, Weihao and Lin, Kevin Qinghong and Gu, Yuchao and Chen, Zhijie and Yang, Zhenheng and Shou, Mike Zheng},
  journal={arXiv preprint arxiv:2408.12528},
  year={2024}
}

@article{januspro2025,
  title={Janus-Pro: Unified Multimodal Understanding and Generation with Data and Model Scaling},
  author={Xiaokang Chen and Chengyue Wu and Zhiyu Wu and Yiyang Ma and Xingchao Liu and Zizheng Pan and Wen Liu and Zhenda Xie and Xingkai Yu and Chong Ruan and Ping Luo},
  journal={arXiv preprint arXiv:2501.17811},
  year={2025},
}

@article{bagel,
  title={Emerging properties in unified multimodal pretraining},
  author={Deng, Chaorui and Zhu, Deyao and Li, Kunchang and Gou, Chenhui and Li, Feng and Wang, Zeyu and Zhong, Shu and Yu, Weihao and Nie, Xiaonan and Song, Ziang and others},
  journal={arXiv preprint arXiv:2505.14683},
  year={2025}
}

@article{team2025nextstep,
  title={NextStep-1: Toward Autoregressive Image Generation with Continuous Tokens at Scale},
  author={Team, NextStep and Han, Chunrui and Li, Guopeng and Wu, Jingwei and Sun, Quan and Cai, Yan and Peng, Yuang and Ge, Zheng and Zhou, Deyu and Tang, Haomiao and others},
  journal={arXiv preprint arXiv:2508.10711},
  year={2025}
}

@article{cai2025hidream,
  title={HiDream-I1: A High-Efficient Image Generative Foundation Model with Sparse Diffusion Transformer},
  author={Cai, Qi and Chen, Jingwen and Chen, Yang and Li, Yehao and Long, Fuchen and Pan, Yingwei and Qiu, Zhaofan and Zhang, Yiheng and Gao, Fengbin and Xu, Peihan and others},
  journal={arXiv preprint arXiv:2505.22705},
  year={2025}
}

@article{ghosh2023geneval,
  title={Geneval: An object-focused framework for evaluating text-to-image alignment},
  author={Ghosh, Dhruba and Hajishirzi, Hannaneh and Schmidt, Ludwig},
  journal={Advances in Neural Information Processing Systems},
  volume={36},
  pages={52132--52152},
  year={2023}
}

@article{dpg,
  title={Ella: Equip diffusion models with llm for enhanced semantic alignment},
  author={Hu, Xiwei and Wang, Rui and Fang, Yixiao and Fu, Bin and Cheng, Pei and Yu, Gang},
  journal={arXiv preprint arXiv:2403.05135},
  year={2024}
}

@article{gao2025seedream,
  title={Seedream 3.0 technical report},
  author={Gao, Yu and Gong, Lixue and Guo, Qiushan and Hou, Xiaoxia and Lai, Zhichao and Li, Fanshi and Li, Liang and Lian, Xiaochen and Liao, Chao and Liu, Liyang and others},
  journal={arXiv preprint arXiv:2504.11346},
  year={2025}
}

@article{gong2025seedream,
  title={Seedream 2.0: A native chinese-english bilingual image generation foundation model},
  author={Gong, Lixue and Hou, Xiaoxia and Li, Fanshi and Li, Liang and Lian, Xiaochen and Liu, Fei and Liu, Liyang and Liu, Wei and Lu, Wei and Shi, Yichun and others},
  journal={arXiv preprint arXiv:2503.07703},
  year={2025}
}

@misc{gptimage,
  title={GPT-Image-1},
  author={OpenAI},
  year={2025},
  url={https://openai.com/index/introducing-4o-image-generation/},
}

@article{niu2025wise,
  title={WISE: A World Knowledge-Informed Semantic Evaluation for Text-to-Image Generation},
  author={Niu, Yuwei and Ning, Munan and Zheng, Mengren and Jin, Weiyang and Lin, Bin and Jin, Peng and Liao, Jiaqi and Ning, Kunpeng and Feng, Chaoran and Zhu, Bin and Yuan, Li},
  journal={arXiv preprint arXiv:2503.07265},
  year={2025}
}

@article{wu2024liquid,
  title={Liquid: Language models are scalable and unified multi-modal generators},
  author={Wu, Junfeng and Jiang, Yi and Ma, Chuofan and Liu, Yuliang and Zhao, Hengshuang and Yuan, Zehuan and Bai, Song and Bai, Xiang},
  journal={arXiv preprint arXiv:2412.04332},
  year={2024}
}

@article{bengio2015scheduled,
  title={Scheduled sampling for sequence prediction with recurrent neural networks},
  author={Bengio, Samy and Vinyals, Oriol and Jaitly, Navdeep and Shazeer, Noam},
  journal={Advances in neural information processing systems},
  volume={28},
  year={2015}
}

@article{ren2025beyond,
  title={Beyond next-token: Next-x prediction for autoregressive visual generation},
  author={Ren, Sucheng and Yu, Qihang and He, Ju and Shen, Xiaohui and Yuille, Alan and Chen, Liang-Chieh},
  journal={arXiv preprint arXiv:2502.20388},
  year={2025}
}

@article{chen2024diffusion,
  title={Diffusion forcing: Next-token prediction meets full-sequence diffusion},
  author={Chen, Boyuan and Mart{\'\i} Mons{\'o}, Diego and Du, Yilun and Simchowitz, Max and Tedrake, Russ and Sitzmann, Vincent},
  journal={Advances in Neural Information Processing Systems},
  volume={37},
  pages={24081--24125},
  year={2024}
}

@article{huang2025self,
  title={Self Forcing: Bridging the Train-Test Gap in Autoregressive Video Diffusion},
  author={Huang, Xun and Li, Zhengqi and He, Guande and Zhou, Mingyuan and Shechtman, Eli},
  journal={arXiv preprint arXiv:2506.08009},
  year={2025}
}

@article{schuhmann2022laion,
  title={Laion-5b: An open large-scale dataset for training next generation image-text models},
  author={Schuhmann, Christoph and Beaumont, Romain and Vencu, Richard and Gordon, Cade and Wightman, Ross and Cherti, Mehdi and Coombes, Theo and Katta, Aarush and Mullis, Clayton and Wortsman, Mitchell and others},
  journal={Advances in neural information processing systems},
  volume={35},
  pages={25278--25294},
  year={2022}
}

@article{chang2023muse,
  title={Muse: Text-to-image generation via masked generative transformers},
  author={Chang, Huiwen and Zhang, Han and Barber, Jarred and Maschinot, AJ and Lezama, Jose and Jiang, Lu and Yang, Ming-Hsuan and Murphy, Kevin and Freeman, William T and Rubinstein, Michael and others},
  journal={arXiv preprint arXiv:2301.00704},
  year={2023}
}

@article{jiao2025flexvar,
  title={Flexvar: Flexible visual autoregressive modeling without residual prediction},
  author={Jiao, Siyu and Zhang, Gengwei and Qian, Yinlong and Huang, Jiancheng and Zhao, Yao and Shi, Humphrey and Ma, Lin and Wei, Yunchao and Jie, Zequn},
  journal={arXiv preprint arXiv:2502.20313},
  year={2025}
}

@misc{flux2024,
    author={Black Forest Labs},
    title={FLUX},
    year={2024},
    howpublished={\url{https://github.com/black-forest-labs/flux}},
}

@inproceedings{rombach2022high,
  title={High-resolution image synthesis with latent diffusion models},
  author={Rombach, Robin and Blattmann, Andreas and Lorenz, Dominik and Esser, Patrick and Ommer, Bj{\"o}rn},
  booktitle={Proceedings of the IEEE/CVF conference on computer vision and pattern recognition},
  pages={10684--10695},
  year={2022}
}

@article{podell2023sdxl,
  title={Sdxl: Improving latent diffusion models for high-resolution image synthesis},
  author={Podell, Dustin and English, Zion and Lacey, Kyle and Blattmann, Andreas and Dockhorn, Tim and M{\"u}ller, Jonas and Penna, Joe and Rombach, Robin},
  journal={arXiv preprint arXiv:2307.01952},
  year={2023}
}

@misc{SD35,
  title={Stable Diffusion 3.5},
  author={Stability-AI},
  year={2024},
  url={https://github.com/Stability-AI/sd3.5},
}

@article{seedream3,
  title={Seedream 3.0 technical report},
  author={Gao, Yu and Gong, Lixue and Guo, Qiushan and Hou, Xiaoxia and Lai, Zhichao and Li, Fanshi and Li, Liang and Lian, Xiaochen and Liao, Chao and Liu, Liyang and others},
  journal={arXiv preprint arXiv:2504.11346},
  year={2025}
}

@article{hidream,
  title={HiDream-I1: A High-Efficient Image Generative Foundation Model with Sparse Diffusion Transformer},
  author={Cai, Qi and Chen, Jingwen and Chen, Yang and Li, Yehao and Long, Fuchen and Pan, Yingwei and Qiu, Zhaofan and Zhang, Yiheng and Gao, Fengbin and Xu, Peihan and others},
  journal={arXiv preprint arXiv:2505.22705},
  year={2025}
}

@article{januspro,
  title={Janus-pro: Unified multimodal understanding and generation with data and model scaling},
  author={Chen, Xiaokang and Wu, Zhiyu and Liu, Xingchao and Pan, Zizheng and Liu, Wen and Xie, Zhenda and Yu, Xingkai and Ruan, Chong},
  journal={arXiv preprint arXiv:2501.17811},
  year={2025}
}

@article{QwenImage,
  title={Qwen-image technical report},
  author={Wu, Chenfei and Li, Jiahao and Zhou, Jingren and Lin, Junyang and Gao, Kaiyuan and Yan, Kun and Yin, Sheng-ming and Bai, Shuai and Xu, Xiao and Chen, Yilei and others},
  journal={arXiv preprint arXiv:2508.02324},
  year={2025}
}

@article{ye2025imgedit,
  title={Imgedit: A unified image editing dataset and benchmark},
  author={Ye, Yang and He, Xianyi and Li, Zongjian and Lin, Bin and Yuan, Shenghai and Yan, Zhiyuan and Hou, Bohan and Yuan, Li},
  journal={arXiv preprint arXiv:2505.20275},
  year={2025}
}

@article{zhang2023magicbrush,
  title={Magicbrush: A manually annotated dataset for instruction-guided image editing},
  author={Zhang, Kai and Mo, Lingbo and Chen, Wenhu and Sun, Huan and Su, Yu},
  journal={Advances in Neural Information Processing Systems},
  volume={36},
  pages={31428--31449},
  year={2023}
}

@inproceedings{brooks2023instructpix2pix,
  title={Instructpix2pix: Learning to follow image editing instructions},
  author={Brooks, Tim and Holynski, Aleksander and Efros, Alexei A},
  booktitle={Proceedings of the IEEE/CVF conference on computer vision and pattern recognition},
  pages={18392--18402},
  year={2023}
}

@inproceedings{yu2025anyedit,
  title={Anyedit: Mastering unified high-quality image editing for any idea},
  author={Yu, Qifan and Chow, Wei and Yue, Zhongqi and Pan, Kaihang and Wu, Yang and Wan, Xiaoyang and Li, Juncheng and Tang, Siliang and Zhang, Hanwang and Zhuang, Yueting},
  booktitle={Proceedings of the Computer Vision and Pattern Recognition Conference},
  pages={26125--26135},
  year={2025}
}

@article{zhao2024ultraedit,
  title={Ultraedit: Instruction-based fine-grained image editing at scale},
  author={Zhao, Haozhe and Ma, Xiaojian Shawn and Chen, Liang and Si, Shuzheng and Wu, Rujie and An, Kaikai and Yu, Peiyu and Zhang, Minjia and Li, Qing and Chang, Baobao},
  journal={Advances in Neural Information Processing Systems},
  volume={37},
  pages={3058--3093},
  year={2024}
}

@inproceedings{xiao2025omnigen,
  title={Omnigen: Unified image generation},
  author={Xiao, Shitao and Wang, Yueze and Zhou, Junjie and Yuan, Huaying and Xing, Xingrun and Yan, Ruiran and Li, Chaofan and Wang, Shuting and Huang, Tiejun and Liu, Zheng},
  booktitle={Proceedings of the Computer Vision and Pattern Recognition Conference},
  pages={13294--13304},
  year={2025}
}

@article{zhang2025context,
  title={In-context edit: Enabling instructional image editing with in-context generation in large scale diffusion transformer},
  author={Zhang, Zechuan and Xie, Ji and Lu, Yu and Yang, Zongxin and Yang, Yi},
  journal={arXiv preprint arXiv:2504.20690},
  year={2025}
}

@article{liu2025step1x,
  title={Step1x-edit: A practical framework for general image editing},
  author={Liu, Shiyu and Han, Yucheng and Xing, Peng and Yin, Fukun and Wang, Rui and Cheng, Wei and Liao, Jiaqi and Wang, Yingming and Fu, Honghao and Han, Chunrui and others},
  journal={arXiv preprint arXiv:2504.17761},
  year={2025}
}

@article{shao2024deepseekmath,
  title={Deepseekmath: Pushing the limits of mathematical reasoning in open language models},
  author={Shao, Zhihong and Wang, Peiyi and Zhu, Qihao and Xu, Runxin and Song, Junxiao and Bi, Xiao and Zhang, Haowei and Zhang, Mingchuan and Li, YK and Wu, Yang and others},
  journal={arXiv preprint arXiv:2402.03300},
  year={2024}
}

@article{liao2025mogao,
  title={Mogao: An omni foundation model for interleaved multi-modal generation},
  author={Liao, Chao and Liu, Liyang and Wang, Xun and Luo, Zhengxiong and Zhang, Xinyu and Zhao, Wenliang and Wu, Jie and Li, Liang and Tian, Zhi and Huang, Weilin},
  journal={arXiv preprint arXiv:2505.05472},
  year={2025}
}

@article{argrpo,
  title={Ar-grpo: Training autoregressive image generation models via reinforcement learning},
  author={Yuan, Shihao and Liu, Yahui and Yue, Yang and Zhang, Jingyuan and Zuo, Wangmeng and Wang, Qi and Zhang, Fuzheng and Zhou, Guorui},
  journal={arXiv preprint arXiv:2508.06924},
  year={2025}
}

@article{ma2025stage,
  title={STAGE: Stable and Generalizable GRPO for Autoregressive Image Generation},
  author={Ma, Xiaoxiao and Qiu, Haibo and Zhang, Guohui and Zeng, Zhixiong and Yang, Siqi and Ma, Lin and Zhao, Feng},
  journal={arXiv preprint arXiv:2509.25027},
  year={2025}
}

@article{wang2025simplear,
  title={Simplear: Pushing the frontier of autoregressive visual generation through pretraining, sft, and rl},
  author={Wang, Junke and Tian, Zhi and Wang, Xun and Zhang, Xinyu and Huang, Weilin and Wu, Zuxuan and Jiang, Yu-Gang},
  journal={arXiv preprint arXiv:2504.11455},
  year={2025}
}

@article{jiang2025t2i,
  title={T2i-r1: Reinforcing image generation with collaborative semantic-level and token-level cot},
  author={Jiang, Dongzhi and Guo, Ziyu and Zhang, Renrui and Zong, Zhuofan and Li, Hao and Zhuo, Le and Yan, Shilin and Heng, Pheng-Ann and Li, Hongsheng},
  journal={arXiv preprint arXiv:2505.00703},
  year={2025}
}

@article{tong2025delving,
  title={Delving into RL for Image Generation with CoT: A Study on DPO vs. GRPO},
  author={Tong, Chengzhuo and Guo, Ziyu and Zhang, Renrui and Shan, Wenyu and Wei, Xinyu and Xing, Zhenghao and Li, Hongsheng and Heng, Pheng-Ann},
  journal={arXiv preprint arXiv:2505.17017},
  year={2025}
}

@article{zhang2025group,
  title={Group Critical-token Policy Optimization for Autoregressive Image Generation},
  author={Zhang, Guohui and Yu, Hu and Ma, Xiaoxiao and Zhang, JingHao and Pan, Yaning and Yao, Mingde and Xiao, Jie and Huang, Linjiang and Zhao, Feng},
  journal={arXiv preprint arXiv:2509.22485},
  year={2025}
}

@article{chen2025sharegpt,
  title={ShareGPT-4o-Image: Aligning Multimodal Models with GPT-4o-Level Image Generation},
  author={Chen, Junying and Cai, Zhenyang and Chen, Pengcheng and Chen, Shunian and Ji, Ke and Wang, Xidong and Yang, Yunjin and Wang, Benyou},
  journal={arXiv preprint arXiv:2506.18095},
  year={2025}
}

@article{lin2025uniworld,
  title={Uniworld: High-resolution semantic encoders for unified visual understanding and generation},
  author={Lin, Bin and Li, Zongjian and Cheng, Xinhua and Niu, Yuwei and Ye, Yang and He, Xianyi and Yuan, Shenghai and Yu, Wangbo and Wang, Shaodong and Ge, Yunyang and others},
  journal={arXiv preprint arXiv:2506.03147},
  year={2025}
}

@article{wu2025omnigen2,
  title={OmniGen2: Exploration to Advanced Multimodal Generation},
  author={Wu, Chenyuan and Zheng, Pengfei and Yan, Ruiran and Xiao, Shitao and Luo, Xin and Wang, Yueze and Li, Wanli and Jiang, Xiyan and Liu, Yexin and Zhou, Junjie and others},
  journal={arXiv preprint arXiv:2506.18871},
  year={2025}
}

@article{labs2025kontext,
  title={FLUX. 1 Kontext: Flow Matching for In-Context Image Generation and Editing in Latent Space},
  author={Labs, Black Forest and Batifol, Stephen and Blattmann, Andreas and Boesel, Frederic and Consul, Saksham and Diagne, Cyril and Dockhorn, Tim and English, Jack and English, Zion and Esser, Patrick and others},
  journal={arXiv preprint arXiv:2506.15742},
  year={2025}
}

@inproceedings{sushko2025realedit,
  title={Realedit: Reddit edits as a large-scale empirical dataset for image transformations},
  author={Sushko, Peter and Bharadwaj, Ayana and Lim, Zhi Yang and Ilin, Vasily and Caffee, Ben and Chen, Dongping and Salehi, Mohammadreza and Hsieh, Cheng-Yu and Krishna, Ranjay},
  booktitle={Proceedings of the Computer Vision and Pattern Recognition Conference},
  pages={13403--13413},
  year={2025}
}

@article{russakovsky2015imagenet,
  title={Imagenet large scale visual recognition challenge},
  author={Russakovsky, Olga and Deng, Jia and Su, Hao and Krause, Jonathan and Satheesh, Sanjeev and Ma, Sean and Huang, Zhiheng and Karpathy, Andrej and Khosla, Aditya and Bernstein, Michael and others},
  journal={International journal of computer vision},
  volume={115},
  number={3},
  pages={211--252},
  year={2015},
  publisher={Springer}
}

@misc{gemini2flash,
  title={Gemini 2.0 Flash},
  author={Google},
  year={2025},
  howpublished={\url{https://developers.googleblog.com/en/experiment-with-gemini-20-flash-native-image-generation}},
}

@misc{gpt4o,
  title={GPT-4o},
  author={OpenAI},
  year={2025},
  howpublished={\url{https://openai.com/index/introducing-4o-image-generation}},
}

@article{jiang2025infiniteyou,
  title={InfiniteYou: Flexible photo recrafting while preserving your identity},
  author={Jiang, Liming and Yan, Qing and Jia, Yumin and Liu, Zichuan and Kang, Hao and Lu, Xin},
  journal={arXiv preprint arXiv:2503.16418},
  year={2025}
}

@article{uno,
  title={Less-to-more generalization: Unlocking more controllability by in-context generation},
  author={Wu, Shaojin and Huang, Mengqi and Wu, Wenxu and Cheng, Yufeng and Ding, Fei and He, Qian},
  journal={arXiv preprint arXiv:2504.02160},
  year={2025}
}

@misc{googleGemini2,
  author={Google DeepMind},
  title={Gemini 2.0},
  howpublished={\url{https://gemini.google.com/}},
  year={2025}
}

@article{lin2025uniworldv1,
  title={UniWorld-V1: High-Resolution Semantic Encoders for Unified Visual Understanding and Generation},
  author={Lin, Bin and Li, Zongjian and Cheng, Xinhua and Niu, Yuwei and Ye, Yang and He, Xianyi and Yuan, Shenghai and Yu, Wangbo and Wang, Shaodong and Ge, Yunyang and others},
  journal={arXiv preprint arXiv:2506.03147},
  year={2025}
}

@article{prismbench,
  title={Flux-reason-6m \& prism-bench: A million-scale text-to-image reasoning dataset and comprehensive benchmark},
  author={Fang, Rongyao and Yu, Aldrich and Duan, Chengqi and Huang, Linjiang and Bai, Shuai and Cai, Yuxuan and Wang, Kun and Liu, Si and Liu, Xihui and Li, Hongsheng},
  journal={arXiv preprint arXiv:2509.09680},
  year={2025}
}

@inproceedings{yao2025reconstruction,
  title={Reconstruction vs. generation: Taming optimization dilemma in latent diffusion models},
  author={Yao, Jingfeng and Yang, Bin and Wang, Xinggang},
  booktitle={Proceedings of the Computer Vision and Pattern Recognition Conference},
  pages={15703--15712},
  year={2025}
}

@article{yu2024representation,
  title={Representation alignment for generation: Training diffusion transformers is easier than you think},
  author={Yu, Sihyun and Kwak, Sangkyung and Jang, Huiwon and Jeong, Jongheon and Huang, Jonathan and Shin, Jinwoo and Xie, Saining},
  journal={arXiv preprint arXiv:2410.06940},
  year={2024}
}

@article{gokaslan2023commoncanvas,
  title={CommonCanvas: An Open Diffusion Model Trained with Creative-Commons Images},
  author={Gokaslan, Aaron and Cooper, A Feder and Collins, Jasmine and Seguin, Landan and Jacobson, Austin and Patel, Mihir and Frankle, Jonathan and Stephenson, Cory and Kuleshov, Volodymyr},
  journal={arXiv preprint arXiv:2310.16825},
  year={2023}
}

@misc {BoerBohan2024Megalith10m,
  author = {Boer Bohan, Ollin},
  title = {Megalith-10m},
  howpublished = {\url{https://huggingface.co/datasets/madebyollin/megalith-10m}},
  url = {https://huggingface.co/datasets/madebyollin/megalith-10m},
  type = {dataset},
  year = {2024},
  month = {June},
  note = {Accessed: 2024-10-07}
}

@inproceedings{wang2023imagen,
	title        = {Imagen editor and editbench: Advancing and evaluating text-guided image inpainting},
	author       = {Wang, Su and Saharia, Chitwan and Montgomery, Ceslee and Pont-Tuset, Jordi and Noy, Shai and Pellegrini, Stefano and Onoe, Yasumasa and Laszlo, Sarah and Fleet, David J and Soricut, Radu and others},
	year         = 2023,
	booktitle    = {Proceedings of the IEEE/CVF conference on computer vision and pattern recognition},
	pages        = {18359--18369}
}

@inproceedings{sheynin2024emu,
	title        = {Emu edit: Precise image editing via recognition and generation tasks},
	author       = {Sheynin, Shelly and Polyak, Adam and Singer, Uriel and Kirstain, Yuval and Zohar, Amit and Ashual, Oron and Parikh, Devi and Taigman, Yaniv},
	year         = 2024,
	booktitle    = {Proceedings of the IEEE/CVF Conference on Computer Vision and Pattern Recognition},
	pages        = {8871--8879}
}

@inproceedings{zhang2024hive,
	title        = {Hive: Harnessing human feedback for instructional visual editing},
	author       = {Zhang, Shu and Yang, Xinyi and Feng, Yihao and Qin, Can and Chen, Chia-Chih and Yu, Ning and Chen, Zeyuan and Wang, Huan and Savarese, Silvio and Ermon, Stefano and others},
	year         = 2024,
	booktitle    = {Proceedings of the IEEE/CVF Conference on Computer Vision and Pattern Recognition},
	pages        = {9026--9036}
}

@article{hui2024hq,
	title        = {Hq-edit: A high-quality dataset for instruction-based image editing},
	author       = {Hui, Mude and Yang, Siwei and Zhao, Bingchen and Shi, Yichun and Wang, Heng and Wang, Peng and Zhou, Yuyin and Xie, Cihang},
	year         = 2024,
	journal      = {arXiv preprint arXiv:2404.09990}
}

@article{shen2024many,
	title        = {Many-to-many image generation with auto-regressive diffusion models},
	author       = {Shen, Ying and Zhang, Yizhe and Zhai, Shuangfei and Huang, Lifu and Susskind, Joshua M and Gu, Jiatao},
	year         = 2024,
	journal      = {arXiv preprint arXiv:2404.03109}
}

@article{pan2025ice,
	title        = {ICE-Bench: A Unified and Comprehensive Benchmark for Image Creating and Editing},
	author       = {Pan, Yulin and He, Xiangteng and Mao, Chaojie and Han, Zhen and Jiang, Zeyinzi and Zhang, Jingfeng and Liu, Yu},
	year         = 2025,
	journal      = {arXiv preprint arXiv:2503.14482}
}

@article{cui2025emu35,
  title={Emu3. 5: Native multimodal models are world learners},
  author={Cui, Yufeng and Chen, Honghao and Deng, Haoge and Huang, Xu and Li, Xinghang and Liu, Jirong and Liu, Yang and Luo, Zhuoyan and Wang, Jinsheng and Wang, Wenxuan and others},
  journal={arXiv preprint arXiv:2510.26583},
  year={2025}
}

@article{li2025onecat,
  title={Onecat: Decoder-only auto-regressive model for unified understanding and generation},
  author={Li, Han and Peng, Xinyu and Wang, Yaoming and Peng, Zelin and Chen, Xin and Weng, Rongxiang and Wang, Jingang and Cai, Xunliang and Dai, Wenrui and Xiong, Hongkai},
  journal={arXiv preprint arXiv:2509.03498},
  year={2025}
}

@article{kou2024orthus,
  title={Orthus: Autoregressive interleaved image-text generation with modality-specific heads},
  author={Kou, Siqi and Jin, Jiachun and Liu, Zhihong and Liu, Chang and Ma, Ye and Jia, Jian and Chen, Quan and Jiang, Peng and Deng, Zhijie},
  journal={arXiv preprint arXiv:2412.00127},
  year={2024}
}

@article{xie2025show,
  title={Show-o2: Improved Native Unified Multimodal Models},
  author={Xie, Jinheng and Yang, Zhenheng and Shou, Mike Zheng},
  journal={arXiv preprint arXiv:2506.15564},
  year={2025}
}

@misc{labs2025flux1kontextflowmatching,
      title={FLUX.1 Kontext: Flow Matching for In-Context Image Generation and Editing in Latent Space},
      author={Black Forest Labs and Stephen Batifol and Andreas Blattmann and Frederic Boesel and Saksham Consul and Cyril Diagne and Tim Dockhorn and Jack English and Zion English and Patrick Esser and Sumith Kulal and Kyle Lacey and Yam Levi and Cheng Li and Dominik Lorenz and Jonas Müller and Dustin Podell and Robin Rombach and Harry Saini and Axel Sauer and Luke Smith},
      year={2025},
      eprint={2506.15742},
      archivePrefix={arXiv},
      primaryClass={cs.GR},
      url={https://arxiv.org/abs/2506.15742},
}

@article{hu2022lora,
  title={Lora: Low-rank adaptation of large language models.},
  author={Hu, Edward J and Shen, Yelong and Wallis, Phillip and Allen-Zhu, Zeyuan and Li, Yuanzhi and Wang, Shean and Wang, Lu and Chen, Weizhu and others},
  journal={ICLR},
  volume={1},
  number={2},
  pages={3},
  year={2022}
}

@inproceedings{liu2024improved,
  title={Improved baselines with visual instruction tuning},
  author={Liu, Haotian and Li, Chunyuan and Li, Yuheng and Lee, Yong Jae},
  booktitle={Proceedings of the IEEE/CVF conference on computer vision and pattern recognition},
  pages={26296--26306},
  year={2024}
}

@inproceedings{siglip,
  title={Sigmoid loss for language image pre-training},
  author={Zhai, Xiaohua and Mustafa, Basil and Kolesnikov, Alexander and Beyer, Lucas},
  booktitle={Proceedings of the IEEE/CVF international conference on computer vision},
  pages={11975--11986},
  year={2023}
}

@inproceedings{teed2020raft,
  title={Raft: Recurrent all-pairs field transforms for optical flow},
  author={Teed, Zachary and Deng, Jia},
  booktitle={European conference on computer vision},
  pages={402--419},
  year={2020},
  organization={Springer}
}

@inproceedings{wang2025koala,
  title={Koala-36m: A large-scale video dataset improving consistency between fine-grained conditions and video content},
  author={Wang, Qiuheng and Shi, Yukai and Ou, Jiarong and Chen, Rui and Lin, Ke and Wang, Jiahao and Jiang, Boyuan and Yang, Haotian and Zheng, Mingwu and Tao, Xin and others},
  booktitle={Proceedings of the Computer Vision and Pattern Recognition Conference},
  pages={8428--8437},
  year={2025}
}

@article{li2024omnicorpus,
  title={Omnicorpus: A unified multimodal corpus of 10 billion-level images interleaved with text},
  author={Li, Qingyun and Chen, Zhe and Wang, Weiyun and Wang, Wenhai and Ye, Shenglong and Jin, Zhenjiang and Chen, Guanzhou and He, Yinan and Gao, Zhangwei and Cui, Erfei and others},
  journal={arXiv preprint arXiv:2406.08418},
  year={2024}
}

@inproceedings{su2025nemotron,
  title={Nemotron-cc: Transforming common crawl into a refined long-horizon pretraining dataset},
  author={Su, Dan and Kong, Kezhi and Lin, Ying and Jennings, Joseph and Norick, Brandon and Kliegl, Markus and Patwary, Mostofa and Shoeybi, Mohammad and Catanzaro, Bryan},
  booktitle={Proceedings of the 63rd Annual Meeting of the Association for Computational Linguistics (Volume 1: Long Papers)},
  pages={2459--2475},
  year={2025}
}

@article{zhou2025megamath,
  title={Megamath: Pushing the limits of open math corpora},
  author={Zhou, Fan and Wang, Zengzhi and Ranjan, Nikhil and Cheng, Zhoujun and Tang, Liping and He, Guowei and Liu, Zhengzhong and Xing, Eric P},
  journal={arXiv preprint arXiv:2504.02807},
  year={2025}
}
